\documentclass{article}

\usepackage[dblblindworkshop, final]{neurips_2026}
\workshoptitle{Interpretability as a Science}

\usepackage{times}
\usepackage{latexsym}
\usepackage[T2A,T1]{fontenc}
\usepackage[utf8]{inputenc}
\usepackage[russian,english]{babel}
\usepackage{microtype}
\usepackage{inconsolata}
\usepackage{graphicx}
\usepackage{booktabs}
\usepackage{longtable}
\usepackage{amsmath}
\usepackage{amssymb}
\usepackage{tikz}
\usetikzlibrary{arrows.meta, positioning, calc, fit, intersections}
\usepackage{subcaption}
\usepackage{xcolor}
\usepackage{xurl}
\usepackage{hyperref}
\hypersetup{colorlinks=true, linkcolor=[rgb]{0.0,0.25,0.55}, citecolor=[rgb]{0.0,0.40,0.25}, urlcolor=[rgb]{0.0,0.25,0.55}}
\usepackage{cleveref}
\usepackage{graphicx}
\graphicspath{{figures/}{screenshots/}}

\usepackage{iftex}
\iftutex
  \usepackage{fontspec}
  \usepackage{xeCJK}
  \newfontfamily\devafont{NotoSerifDevanagari-Regular.ttf}[Script=Devanagari]
  \newfontfamily\mlymfont{NotoSerifMalayalam-Regular.ttf}[Script=Malayalam]
  \newfontfamily\thaiscriptfont{NotoSerifThai-Regular.ttf}[Script=Thai]
  \newfontfamily\bengalifont{NotoSerifBengali-Regular.ttf}[Script=Bengali]
  \newfontfamily\arabscriptfont{NotoNaskhArabic-Regular.ttf}[Script=Arabic]
  \newCJKfontfamily\korfont{NotoSerifCJKkr-subset.otf}
  \babelfont[russian]{rm}[BoldFont=NotoSerif-Bold.ttf, ItalicFont=NotoSerif-Italic.ttf, BoldItalicFont=NotoSerif-BoldItalic.ttf]{NotoSerif-Regular.ttf}
  \newfontfamily\cyrfont{NotoSerif-Regular.ttf}[BoldFont=NotoSerif-Bold.ttf, ItalicFont=NotoSerif-Italic.ttf, BoldItalicFont=NotoSerif-BoldItalic.ttf, Script=Cyrillic]
  \newfontfamily\cyrttfont{DejaVuSansMono.ttf}[BoldFont=DejaVuSansMono-Bold.ttf, Script=Cyrillic]
  \usepackage{newunicodechar}
  \newfontfamily\symbolfont{DejaVuSans.ttf}
  \newunicodechar{↵}{{\symbolfont\char"21B5}}
\else
  \providecommand\devafont{}\providecommand\mlymfont{}\providecommand\thaiscriptfont{}%
  \providecommand\bengalifont{}\providecommand\arabscriptfont{}\providecommand\korfont{}%
  \providecommand\cyrfont{}\providecommand\cyrttfont{}%
\fi
\usepackage{truncate}
\usepackage{tablefootnote}

\newcommand{\TODO}[1]{}

\newlength{\phw}

\newcommand{\Nmodels}{13}

\newif\ifscaffold       \scaffoldfalse
\newif\ifdraftappendix  \draftappendixfalse
\newif\ifcircuitcases   \circuitcasestrue

\title{Comparing Sparsified MLPs and Transcoders: \\ Many Transcoder Features Are Neurons in Disguise}

\author{
  Stepan Shabalin\thanks{\hspace{0.4em}Equal contribution.}\hspace{0.5em}\thanks{\hspace{0.4em}Correspondence to \texttt{sshabalin3@gatech.edu}.} \\
  Anthropic Fellows \\
  Georgia Institute of Technology
  \And
  Mateusz Piotrowski$^*$ \\
  Anthropic
}

\begin{document}
\maketitle

\begin{center}
\begin{minipage}{\linewidth}
  \centering
  \resizebox{\linewidth}{!}{%
  \begin{tikzpicture}[
    >=Latex,
    box/.style={draw, rounded corners, inner sep=4pt, align=center},
    resd/.style={circle, minimum size=4mm, inner sep=0pt, fill=gray!50},
    neuron/.style={circle, minimum size=4mm, inner sep=0pt},
    selneuron/.style={circle, minimum size=4.6mm, inner sep=0pt,
                      fill=orange!50, draw=orange!95!black, line width=1.5pt},
    feature/.style={circle, minimum size=4mm, inner sep=0pt, fill=white, draw=black, line width=0.4pt},
    selfeature/.style={circle, minimum size=4.6mm, inner sep=0pt,
                       fill=teal!65, draw=teal!75!black, line width=1.5pt},
    flow/.style={-Latex, thick, gray!60!black},
    line/.style={thick, gray!60!black},
    encdim/.style={gray!50, thin},
    enca/.style={line width=1.0pt, orange!95!black},
    encb/.style={line width=1.0pt, teal!75!black},
    cossim/.style={<->, dashed, line width=0.7pt, gray!30!black},
    every node/.style={font=\bfseries\normalsize},
    wlab/.style={font=\bfseries\normalsize, inner sep=1pt},
    hdr/.style={font=\bfseries\large},
    cmark/.style={circle, draw=gray!30!black, fill=white, inner sep=0.5pt, minimum size=3.6mm,
                 font=\bfseries\footnotesize},
  ]
  \node[resd] (mlpin) at (0, 0) {};
  \node[above=2mm of mlpin] {MLP input};
  \node[neuron, fill=orange!22] (n1) at (8.0, 2.37) {};
  \node[selneuron] (n2) at (8.0, 1.82) {};
  \node[neuron, fill=orange!36] (n3) at (8.0, 1.27) {};
  \node[selneuron] (n4) at (8.0, 0.72) {};
  \node[box, fit=(n1) (n4), label={[hdr, yshift=-0.5mm]above:Sparsified MLP}] (mlp) {};
  \node[feature] (f1) at (8.0, -0.72) {};
  \node[feature] (f2) at (8.0, -1.22) {};
  \node[selfeature] (f3) at (8.0, -1.72) {};
  \node[feature] (f4) at (8.0, -2.22) {};
  \node[selfeature] (f5) at (8.0, -2.72) {};
  \node[feature] (f6) at (8.0, -3.22) {};
  \node[box, fit=(f1) (f6), label={[hdr, yshift=0.5mm]below:Transcoder}] (tc) {};
  \node[resd] (mlpout) at (13.0, 1.55) {};
  \node[above=2mm of mlpout] {Sparsified MLP output};
  \node[resd] (tcout)  at (13.0, -1.97) {};
  \node[below=2mm of tcout] {Transcoder output};
  \coordinate (encfork) at ([xshift=8mm] mlpin.east);
  \draw[line] (mlpin.east) -- (encfork);
  \coordinate (decfork_mlp) at ([xshift=-5mm] mlpout.west);
  \coordinate (decfork_tc)  at ([xshift=-5mm] tcout.west);
  \draw[flow] (decfork_mlp) -- (mlpout);
  \draw[flow] (decfork_tc)  -- (tcout);
  \draw[encdim] (encfork) -- (mlp.west |- n1);
  \draw[enca, name path=enc_mlp] (encfork) -- (mlp.west |- n2)
    node[pos=0.6, above, sloped, wlab, text=orange!50!black] {$W_{\mathrm{up}}^{(i)}\!\pm\! W_{\mathrm{gate}}^{(i)}$};
  \draw[encdim] (encfork) -- (mlp.west |- n3);
  \draw[enca] (encfork) -- (mlp.west |- n4);
  \draw[encdim] (encfork) -- (tc.west |- f1);
  \draw[encdim] (encfork) -- (tc.west |- f2);
  \draw[encb] (encfork) -- (tc.west |- f3);
  \draw[encdim] (encfork) -- (tc.west |- f4);
  \draw[encb, name path=enc_tc] (encfork) -- (tc.west |- f5)
    node[pos=0.6, below, sloped, wlab, text=teal!40!black] {$W_{\mathrm{enc}}^{(j)}$};
  \draw[encdim] (encfork) -- (tc.west |- f6);
  \draw[encdim] (mlp.east |- n1) -- (decfork_mlp);
  \draw[enca, name path=dec_mlp] (mlp.east |- n2) -- (decfork_mlp)
    node[pos=0.5, above, sloped, wlab, text=orange!50!black] {$W_{\mathrm{down}}^{(i)}$};
  \draw[encdim] (mlp.east |- n3) -- (decfork_mlp);
  \draw[enca] (mlp.east |- n4) -- (decfork_mlp);
  \draw[encdim] (tc.east |- f1) -- (decfork_tc);
  \draw[encdim] (tc.east |- f2) -- (decfork_tc);
  \draw[encb] (tc.east |- f3) -- (decfork_tc);
  \draw[encdim] (tc.east |- f4) -- (decfork_tc);
  \draw[encb, name path=dec_tc] (tc.east |- f5) -- (decfork_tc)
    node[pos=0.5, below, sloped, wlab, text=teal!40!black] {$W_{\mathrm{dec}}^{(j)}$};
  \draw[encdim] (tc.east |- f6) -- (decfork_tc);
  \path[overlay, name path=enc_vline] (5.0, -8) -- (5.0, 8);
  \path[name intersections={of=enc_mlp and enc_vline, by=enc_topi}];
  \path[name intersections={of=enc_tc  and enc_vline, by=enc_boti}];
  \draw[cossim] (enc_topi) -- (enc_boti);
  \path[overlay, name path=dec_vline] (10.5, -8) -- (10.5, 8);
  \path[name intersections={of=dec_mlp and dec_vline, by=dec_topi}];
  \path[name intersections={of=dec_tc  and dec_vline, by=dec_boti}];
  \draw[cossim] (dec_topi) -- (dec_boti);
  \node[anchor=west, fill=white, inner sep=1pt] at (5.12, 0) {$\cos\!\big(W_{\mathrm{up}}^{(i)}\!\pm\! W_{\mathrm{gate}}^{(i)},\,W_{\mathrm{enc}}^{(j)}\big) > 0.7$};
  \node[anchor=west, fill=white, inner sep=1pt] at (10.62, 0) {$\cos\!\big(W_{\mathrm{down}}^{(i)},\, W_{\mathrm{dec}}^{(j)}\big) > 0.7$};
  \end{tikzpicture}
  }
  
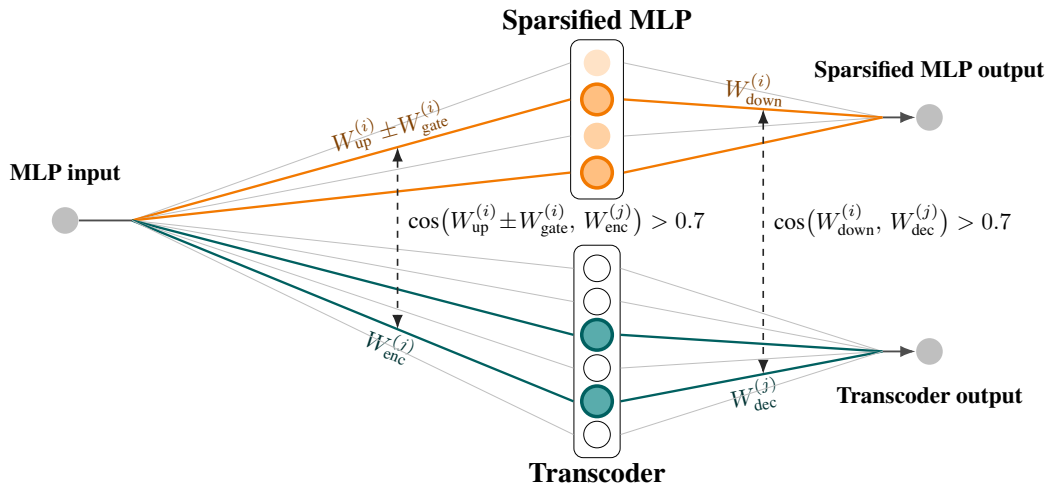
\captionof{figure}{We find some transcoder features are
    \emph{neuron features}: their encoder and decoder match an individual
    sparsified MLP neuron at cosine similarity ${>}\,0.7$. Over $50\%$ of activations in late layers are from neuron features; neuron features are discovered across training runs and carry most of the faithfulness to the original model.}
    \label{fig:schematic}
\end{minipage}
\end{center}

\begin{abstract}
MLP neurons in transformer language models are widely understood to be polysemantic,
motivating transcoders --- sparse, learned approximations of MLP layers --- as interpretable
replacement models for mechanistic analysis. However, recent work shows that MLP neurons can be used to recover sparse, interpretable circuits \citep{arora2026circuits, ye2025ffkvsae}, raising the question of whether transcoders are necessary for circuit analysis.

We define \emph{neuron features} as transcoder features whose weights are strongly aligned with individual MLP neurons, and show that they account for a substantial fraction of active features in late layers across \Nmodels{} open-source transcoders. Only neuron features have a stable correspondence with something in the underlying MLP: other features vary more between training runs, do not match the MLP's Jacobian, and change less significantly from their random initialization.
Our findings reframe MLP neurons as a stronger basis for interpretability than commonly assumed, and call for better methods of attributing the non-neuron structure that is used by the model.

\end{abstract}

\section{Introduction}
\label{sec:intro}

Polysemanticity is the standard obstacle to direct mechanistic analysis of
MLPs, and the field has responded by replacing neurons with learned sparse
dictionaries as the basic units of analysis.

Individual neurons activate on many unrelated concepts
\citep{elhage2022superposition}, and are widely taken to be the wrong
basic unit for mechanistic analysis.
In response, the field has moved to
\emph{learned sparse dictionaries} over the model as the substrate for investigation.

\begin{figure}[!t]
  \centering
  \begin{subfigure}[t]{0.44\linewidth}
    \vspace{0pt}%
    \centering
    \includegraphics[width=\linewidth]{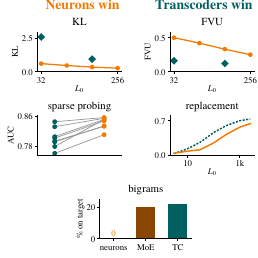}
    \caption{Summary of our evaluations: transcoders win on reconstruction and fine-grained concept recovery, but pruned neurons make for a more faithful replacement model.}
    \label{fig:tradeoff}
  \end{subfigure}\hspace{0.06\linewidth}
  \begin{subfigure}[t]{0.44\linewidth}
    \vspace{0pt}%
    \centering
    \includegraphics[width=\linewidth]{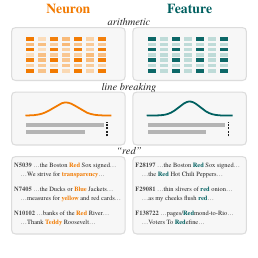}
    \caption{We present case studies in which we show behaviors previously interpreted through transcoders or neurons are reflected in both through the framework of neuron features.}
    \label{fig:casestudies}
  \end{subfigure}
  \caption{A summary of findings from our evaluations and case studies.}
\end{figure}

Sparse autoencoders trained on MLP activations
\citep{bricken2023monosemanticity, cunningham2023sparse}, scaled to
frontier models \citep{templeton2024scaling}, and transcoders,
sparse piecewise linear approximations of the MLP layer's input-output map
\citep{dunefsky2024transcoders}, underlie modern attribution graph
methods \citep{ameisen2025circuit, lindsey2025biology} and case studies
of frontier model behavior \citep{gurnee2025manifolds}.
This choice is motivated largely by automated interpretability metrics: learned dictionary
features score as more interpretable than individual MLP neurons
\citep{paulo2024autointerp, paulo2025transcodersbeat}.

Two recent results contradict this model of neuron polysemanticity. \citet{arora2026circuits} find that, when applying algorithms for transcoder-based attribution to raw MLP neurons, it is possible to (in some settings) recover circuits as sparse and faithful as
those built from learned features. \citet{ye2025ffkvsae} shows interpreting thresholded neurons as SAEs can match learned dictionaries on standard benchmarks.

Motivated by these findings, we set out to answer the question: what do transcoders contribute to circuit analysis that MLP neurons do not have out of the box?

In this paper, we describe a class of transcoder features,
\emph{neuron features} (Section~\ref{sec:nfc}), whose encoder, decoder,
and activation pattern all track a single MLP neuron.
Across \Nmodels{}
open-source transcoders on models from the Gemma, Qwen, and Llama families, neuron
features account for a substantial (near $50\%$) fraction of active
features in late layers (Section~\ref{sec:nfc:prevalence}) and exhaust
most of the structure learned by transcoders that aligns with the underlying
MLP.

Once neuron features are accounted
for, what remains of the transcoder varies across training
seeds (Section~\ref{sec:nfc:cross-seed}), does not match the MLP's first-order computation (Jacobian) (Section~\ref{sec:nfc:faithfulness}), but does represent many additional interpretable concepts not in the original decoder basis (Appendix~\ref{app:eval-results:bigram}).
Additionally, the training method and scale of the transcoder influence the amount of neuron features (Section~\ref{sec:nfc:prevalence}) and the extent to which claims about non-neuron features apply.

\noindent Our main contributions are as follows:
\begin{itemize}
  \item
  We identify and characterize \textbf{neuron features} (Section~\ref{sec:nfc}):
  transcoder features whose encoder and decoder directions align with
  individual MLP neurons and whose activations track those neurons.

  \item We show that \textbf{neuron features are a significant and possibly the
  only piece of consistent neuron-aligned structure transcoders learn}, with three
  independent pieces of evidence: linearizations
  of the transcoder match the MLP only on neuron features; only neuron
  features reoccur across random seeds; and the remaining features'
  weights remain close to the initialization
  (Sections~\ref{sec:nfc:faithfulness},~\ref{sec:nfc:cross-seed},~\ref{sec:nfc:init-distance}).

  We show that the
  non-neuron portion of transcoder dictionaries can be steered: with
  training-time alignment interventions, the resulting dictionary's features are
  sparse combinations of neurons, at little cost to reconstruction
  (Section~\ref{sec:nonneuron:ni},Appendix~\ref{sec:nonneuron:snrl}).
\end{itemize}

\section{Evaluating thresholded neurons as features}
\label{sec:mlp}

\begin{figure*}[t]
  \centering
  \begin{subfigure}[t]{0.64\textwidth}
    \vspace{0pt}%
    \centering
    \includegraphics{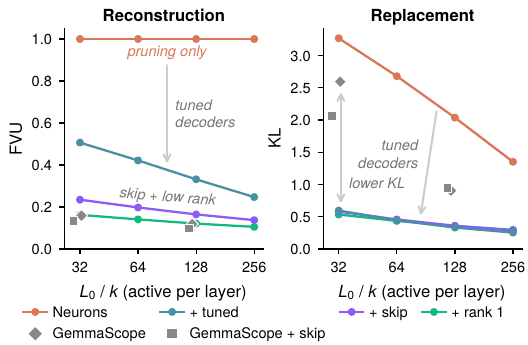}
    \caption{Reconstruction (FVU) vs replacement (KL) for neuron-based
      replacement models on Gemma~3 4B. Let $y$ denote the top-$k$-pruned
      MLP output and $a_i$ the (sign-split) activation of neuron $i$.
      \textbf{Tuned}: a $d_\mathrm{model} \times d_\mathrm{model}$
      linear map applied to $y$, trained jointly on FVU and KL.
      \textbf{Skip}: a learnable skip connection from the layer input.
      \textbf{Rank-1}: above plus a per-neuron rank-1 contribution gated
      on activation.
      See Appendix~\ref{app:setup} for details.
      }
    \label{fig:pareto}
  \end{subfigure}\hfill
  \begin{subfigure}[t]{0.34\textwidth}
    \vspace{0pt}%
    \centering
    \includegraphics{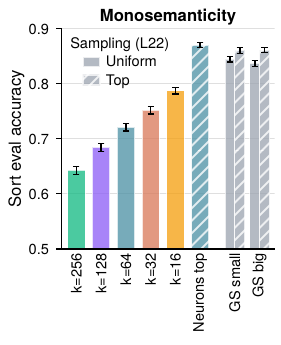}
    \caption{Sort eval accuracy (Gemma~3 4B, L22). Top examples of
      pruned neurons score comparably to Gemmascope features; accuracy
      degrades when sampling quantiles. }
    \label{fig:sort-eval}
  \end{subfigure}
  \caption{Reconstruction and replacement (a) and sort eval monosemanticity (b) for
   neurons vs transcoders. Top neuron activations explain most of the model's computation and are similarly interpretable to transcoder feature activations.}
  \label{fig:replacement-and-sort-eval}
\end{figure*}

In this section, we argue that Gated MLP neurons --- once split by sign and
selected via top-$k$ --- are a strong basis for mechanistic
interpretability:
\begin{itemize}
  \item \textbf{Every neuron has two well-defined encoder/decoder directions}:
    neurons are approximately bilinear at high activations, so $W_\mathrm{up}^{(i)} \pm W_\mathrm{gate}^{(i)}$
    serve as the neuron's effective encoders, and $W_\mathrm{down}^{(i)}$
    as its decoder, see \Cref{sec:mlp:directions}. Top-K thresholding with $k \approx 50$ is a good way to turn continuous activations of gated MLPs into binary on/off values.
  \item \Cref{sec:mlp:kl} shows \textbf{sparsified neurons beat transcoders on KL}, a metric of causal importance, while losing on
    FVU.
 This is evidence much of what transcoders
    learn increases reconstruction only and is not relevant to model behavior. Accordingly, \Cref{sec:nfc:faithfulness} shows transcoders are incosistent with linearizations of the original MLP.
  \item \Cref{sec:mlp:polysemantic} shows that, at broad activation ranges, individual neurons are polysemantic. However, \textbf{top activations of neurons can be as monosemantic as transcoder feature activations}. We operationalise this with sort evals, and decompose polysemantic neurons via clustering in Appendix~\ref{sec:clustering}.
  \item \textbf{Neurons match or exceed learned dictionaries on sparse probing, but lose on detection of fine-grained concepts like bigrams}, see \Cref{sec:mlp:probing}. If very specific features or analysis of early layers is necessary, transcoders may provide an advantage; however, MoEs may make up for this difference by having many more neurons.
  \item \textbf{Attribution graphs built from neurons have lower Replacement Scores} but are useful for circuit analysis, containing interpretable features for distress (\Cref{sec:mlp:attrib}).
\end{itemize}

\subsection{Neurons have single directions}
\label{sec:mlp:directions}

Once split by the sign of its activation, each gated MLP neuron has a single encoder direction and a single decoder direction. We work with gated MLPs \citep{shazeer2020glu}. Writing
$g_i = W_\mathrm{gate}^{(i)}\, x$ and
$u_i = W_\mathrm{up}^{(i)}\, x$ for the $i$th row of each projection,
we refer to the elementwise hidden activations
$a_i = \sigma(g_i)\, u_i$ as the layer's \emph{neurons}. At high
$|a_i|$ the neuron is approximately bilinear and selective for a single
input direction: $W_\mathrm{up}^{(i)} + W_\mathrm{gate}^{(i)}$ when
$u_i > 0$ and $W_\mathrm{up}^{(i)} - W_\mathrm{gate}^{(i)}$ when
$u_i < 0$ (derivation and details in
Appendix~\ref{app:neuron-geometry}).

We refer to $W_\mathrm{up}^{(i)} + W_\mathrm{gate}^{(i)}$ and
$W_\mathrm{up}^{(i)} - W_\mathrm{gate}^{(i)}$ as the neuron's positive
and negative \emph{effective encoders}, and treat each neuron as two
features --- one per sign of $a_i$ --- in all subsequent comparisons.
The two effective encoders are approximately orthogonal (since
$W_\mathrm{up}^{(i)}$ and $W_\mathrm{gate}^{(i)}$ are), and the
corresponding decoders have opposite signs ($+W_\mathrm{down}^{(i)}$
and $-W_\mathrm{down}^{(i)}$).

To place neurons and learned transcoders on equal footing, we sparsify
the MLP via top-$k$ thresholding on the (sign-split) neuron
activations.\footnote{Activation magnitudes vary substantially across layers in
models without post-feedforward LayerNorm, so we threshold layer-wise,
mirroring the implicit selection in TopK transcoders.} Under this
pruning, the MLP itself can be seen as a narrow transcoder: the input is encoded as the top-$k$ sign-split activations, and $W_\mathrm{down}$ reconstructs the layer output (see
\S\ref{sec:mlp:kl}).

\subsection{Replacement and reconstruction}
\label{sec:mlp:kl}

Following \citet{ameisen2025circuit}, we aim to construct a replacement
model that matches the original's predictions while being much
sparser. \citet{ameisen2025circuit} report that neurons are
Pareto-worse than transcoders on both reconstruction (FVU\footnote{Fraction of variance unexplained, normalized L2 loss.}) and
replacement (KL divergence); we replicate this on Gemma~3 4B, where untrained
pruned neurons have high KL and FVU above 1. We find replacement performance
can be recovered (\Cref{fig:pareto}) by training a linear map applied to the pruned MLP
output, $\hat{y} = W_\mathrm{recon}\, y$
(more details in Appendix~\ref{app:setup:rep}).

With this correction, KL divergence drops
below that of transcoders, though FVU remains substantially higher.
Some of the remaining FVU gap is closed by adding terms to the reconstruction (\Cref{fig:pareto}) which improve
FVU but not KL.

The pattern of neurons after minor finetuning producing replacement models that outperform transcoders on KL replicates across models (Appendix~\ref{app:setup}). These results indicate that the neurons already capture the
structure relevant to replacement, and reconstruction error can be reduced without affecting KL, meaning
that some of the structure transcoders capture is not relevant to
model behavior.

\providecommand{\peak}[1]{\colorbox{gray!18}{\textbf{#1}}}
\providecommand{\fcard}[6]{%
  \noindent\fbox{%
  \begin{minipage}{\dimexpr\linewidth-2\fboxsep-2\fboxrule\relax}\small
  \textbf{#1}\quad L#2 / N#3\\
  \emph{#4}\\[-5pt]
  \rule{\linewidth}{0.2pt}\\[1pt]
  \begin{tabular}{@{}r@{\hspace{4pt}}r@{\hspace{2pt}}p{0.5\linewidth}@{}}
  #5
  \end{tabular}\\[2pt]
  \rule{\linewidth}{0.2pt}\\[2pt]
  #6
  \end{minipage}}\vspace{6pt}}
\providecommand{\ctxrow}[4]{#1 & #2 & \peak{#3}\hspace{2pt}#4 \\}

\subsection{Sparse probing, bigrams, and concept representation}
\label{sec:mlp:probing}

Transcoders are expected to contain features for common concepts beyond ones already present in neurons.
Sparse probing is designed to find features that are sensitive to specific concepts, so we would expect transcoders to outperform neurons because they can represent many more specific concepts.

\citet{kantamneni2025sparseprobing} report that SAEs do not outperform
standard logistic regression baselines on sparse probing their 113 binary classification tasks. We extend the comparison to
top-$k$-sparsified neurons against transcoders. Across eight transcoders, spanning Gemma~2/3, Qwen~3, and
Llama~3.1/3.2, neurons match or exceed the transcoder on every
model (Table~\ref{tab:sparse-probing}).

We investigated which tasks the dictionary outperforms neurons on and
found these are consistently the bigram-detection tasks from
\citet{gurnee2023haystack}. We therefore extended the benchmark to
1\,500 phrase bigrams sampled from FineWeb
(Appendix~\ref{app:eval-methods:bigram}). Even though transcoders win the comparison,
neurons score a high mean AUC of $0.988$, making it unclear whether transcoders truly find missing bigrams.
Therefore, we sanity-check if the selected feature actually fires
on the target bigram. Inspecting the top max-activating context on
FineWeb, the dictionary feature fires on the target bigram for $75.6\%$
of bigrams, while neurons fire on close to $0\%$.

Bigrams are an example of commonly occuring features that are individually rare, and so are likely to have sparse activations. We find bigrams are effortlessly captured by transcoders while not represented monosemantically at all by neurons. At the same time, sparse probing can show high accuracy even if a concept is not represented in the feature, so it is difficult to make a conclusion about whether transcoders are better at representing other types of features.

There is, however, a reason to expect modern MoE models to be more capable of representing fine-grained features: as \Cref{tab:bigram-probing} shows, Gemma 4 26B A4B's neurons match bigrams in $20.1\%$ of cases, which is much higher than any dense model could hope to achieve.

\subsection{Strong attribution graphs from neurons}
\label{sec:mlp:attrib}

\providecommand{\peak}[1]{\colorbox{gray!18}{\textbf{#1}}}
\providecommand{\fcard}[6]{%
  \noindent\fbox{%
  \begin{minipage}{\dimexpr\linewidth-2\fboxsep-2\fboxrule\relax}\small
  \textbf{#1}\quad L#2 / N#3\\
  \emph{#4}\\[-5pt]
  \rule{\linewidth}{0.2pt}\\[1pt]
  \begin{tabular}{@{}r@{\hspace{4pt}}r@{\hspace{2pt}}p{0.5\linewidth}@{}}
  #5
  \end{tabular}\\[2pt]
  \rule{\linewidth}{0.2pt}\\[2pt]
  #6
  \end{minipage}}\vspace{6pt}}
\providecommand{\ctxrow}[4]{#1 & #2 & \peak{#3}\hspace{2pt}#4 \\}

We construct attribution graphs from MLP neurons by treating the
neuron activations as a feature basis for attribution graphs with RelP
\citep{lindsey2025biology,arora2026circuits,jafari2025relp}. Details are in
Appendix~\ref{app:circuits}.

We evaluate Replacement Score at matched $L_0$ across neuron
attribution and standard transcoders, following
\citet{lindsey2025biology} for selecting 30 random representative prompts.\footnote{Short Pile sequences
with rare, non-repeated target tokens that the base model predicts
correctly.} \Cref{fig:rs-l0-pareto} shows that Gemmascopes but not TopK transcoders have higher replacement score around the practical $L_0$ range of 10--100 than sparsified MLPs. In the limit, full MLPs have a replacement score of 1.0, but transcoders shift the replacement score curve to the left.

We show that neuron attribution graphs discover relevant features for
\emph{distress} (strong negative emotional states) on Gemma~3 27B-it.
We adapt the distress elicitation of
\citet{soligo2026gemma},\footnote{Three turns of impossible arithmetic
puzzles with firm user rejections, condensed into a single-turn
steered prompt for token efficiency.} then construct a short prompt
where the model can be distressed and attribute to the steering
vector following \citet{macar2026introspection}. We
take a base set MLP neurons downstream of the steering vector input node and
filter the candidate pool with Claude Haiku to leave
features interpretably corresponding to apology and distress. Details are in
Appendix~\ref{app:distress-setup}; in addition to the representative feature
in \Cref{fig:distress-card}, features from the circuit are shown in
Appendix~\ref{app:distress-cards}.

\subsection{Polysemanticity of neurons}
\label{sec:mlp:polysemantic}

To compare the interpretability of individual neurons and transcoder features, we use sort evals \citep{anthropicaug2024}, a measure of interpretability of feature activations.
The evaluation operationalises monosemanticity as
discriminability.
We run sort evals separately on individual sign-split neurons and on transcoder features.
Activating samples are drawn from two sources: the top-$k$
activations only, and from quantiles of the full activation distribution.

At the top of each feature's activation distribution, neurons score comparably to transcoder features (Figure~\ref{fig:sort-eval}). Under quantile sampling (higher $k$), scores of neurons degrade noticeably compared to transcoder features. In Appendix~\ref{sec:clustering}, we describe how a neuron's broader activations form multiple coherent sub-clusters, a model for the presence of polysemanticity.

\begin{figure*}[t!]
  \centering
  \includegraphics{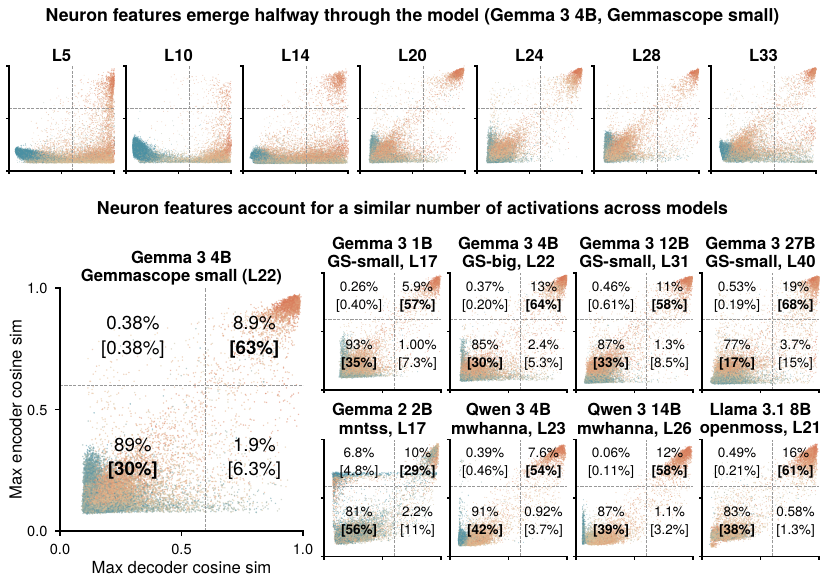}
  \caption{A distinct cluster of features has encoder and decoder
      cosine similarity ${>}\,0.6$ to one neuron. The cluster sharpens
      through mid layers (top) and recurs across models (bottom). Points are colored
      by firing frequency, from blue (rare) to orange
      (frequent). Each quadrant shows the fraction of features in it in the dictionary and all feature activations (bold, bracketed).}
  \label{fig:nfc-scatter}
\end{figure*}

\begin{figure*}[t!]
  \centering
  \begin{minipage}[t]{0.52\linewidth}
  \vspace{0pt}%
    \centering
    \includegraphics{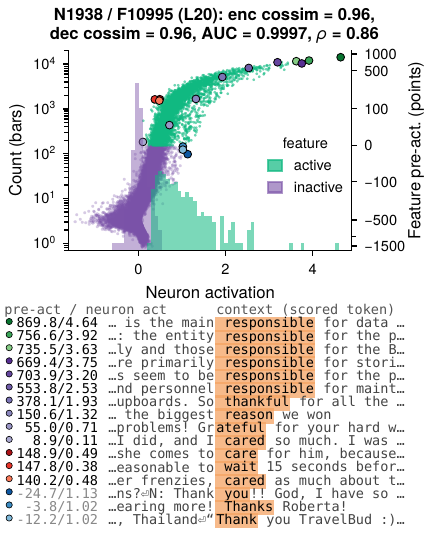}
    \captionof{figure}{Neuron and neuron feature pair (L20 N1938, F10995): the neuron's
      post-activation correlates with feature pre-activation (AUC${=}\,0.9997$,
      Spearman $\rho{=}0.86$). Green: feature is active, purple: feature is inactive.
      }
    \label{fig:nfc-cofiring}
  \end{minipage}\hfill
  \begin{minipage}[t]{0.46\linewidth}
  \vspace{0pt}%
    \centering
    \includegraphics{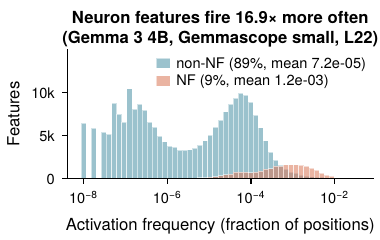}
    \captionof{figure}{Activation frequency distribution for neuron features
      vs.\ non-neuron features.}
    \label{fig:nf-freq-hist}

    \vspace{6pt}
    \centering
    \centering
    \footnotesize
    \setlength{\tabcolsep}{3pt}
    \begin{tabular}{l@{\hskip 6pt}ccc@{\hskip 6pt}ccc}
      \toprule
      & \multicolumn{3}{c}{count in dict}
      & \multicolumn{3}{c}{\% in activations} \\
      \cmidrule(lr){2-4} \cmidrule(lr){5-7}
      $M$ & shared & NF & both & shared & NF & both \\
      \midrule
      32k  &  9.8k &  8.1k &  7.6k & 29.6 & 25.0 & 23.3 \\
      65k  & 13.0k & 10.2k &  9.6k & 24.9 & 20.2 & 18.5 \\
      128k & 16.5k & 11.8k & 11.1k & 21.6 & 16.3 & 15.3 \\
      256k & 20.4k & 14.2k & 12.6k & 20.4 & 19.6 & 15.8 \\
      1M   & 22.0k & 11.5k & 10.5k & 13.7 & 12.8 & 10.5 \\
      \bottomrule
    \end{tabular}
    \captionof{table}{Counts in dictionaries and percentages in activations of: features with strong matches in another seed (\emph{shared}), neuron features (\emph{NF}), and their intersection
      (\emph{both})
      }
    \label{tab:cross-seed}
  \end{minipage}
\end{figure*}

\section{Many transcoder features correspond to neurons}
\label{sec:nfc}

\subsection{Definition}
\label{sec:nfc:def}

When comparing attribution graphs for the same model generated with
per-layer transcoders and
top-$k$-thresholded neurons, we find a
striking degree of overlap in the features the two methods select. For
many transcoder feature nodes in a given layer, there exists a neuron node whose
top and bottom logits, circuit connections, and maximum activations are all similar. Generalizing this observation, we call a transcoder feature a
\emph{neuron feature} when its weights, activation pattern, top/bottom
logits, and role in circuit correspond to those of a single MLP neuron.

To turn this label into a quantitative criterion, we plot per-feature
encoder and decoder cosine similarity to the best-matched neuron
(Figure~\ref{fig:nfc-scatter}). In most transcoders we evaluate, the
joint distribution is bimodal, with a tight high-similarity cluster of
features aligned with the same neuron on both sides. Operationally, we
define a transcoder feature as a neuron feature if its encoder and
decoder both exceed a cosine threshold to the same neuron's effective
encoder ($W_{\mathrm{up}}^{(i)}\!\pm\! W_{\mathrm{gate}}^{(i)}$) and
decoder ($W_{\mathrm{down}}^{(i)}$).\footnote{The threshold separating
the cluster from the bulk varies between $0.6$ and $0.9$ across
transcoders.} In the following section, we will argue neuron features are important to the transcoder's computation.

\subsection{Prevalence}
\label{sec:nfc:prevalence}

Across the 13 open-source transcoders we evaluate
(Table~\ref{tab:nfc-multimodel}), neuron features visually form a separate cluster and can be separated by a threshold of 0.6 or 0.7 cosine similarity in both encoder and decoder weights to some neuron (Figure~\ref{fig:nfc-scatter}); we use this as the definition for the claims below.

\textbf{Although neuron features remain a small fraction of the
dictionary, they account for over half of all feature activations in late layers.}
Table~\ref{tab:transcoders}: the fraction of neuron features in the dictionary
for late layers rises as $5.9\% \to 8.9\% \to 15.6\% \to 37.1\%$ from 1B to 27B, with the share in activations going from
$57.9\%$ to $\mathbf{82.8\%}$.\footnote{The fraction in dictionary rises monotonically with
model size in the Gemma family: $3.6 \to 7.2 \to 10.7 \to 17.7\%$ from 1B
to 27B, explained by the rising fraction of neurons in the model at a
constant transcoder width and training setup (see Appendix~\ref{app:non-neuron:scaling}).} This finding is replicated across models (\Cref{fig:nfc-scatter}).
Neuron features are much more prevalent in activations because they are $>10\times$ more frequent on average
(Figure~\ref{fig:nf-freq-hist}).

\textbf{Neuron features are only prevalent in activations halfway into the model's layers.}\footnote{Early-layer prevalence is variable and depends on initialization
    and training dynamics (\S\ref{sec:nonneuron:ni}). One possible
    cause is that early-layer MLP neurons are more polysemantic, so the
    transcoder objective is less likely to select them as feature
    targets than in late layers.}
    Figure~\ref{fig:nfc-per-layer}: on Gemma~3 4B and
    Qwen~3 4B, the percentage of neuron features climbs from near zero in
    early layers to over $50\%$ through the middle of the
    model.\footnote{In Qwen, this claim is not strictly true, but the transcoders  for that model are weak and do not have as many neuron features in general. The claim about attribution graphs is a stronger source of evidence.}

\textbf{In circuits, neuron features are common in late layers
    across pruning thresholds.}
    On the late third of layers, neuron features account for
    ${\geq}\,40\%$ of nodes in attribution graphs for every
    transcoder we tested except for one.\footnote{Llama 3.2 1B mntss (${\sim}\,38\%$). Full cross-model comparison and
    methodology in Appendix~\ref{app:circuits:cross-model}.} The neuron features are important to the circuit, as pruning the circuit by influence \citep{lindsey2025biology} keeps their amount roughly the same (e.g.\ $56$--$61\%$ on Gemma 3 4B from
    $L_0\,{=}\,10$ to $L_0\,{=}\,100$ features per token). The share of neuron features in early layers is much lower (${\leq}\,33\%$ across transcoders) and
    drops further with tighter pruning.

\textbf{Each MLP neuron eventually corresponds to at least one
transcoder feature given sufficient transcoder width.} Across a
$32{\text{k}}\to 1{\text{M}}$ width sweep of TopK transcoders at
Gemma~3 4B L22 (k${=}\,64$), the fraction of neurons matched by at
least one transcoder feature at $|\cos|\,{\geq}\,0.7$ rises from $78\%$
at $32$k features to $93\%$ at $256$k, then plateaus through $1$M
(full table in Appendix~\ref{app:nf-coverage-sweep}). At
Gemmascope L0-small ($262$k features), most matched neurons receive a
small handful of features (median $2$, mean $2.5$): one per sign of
the activation, plus the occasional second copy
(Figure~\ref{fig:neuron-coverage}, left).

\textbf{Neuron features fire roughly $3\times$ less often than their
matched neurons.} The median ratio of feature firing rate to
neuron firing rate on Gemma~3 4B L22 is $0.34$
(Figure~\ref{fig:neuron-coverage}, right); features fire on a
subset of the contexts that activate their neuron.

\subsection{Neuron features activate similarly to neurons}
\label{sec:nfc:cofiring}

We show that neuron features also activate on similar tokens to their neurons, and that, given a neuron feature activates, its neuron is likely also highly active.

\textbf{Neuron activations predict when their neuron features fire.} For each neuron feature, we threshold its neuron's per-token
post-activation to predict whether the feature fires at
that token. Across neuron features, the median AUC of this classifier is at least $0.977$ (we report $\max(\mathrm{AUC}, 1-\mathrm{AUC})$, because a feature may match either sign of a neuron). Figure~\ref{fig:nfc-cofiring} shows an $87.8$th percentile example at $0.9997$ AUC; more examples can be seen in Appendix~\ref{app:cofiring-panels}.

To find whether a neuron feature's neuron activates when the feature is active, we measure the rank of the neuron in the top K selection conditional on the rank of the feature.
 On Gemma~3 4B L22
(Figure~\ref{fig:rank-distribution}),
the median rank of the neuron is as low as $12$ for the top-firing feature and climbs to ${\sim}250$ for the $32^\text{nd}$ highest-firing.
\footnote{Given we take top 64--256 neurons per layer for circuit pruning, this means neuron features in low-L0 transcoders are $\approx 50\%$ likely to have a corresponding neuron in the neuron circuit.}

\subsection{Faithfulness concentrates in neuron features}
\label{sec:nfc:faithfulness}

\begin{figure*}[t!]
  \centering
  \begin{minipage}[t]{0.48\linewidth}
  \vspace{0pt}%
    \centering
    \includegraphics{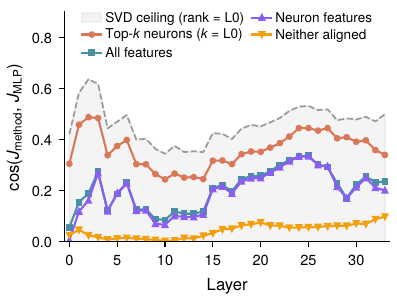}
    \captionof{figure}{Per-layer cosine similarity between the MLP's Jacobian and
      the transcoder's, Gemma~3 4B with Gemmascope L0-small. Top-$k$-pruned
      neurons, without any learned correction, match or exceed the full
      transcoder across late layers.}
    \label{fig:faithfulness-by-layer}
  \end{minipage}\hfill
  \begin{minipage}[t]{0.48\linewidth}
  \vspace{0pt}%
    \centering
    \includegraphics{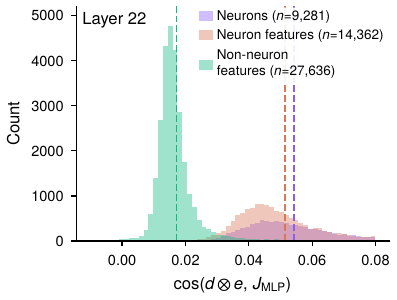}
    \captionof{figure}{Per-feature rank-1 Jacobian alignment:
      neuron features (orange) have higher Jacobian alignment
      and explain most of the faithfulness; non-neuron features (green) have lower faithfulness.}
    \label{fig:appendix-faithfulness-rank1}
  \end{minipage}
\end{figure*}

A non-neuron feature could still be a sparse combination of a few
neurons (see \S\ref{sec:clustering}), or it
could be a diffuse combination of many neurons that picks out a specific
direction. In either case, a non-neuron feature which models some
specific structure in the MLP should add something to the reconstruction
of the Jacobian. Each transcoder feature has a \emph{rank-1 Jacobian
contribution} given by the outer product of its decoder direction with
its effective encoder direction, scaled by its activation derivative at
a given token; these sum to the transcoder's Jacobian. Following
\citet{lindsey2025bulk}, we define
\emph{Jacobian alignment} between two such matrices as their Frobenius cosine
similarity.

\textbf{Jacobian alignment comes from neuron features.} Per feature, neuron-aligned features have higher Jacobian alignment
with the MLP than non-neuron features
(Figure~\ref{fig:appendix-faithfulness-rank1}, Gemma~3 4B L22). Summed over
all features, the transcoder's Jacobian alignment with the MLP is
$\approx 0.3$; restricting the sum to the neuron-aligned subset alone
matches that, while restricting to the non-aligned subset reaches only
$\approx 0.05$.
All layers behave similarly (Figure~\ref{fig:faithfulness-by-layer}): the combination of neuron features' Jacobians matches the full transcoder's Jacobian almost exactly.
Jacobian alignment scores for other models are in Appendix~\ref{app:non-neuron}.

We perform a preliminary investigation of unfaithfulness in cross-layer transcoders \citep{lindsey2025biology}, and find that, in two open source cross-layer transcoders on Gemma 2 2B and Llama 3.2 1B, the extent to which a feature uses its cross-layer decoder is negatively correlated with alignment to the closest neuron (Pearson $\rho$ from $-0.59$ to $-0.93$ across layers and models;
Appendix~\ref{app:clt-crosslayer}), with a clear cluster of neuron features with weak cross-layer decoders. While preliminary, combined with the above results, this suggests cross-layer connections are not faithful to the original model's computation.

\subsection{Cross-seed stability: most features reliably learned by transcoders are neuron features}
\label{sec:nfc:cross-seed}

\citet{paulo2025different} show that SAEs trained on different random
seeds disagree on most features. We apply the same analysis to our
transcoders: we train two transcoders that differ only in the
initialization seed, then ask, for each feature in run~A, whether there exists a feature in run~B with high (${>}\,0.7$) encoder and decoder cosine similarity. For a feature to be considered ``real'', it should reoccur between seeds.

Table~\ref{tab:cross-seed} shows the prevalence of such features among neuron and non-neuron features in transcoders trained on Gemma~3 4B L22 at $k{=}64$ and several widths $M$. We count features that are shared between seeds, neuron features, and features that are both. We report the fraction of such features in the dictionary and activations because neuron features are overrepresented in activations. We make two broad conclusions:

\textbf{Most features shared between seeds are neuron features.} Given a feature is shared between seeds, it is $>50\%$ likely to be a neuron feature for most training configurations.

\textbf{Scaling transcoder width doesn't lead to the discovery of many new stable features.} The number of neuron features stabilizes early, and the number of shared features grows sublinearly, meaning there is not much reason to expect naive scaling of transcoders to overcome their weaknesses at identifying consistent features, their main advantage over using a fixed neuron basis; even the 1M-feature dictionary only discovered 22 thousand stable features, which is comparable to the $20{,}480$ neurons found in the original model after accounting for sign.

Appendix~\ref{app:non-neuron:scaling} details the matching procedure and shows results for
JumpReLU transcoders.

\subsection{Non-neuron features barely move from initialization}
\label{sec:nfc:init-distance}

We see evidence that neuron features are updated significantly more through training, suggesting both that neuron features are more important and that non-neuron features are somewhat arbitrary and thus dependent on initialization (\Cref{sec:nfc:cross-seed}). We show two different variations of this behavior in our two training pipelines: the TopK pipeline and the
JumpReLU pipeline (training details in Appendix~\ref{app:setup}).

We train two $262$k TopK transcoders ($k{=}64$) on Gemma~3 4B L22 for $15{,}000$
steps and save the random initializations at step $0$. We also train a JumpReLU transcoder of width $262$k with the training settings outlined in \Cref{app:setup}; however, unlike other runs in that pipeline, its encoder is initialized at the same width-independent scale as the TopK pipeline, with per-row norm $\sqrt{1/3} \approx 0.577$, instead of the default $\sqrt{d_\mathrm{model}/M} \approx 0.1$.

\begin{figure*}[h]
  \centering
  \hspace*{\fill}\includegraphics[width=0.87250\linewidth,trim={80bp 457.5bp 0 0},clip]{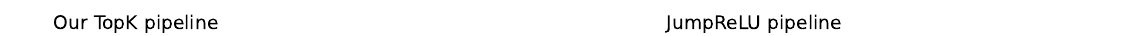}\\[-1pt]
  \includegraphics[width=\linewidth,trim={0 0 0 240bp},clip]{non-neuron/non_neuron_2x2.pdf}
  \caption{Encoder cosine
    similarity to the initialization versus firing frequency,
    split by neuron vs non-neuron features. \emph{Left:} our TopK
    pipeline ($k{=}64$, $M{=}262$k, Gemma~3 4B L22, $15{,}000$ steps).
    \emph{Right:} JumpReLU pipeline ($M{=}262$k, Gemma~3 4B L22) with
    the encoder init scale matched to the TopK pipeline; this is the
    only JumpReLU checkpoint we saved the initialization for.
    In TopK transcoders, neuron features rotate
    away from their random initialization, while non-neuron features
    stay closer to init. There is a distinct cloud of slightly-rotated features which is separate from neuron features and features that did not change at all. In the JumpReLU transcoder, despite the fact that this run reuses the width-independent initialization from the TopK pipeline, most of the alive features rotate significantly, even non-neuron features; however, features that are dead are still distributed uniformly. In both cases, the higher firing frequency of neuron features does not fully account for them moving away from initialization.}
  \label{fig:non-neuron-2x2}
\end{figure*}

\textbf{In TopK training, non-neuron features stay close to their random initialization.} \Cref{fig:non-neuron-2x2} shows that neuron features become orthogonal to the initialization through training: the median
$\cos\!\big(W_\mathrm{enc}^\mathrm{trained},\, W_\mathrm{enc}^\mathrm{init}\big)$ is close to zero in both settings.   In TopK transcoders, non-neuron features stay close to the initialization (median cosine similarity
$+0.48$); while many are very close to init (cosine similarity $>0.8$), the cosine similarity of non-neuron features to the initialization is distributed almost uniformly across the entire range.

\subsection{Neuron initialization preserves alignment through training}
\label{sec:nonneuron:ni}

The above results are partly explained by our finding that, for some training setups, non-neuron features barely move from random initialization during training (\Cref{sec:nfc:init-distance}). Therefore, the absence of neuron features for some neurons in narrower transcoders may be an artifact of the training procedure. If we instead initialize the encoder at neuron directions, training will preserve that alignment, increasing the number of neuron features. This is a technique we refer to as \emph{neuron initialization} (NI): each encoder row is set to a random sign-split $\pm(\mathrm{gate} \pm \mathrm{up})$ direction (with $10\%$ Gaussian noise added to break symmetry), the decoder is set to zero, and the transcoder is otherwise trained the same way.

NI improves on standard transcoder training in speed of convergence to the same loss and reduces the number of dead features (see Appendix~\ref{app:ninit-sweep}). More importantly, it reveals structure in the encoder the standard objective cannot learn: the fraction of features with encoder similarity~$>0.5$ to a neuron rises from $11\%$ under random initialization to $92\%$ under NI on Gemma~3 4B L22 at $256$k width, and from $12\%$ to $93\%$ on the decoder side. Together, the aligned features account for $95\%$ of activation mass under NI, up from $65\%$. %

\subsection{Case study: linebreaking}
\label{sec:nfc:linebreaking}

\begin{figure*}[t]
  \centering
  \begin{minipage}[t]{0.49\linewidth}
  \vspace{0pt}%
    \centering
    \includegraphics{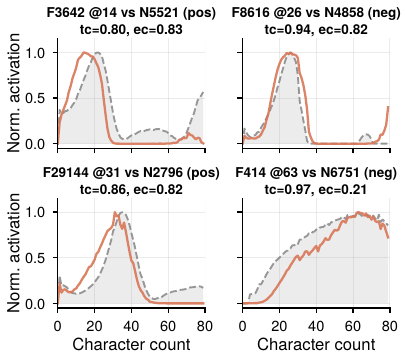}
  \captionof{figure}{Tuning curves across character count for character counting features (\citep{gurnee2025manifolds}; orange) and neurons (grey).
    \protect\footnotemark
    }
    \label{fig:linebreaking-main}
  \end{minipage}\hfill
  \begin{minipage}[t]{0.49\linewidth}
  \vspace{0pt}%
    \centering
    \includegraphics[width=\linewidth]{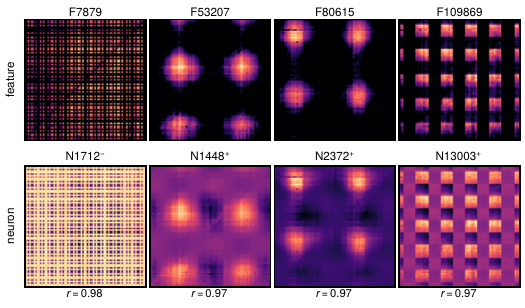}
    \captionof{figure}{Activations for different values of operands for arithmetic neurons from \citet{feucht2026arithmetic}; we discover their corresponding neuron features in a transcoder (top) and show they are causal for addition and play similar roles.
      }
    \label{fig:arith-grids-body}
  \end{minipage}
\end{figure*}

\textbf{Line-position features are mostly neuron features.} \citet{gurnee2025manifolds} identify Claude 3.5 Haiku features whose
firing rates peak at specific character positions within a line and
together let the model predict line breaks. In Gemma~3 4B with
Gemmascope L0-small, 7 of 8 clean peaked features at layer~$20$ track a
single neuron's tuning curve over character
position (Figure~\ref{fig:linebreaking-main}). Most of these are also
neuron features according to our definition, but one is similar in only
the tuning curve and not weights. See Appendix~\ref{app:linebreaking} for details.

\footnotetext{Gemma~3 4B layer~$20$, Gemmascope L0-small. For F3642 / N5521$^{+}$, F8616 / N4858$^{-}$,
    and F29144 / N2796$^{+}$ the feature and neuron curves nearly
    coincide with some offset. Counterexample: F414 / N6751$^{-}$ have a
    similar, round shape, with the feature being sharper than the
    neuron, but their weights are not aligned.}

\subsection{Case study: arithmetic}

\textbf{Transcoder features important to arithmetic are neuron features.} \citet{feucht2026arithmetic} localize base-10 addition in Llama-3.1-8B to 28
neurons of the layer-18 MLP, which write into a low-dimensional output subspace
and organize into Fourier periods. We find similar structure in three open source transcoders, with neuron features for these neurons being the majority of the features which write to the subspace, and the only ones with strong causal effects: ablating their
contributions drops addition accuracy from $94.3\%$ to $30.6\%$. These features have similar grid activation patterns to their neurons (Appendix~\ref{fig:arith-grids-body}); there also exists a weaker population of less aligned features which we consider an artifact of transcoder training. Described in detail in Appendix~\ref{app:arith}.

\subsection{Case study: color red, neuron features corresponding to multiple neurons}

This paper largely focuses on cases where neurons and their neuron features can be assumed to be monosemantic, and transcoders don't contribute anything useful through their change of basis and feature splitting. We still believe transcoders can be useful in clarifying specific detail in features and making specific neurons easier to understand; to demonstrate this, we studied a simple prompt where the input is simple and activates multiple related neurons.

In Appendix~\ref{sec:clustering:red}, we study attribution graphs for predicting the translation of the Russian word for ``red'', and find that Gemma~3 4B activates three polysemantic layer-22 neurons rather than a dedicated ``red''. Transcoder features at this site do correspond to neurons, but make the meaning cleaner by mixing together 2--4 neurons, and do discover monosemantic ``red'' features. When we consider activations of the neuron on which the transcoder feature is also active, we find that they form a clear cluster separate from other, unrelated activations.

While a more broad and in-depth study is necessary to make robust conclusions, we see this as preliminary evidence that there exists real polysemanticity that transcoders can resolve by recombining the neuron basis, especially with interventions such as those described in \Cref{sec:clustering}.

\section{Related work}
\label{sec:related}

\paragraph{Neuron-based interpretability in LLMs.}
Several lines of work analyze individual neurons in language models.
Polysemanticity has been characterized empirically across models
\citep{voita2023neurons, gurnee2024universal}, and circuits routed through
neurons have been documented for arithmetic and counting tasks
\citep{hanna2023greaterthan, nikankin2025arithmetic}. Automated descriptions
of single neurons scaled from \citet{bills2023autointerp} to model-level
catalogs \citep{transluce2024neurondesc} and head-to-head comparison with
learned features \citep{paulo2024autointerp}. Closest to our setting,
\citet{arora2026circuits} show that language-model circuits are sparse in
the neuron basis with appropriate attribution.

\paragraph{From SAEs to attribution graphs.}
Sparse autoencoders were introduced as an interpretability primitive on
language model activations \citep{bricken2023monosemanticity,
cunningham2023sparse} and subsequently scaled to frontier models
\citep{templeton2024scaling}. \citet{dunefsky2024transcoders} introduced
transcoders, which approximate an MLP layer with a sparse, wider
replacement. Attribution graph methods build on these dictionaries:
\citet{marks2025sparse} introduced linear-approximation attribution with
stop-gradient on the reconstruction error, and
\citet{ameisen2025circuit, lindsey2025biology} extended the framework to
full attribution graphs over transcoders. \citet{gurnee2025manifolds}
use transcoder feature families to investigate model behavior across
scales and report features that recur consistently across model sizes,
though they do not compare against individual neurons.

\paragraph{Comparing neurons to learned features.}
On standard interpretability evaluations, learned features score as more
interpretable than individual neurons \citep{paulo2024autointerp};
transcoders specifically score above SAEs on the same benchmarks
\citep{paulo2025transcodersbeat}. Our findings provide an account of this
gap rather than closing it. Standardized benchmark suites for sparse
dictionaries have been developed \citep{karvonen2025saebench}, and
neuron-versus-feature comparisons in the attribution graph setting appear
in \citet{arora2026circuits}.

\paragraph{Polysemantic neurons as combinations of features.}
Sparse probing \citep{gurnee2023haystack} shows that concepts can often be
read out from small numbers of neurons, motivating the view that
polysemantic neurons are combinations of underlying features.
\citet{adler2025fcc} formalize this in small Boolean networks via feature
channel coding. Post-hoc methods decompose polysemantic neurons into
clusters of activations \citep{dreyer2024pure, kopf2025prism,
liu2026neuronscope}; our proposed methods are training-time interventions that make transcoders applicable to the same problem.

\paragraph{Known artifacts of dictionary training.}
Feature splitting and absorption \citep{chanin2025absorption} do not
appear in our transcoders by default, but the training-time interventions
in Section~\ref{sec:clustering} can induce them. Cross-seed instability of
SAE and transcoder features \citep{paulo2025different} concentrates
outside the neuron-aligned subset. Our findings are also suggestive of an
account of cross-layer transcoder unfaithfulness \citep{lange2026clt},
though the supporting evidence in this paper is limited.

\section{Discussion}
\label{sec:discussion}

We have shown that processing neuron activations can produce a set of features with comparable monosemanticity to transcoder features which also explains much of the behavior of the original model. We defined and characterized a class of transcoder features --- neuron features --- and demonstrated that they are prevalent in typical activations, circuits from attribution graphs, and cases previously studied in literature. We have shown that non-neuron features don't contribute to explaining the model's local behavior, and are less likely to reappear if a transcoder is simply retrained with a different seed.

We have demonstrated similar patterns in transcoders for mixture-of-experts models (Appendix~\ref{app:moe-nf}), cross-layer transcoders (Appendix~\ref{app:clt-crosslayer}), and MLP output SAEs (Appendix~\ref{app:mlpout-saes}). Using toy models, we show that the phenomenon of replicating teacher features  imperfectly is not specific to transcoders, and show an example of neurons being represented without having an exact copy in a student narrower than the teacher (Appendix~\ref{app:distill-toy}).

Our findings cast doubt on the feasibility of resolving superposition of MLP neurons within the same layer using dictionary learning methods. While we include examples of monosemantic features discovered by transcoders not present in the original basis --- bigrams (Appendix~\ref{app:eval-results:bigram}) and color (Appendix~\ref{sec:clustering:red}) -- we are uncertain about the extent to which these features are truly represented by the original model. We remain cautiously optimistic about resolving cross-layer superposition in neuron circuits using supernode graphs, and believe further case studies of model behavior with these tools can help solidify our understanding of abstractions used by the model.

We make the following recommendations to researchers:
\begin{enumerate}
    \item Avoiding the use of transcoders with a width expansion factor of less than $4\times$; for MoE models, this practically means not using transcoders at all (\Cref{app:moe-nf}).
    \item When comparing dictionary learning methods to neuron baselines, thresholding the neurons with TopK of around 64, and considering positive and negative activations of a given neuron as distinct semantic units; when training dictionary learning methods, considering a neuron-inspired initialization (\Cref{sec:nonneuron:ni}).
    \item When evaluating dictionary learning methods, studying more closely the elements which reproduce across seeds and align with the computation of the original module.
\end{enumerate}

\section{Limitations}
\label{sec:limitations}
Our results are restricted to base models, and we mostly use English-language data. We evaluate \Nmodels{} transcoders on models in the 1B--27B parameter range, but they are all dense, and we only use per-layer transcoders. Our preliminary investigations have shown MoE architectures may have interesting cases of same-layer superposition and that cross-layer transcoders' faithfulness may be hampered by existing training schemes; these results are not developed further. Given that most open source frontier models are MoEs, developing techniques that work on them is necessary to understand more complex capabilities.

Throughout the paper, we characterize only sparse, high-magnitude activations of neurons. Even though we believe from informal investigations that low activations and multi-token patterns can be just as or even more interpretable than the sparse activations we considered so far, it is beyond the scope of this paper. We consider interpretability of feature activations on single tokens only and do not consider possible mechanisms which can read activations spread across multiple tokens.

We study only Top-$k$ and JumpReLU dictionaries trained with two different codebases. We have shown that the fraction of features categorized as neuron features depends on dictionary architecture and on the threshold used for selection ($0.6$, $0.7$ or $0.9$ depending on transcoder and objective); we do not offer a comprehensive explanation of how architectural factors decide the appearance of neuron features. We do not quantify the role of features that are almost neuron features which are more prevalent in TopK transcoders and some of which seem to correspond to sparse combinations of neurons.

We do not characterise features that recur across seeds in JumpReLU transcoders without being neuron features, even though there is a substantial amount present. We do not investigate features which approximately co-occur between seeds or attempt to understand if such features lie on continuous models.

Further, our analyses on cross-layer transcoders use only two narrow models and do not make a conclusion as to whether the non-neuron features are faithful and whether our other results generalize to them. Relatedly, for faithfulness evaluations, we only consider first-order, Jacobian faithfulness (as opposed to faithfulness under interventions), and do not extend our analysis to non-JumpReLU transcoders. We do not study MLP output or residual SAEs and evaluate the connection of their features to neurons, even though MLP output SAEs may be a more promising primitive for attribution graph construction and learn fewer neuron features at lower cost.

We explain that the weight-based definition of neuron features is not robust due to the fact that non-neuron features retain parts of their direction at initialization. However, it is not clear if the learned transcoders are learning faithful decompositions of the computation and what their decoders correspond to in general. We do not characterize the behavior of these interventions across models or layers. Our training interventions do not make clusters more prevalent or distinguishable, and baseline transcoders appear to have the structure we are looking for, even if the clustering is less strict and the weight alignment to neurons is lower.

We also did not rigorously test whether it is possible to remove neuron features from transcoders; we believe features must be aligned to neurons to some extent ($>0.4$), but do not show detailed results for what degree of alignment is necessary for the transcoding objective.

We hypothesize that polysemantic neuron activations can be defined as clusters defined by small sets of neurons. We show that dense K-means surprisingly finds such clusters, but do not find a connection to features learned by transcoders. We can only make broad claims about co-firing of features and their closest neurons and do not characterize precise ways feature activation is determined by sets of neurons firing together. We also do not formulate ways to understand diffuse combinations of dozens or hundreds of neurons. We are uncertain about the implications of these results for the framework of computation in superposition and emphasize understanding superposition across layers and in MoE experts as important areas for future work.

We do not fully explain how neuron features and transcoder features co-occur, which neuron features are most important and why they are represented monosemantically. We do not characterise where transcoder decoders come from: the directions they pick out appear meaningful but are not captured by the clustering picture. Furthermore, we do not evaluate the degree to which concepts are represented in transcoders or show ways to find specific concepts that transcoders represent with combinations of neurons; sparse probing and its variations, as we have shown, are not the appropriate
evaluation for this.

We run no systematic human evaluation; all our quantitative interpretability evaluations use automated metrics. This means some of the findings, like neurons finding more relevant features for sparse probing, may find features that are merely correlated with the desired variable and represent a different concept in extremal activations.

\section*{Potential Risks}
The procedure described in Appendix~\ref{app:distress-setup} gives a simple
and inexpensive way to find a small set of neurons that activate on
patterns plausibly associated with model distress in any open-weights
model that exhibits them. Though we do not release specifics of the weight editing technique, this type of weight editing, if applied irresponsibly, could remove a
behavioral signal which, regardless of beliefs about machine
sentience, is currently among the more legible windows into whether a
model is in a difficult internal state. However, a sparse neuron-based weight edit would not be any more effective than orthogonalization \citep{arditi2024refusal}, meaning the only potentially harmful consequence is detailing the procedure for identifying the steering vector.

\label{main-body-end}

\begin{ack}
This work was supported by the Anthropic Fellows Program, a BlueDot rapid grant, and compute from White Circle. We thank Ben Thompson, Gavia Gray, Curt Tigges, Thomas Jiralerspong, McNair Shah, and Parv Mahajan.
\end{ack}

\bibliographystyle{plainnat}
\bibliography{refs}

\begin{thebibliography}{61}
\providecommand{\natexlab}[1]{#1}
\providecommand{\url}[1]{\texttt{#1}}
\expandafter\ifx\csname urlstyle\endcsname\relax
  \providecommand{\doi}[1]{doi: #1}\else
  \providecommand{\doi}{doi: \begingroup \urlstyle{rm}\Url}\fi

\bibitem[Adler et~al.(2025)Adler, Alistarh, and Shavit]{adler2025fcc}
Micah Adler, Dan Alistarh, and Nir Shavit.
\newblock Towards combinatorial interpretability of neural computation.
\newblock \emph{arXiv preprint arXiv:2504.08842}, 2025.

\bibitem[Ahn et~al.(2025)Ahn, Xu, Abreu, Fan, Magakyan, Sharma, Zhan, and
  Langford]{ahn2025dion}
Kwangjun Ahn, Byron Xu, Natalie Abreu, Ying Fan, Gagik Magakyan, Pratyusha
  Sharma, Zheng Zhan, and John Langford.
\newblock Dion: Distributed orthonormalized updates.
\newblock \emph{arXiv preprint arXiv:2504.05295}, 2025.
\newblock URL \url{https://arxiv.org/abs/2504.05295}.

\bibitem[Ameisen et~al.(2025)Ameisen, Lindsey, Pearce, Gurnee, Turner, Chen,
  Citro, Abrahams, Carter, Hosmer, Marcus, Sklar, Templeton, Bricken,
  McDougall, Cunningham, Henighan, Jermyn, Jones, Persic, Qi, Thompson,
  Zimmerman, Rivoire, Conerly, Olah, and Batson]{ameisen2025circuit}
Emmanuel Ameisen, Jack Lindsey, Adam Pearce, Wes Gurnee, Nicholas~L. Turner,
  Brian Chen, Craig Citro, David Abrahams, Shan Carter, Basil Hosmer, Jonathan
  Marcus, Michael Sklar, Adly Templeton, Trenton Bricken, Callum McDougall,
  Hoagy Cunningham, Thomas Henighan, Adam Jermyn, Andy Jones, Andrew Persic,
  Zhenyi Qi, T.~Ben Thompson, Sam Zimmerman, Kelley Rivoire, Thomas Conerly,
  Chris Olah, and Joshua Batson.
\newblock Circuit tracing: Revealing computational graphs in language models.
\newblock \emph{Transformer Circuits Thread}, 2025.
\newblock URL
  \url{https://transformer-circuits.pub/2025/attribution-graphs/methods.html}.

\bibitem[Arditi et~al.(2024)Arditi, Obeso, Syed, Paleka, Panickssery, Gurnee,
  and Nanda]{arditi2024refusal}
Andy Arditi, Oscar Obeso, Aaquib Syed, Daniel Paleka, Nina Panickssery, Wes
  Gurnee, and Neel Nanda.
\newblock Refusal in language models is mediated by a single direction.
\newblock In \emph{Advances in Neural Information Processing Systems}, 2024.

\bibitem[Arora et~al.(2026)Arora, Wu, Steinhardt, and
  Schwettmann]{arora2026circuits}
Aryaman Arora, Zhengxuan Wu, Jacob Steinhardt, and Sarah Schwettmann.
\newblock Language model circuits are sparse in the neuron basis.
\newblock \emph{arXiv preprint arXiv:2601.22594}, 2026.
\newblock \url{https://transluce.org/neuron-circuits}.

\bibitem[Austen(1813)]{austen1813pride}
Jane Austen.
\newblock \emph{Pride and Prejudice}.
\newblock T. Egerton, London, 1813.

\bibitem[Ben~Allal et~al.(2025)Ben~Allal, Lozhkov, Bakouch,
  Mart{\'\i}n~Bl{\'a}zquez, Penedo, Tunstall, Marafioti, Kydl{\'\i}{\v c}ek,
  Piqueres~Lajar{\'\i}n, Srivastav, Lochner, Fahlgren, Nguyen, Fourrier,
  Burtenshaw, Larcher, Zhao, Zakka, Morlon, Raffel, von Werra, and
  Wolf]{allal2025smollm2}
Loubna Ben~Allal, Anton Lozhkov, Elie Bakouch, Gabriel
  Mart{\'\i}n~Bl{\'a}zquez, Guilherme Penedo, Lewis Tunstall, Andr{\'e}s
  Marafioti, Hynek Kydl{\'\i}{\v c}ek, Agust{\'\i}n Piqueres~Lajar{\'\i}n,
  Vaibhav Srivastav, Joshua Lochner, Caleb Fahlgren, Xuan-Son Nguyen,
  Cl{\'e}mentine Fourrier, Ben Burtenshaw, Hugo Larcher, Haojun Zhao, Cyril
  Zakka, Mathieu Morlon, Colin Raffel, Leandro von Werra, and Thomas Wolf.
\newblock {SmolLM2}: When smol goes big -- data-centric training of a small
  language model.
\newblock \emph{arXiv preprint arXiv:2502.02737}, 2025.
\newblock URL \url{https://arxiv.org/abs/2502.02737}.

\bibitem[Bills et~al.(2023)Bills, Cammarata, Mossing, Tillman, Gao, Goh,
  Sutskever, Leike, Wu, and Saunders]{bills2023autointerp}
Steven Bills, Nick Cammarata, Dan Mossing, Henk Tillman, Leo Gao, Gabriel Goh,
  Ilya Sutskever, Jan Leike, Jeff Wu, and William Saunders.
\newblock Language models can explain neurons in language models.
\newblock OpenAI Blog, 2023.
\newblock
  \url{https://openaipublic.blob.core.windows.net/neuron-explainer/paper/index.html}.

\bibitem[Bricken et~al.(2023)Bricken, Templeton, Batson, Chen, Jermyn, Conerly,
  Turner, Anil, Denison, Askell, et~al.]{bricken2023monosemanticity}
Trenton Bricken, Adly Templeton, Joshua Batson, Brian Chen, Adam Jermyn, Tom
  Conerly, Nick Turner, Cem Anil, Carson Denison, Amanda Askell, et~al.
\newblock Towards monosemanticity: Decomposing language models with dictionary
  learning.
\newblock \emph{Transformer Circuits Thread}, 2023.
\newblock
  \url{https://transformer-circuits.pub/2023/monosemantic-features/index.html}.

\bibitem[Bussmann et~al.(2025)Bussmann, Nabeshima, Karvonen, and
  Nanda]{bussmann2025matryoshka}
Bart Bussmann, Noa Nabeshima, Adam Karvonen, and Neel Nanda.
\newblock Learning multi-level features with matryoshka sparse autoencoders.
\newblock \emph{arXiv preprint arXiv:2503.17547}, 2025.

\bibitem[Chanin et~al.(2025)Chanin, Wilken-Smith, Dulka, Bhatnagar, Golechha,
  and Bloom]{chanin2025absorption}
David Chanin, James Wilken-Smith, Tom{\'a}{\v{s}} Dulka, Hardik Bhatnagar,
  Satvik Golechha, and Joseph Bloom.
\newblock {A is for Absorption}: Studying feature splitting and absorption in
  sparse autoencoders.
\newblock In \emph{Advances in Neural Information Processing Systems}, 2025.
\newblock arXiv:2409.14507.

\bibitem[Choi et~al.(2024)Choi, Huang, Meng, Johnson, Steinhardt, and
  Schwettmann]{transluce2024neurondesc}
Dami Choi, Vincent Huang, Kevin Meng, Daniel~D. Johnson, Jacob Steinhardt, and
  Sarah Schwettmann.
\newblock Scaling automatic neuron description.
\newblock Transluce Blog, 2024.
\newblock \url{https://transluce.org/neuron-descriptions}.

\bibitem[Conerly et~al.(2025)Conerly, Cunningham, Templeton, Lindsey, Hosmer,
  and Jermyn]{conerly2025january}
Tom Conerly, Hoagy Cunningham, Adly Templeton, Jack Lindsey, Basil Hosmer, and
  Adam Jermyn.
\newblock Circuits updates --- january 2025.
\newblock \emph{Transformer Circuits Thread}, 2025.
\newblock URL
  \url{https://transformer-circuits.pub/2025/january-update/index.html}.

\bibitem[Cunningham et~al.(2024)Cunningham, Ewart, Riggs, Huben, and
  Sharkey]{cunningham2023sparse}
Hoagy Cunningham, Aidan Ewart, Logan Riggs, Robert Huben, and Lee Sharkey.
\newblock Sparse autoencoders find highly interpretable features in language
  models.
\newblock In \emph{International Conference on Learning Representations}, 2024.
\newblock arXiv:2309.08600.

\bibitem[Dreyer et~al.(2024)Dreyer, Purelku, Vielhaben, Samek, and
  Lapuschkin]{dreyer2024pure}
Maximilian Dreyer, Erblina Purelku, Johanna Vielhaben, Wojciech Samek, and
  Sebastian Lapuschkin.
\newblock {PURE}: Turning polysemantic neurons into pure features by
  identifying relevant circuits.
\newblock In \emph{CVPR Workshop on Explainable AI for Computer Vision
  (XAI4CV)}, 2024.
\newblock arXiv:2404.06453.

\bibitem[Dunefsky et~al.(2024)Dunefsky, Chlenski, and
  Nanda]{dunefsky2024transcoders}
Jacob Dunefsky, Philippe Chlenski, and Neel Nanda.
\newblock Transcoders find interpretable {LLM} feature circuits.
\newblock In \emph{Advances in Neural Information Processing Systems}, 2024.
\newblock arXiv:2406.11944.

\bibitem[Elhage et~al.(2022)Elhage, Hume, Olsson, Schiefer, Henighan, Kravec,
  Hatfield-Dodds, Lasenby, Drain, Chen, et~al.]{elhage2022superposition}
Nelson Elhage, Tristan Hume, Catherine Olsson, Nicholas Schiefer, Tom Henighan,
  Shauna Kravec, Zac Hatfield-Dodds, Robert Lasenby, Dawn Drain, Carol Chen,
  et~al.
\newblock Toy models of superposition.
\newblock \emph{Transformer Circuits Thread}, 2022.
\newblock \url{https://transformer-circuits.pub/2022/toy_model/index.html}.

\bibitem[Feucht et~al.(2026)Feucht, Haklay, Bhalla, Wurgaft, Rager, Sarfati,
  Merullo, McGrath, Lewis, Lubana, Fel, and Geiger]{feucht2026arithmetic}
Sheridan Feucht, Tal Haklay, Usha Bhalla, Daniel Wurgaft, Can Rager,
  Rapha{\"e}l Sarfati, Jack Merullo, Tom McGrath, Owen Lewis, Ekdeep~Singh
  Lubana, Thomas Fel, and Atticus Geiger.
\newblock Arithmetic in the wild: Llama uses base-10 addition to reason about
  cyclic concepts.
\newblock \emph{arXiv preprint arXiv:2605.01148}, 2026.

\bibitem[Gao et~al.(2020)Gao, Biderman, Black, Golding, Hoppe, Foster, Phang,
  He, Thite, Nabeshima, Presser, and Leahy]{gao2020pile}
Leo Gao, Stella Biderman, Sid Black, Laurence Golding, Travis Hoppe, Charles
  Foster, Jason Phang, Horace He, Anish Thite, Noa Nabeshima, Shawn Presser,
  and Connor Leahy.
\newblock The {Pile}: An 800{GB} dataset of diverse text for language modeling.
\newblock \emph{arXiv preprint arXiv:2101.00027}, 2020.

\bibitem[{Gemma Team}(2024)]{gemma2techreport}
{Gemma Team}.
\newblock Gemma 2: Improving open language models at a practical size.
\newblock \emph{arXiv preprint arXiv:2408.00118}, 2024.

\bibitem[{Gemma Team}(2025)]{gemma3techreport}
{Gemma Team}.
\newblock Gemma 3 technical report.
\newblock \emph{arXiv preprint arXiv:2503.19786}, 2025.

\bibitem[{Gemma Team}(2026)]{gemma4techreport}
{Gemma Team}.
\newblock Gemma 4 technical report.
\newblock \emph{arXiv preprint arXiv:2607.02770}, 2026.

\bibitem[Gurnee et~al.(2023)Gurnee, Nanda, Pauly, Harvey, Troitskii, and
  Bertsimas]{gurnee2023haystack}
Wes Gurnee, Neel Nanda, Matthew Pauly, Katherine Harvey, Dmitrii Troitskii, and
  Dimitris Bertsimas.
\newblock Finding neurons in a haystack: Case studies with sparse probing.
\newblock \emph{Transactions on Machine Learning Research}, 2023.
\newblock arXiv:2305.01610.

\bibitem[Gurnee et~al.(2024)Gurnee, Horsley, Guo, Kheirkhah, Sun, Hathaway,
  Nanda, and Bertsimas]{gurnee2024universal}
Wes Gurnee, Theo Horsley, Zifan~Carl Guo, Tara~Rezaei Kheirkhah, Qinyi Sun,
  Will Hathaway, Neel Nanda, and Dimitris Bertsimas.
\newblock Universal neurons in {GPT-2} language models.
\newblock \emph{Transactions on Machine Learning Research}, 2024.
\newblock arXiv:2401.12181.

\bibitem[Gurnee et~al.(2025)Gurnee, Ameisen, Kauvar, Tarng, Pearce, Olah, and
  Batson]{gurnee2025manifolds}
Wes Gurnee, Emmanuel Ameisen, Isaac Kauvar, Julius Tarng, Adam Pearce, Chris
  Olah, and Joshua Batson.
\newblock When models manipulate manifolds: The geometry of a counting task.
\newblock \emph{Transformer Circuits Thread}, 2025.
\newblock arXiv:2601.04480;
  \url{https://transformer-circuits.pub/2025/linebreaks/index.html}.

\bibitem[Hanna et~al.(2023)Hanna, Liu, and Variengien]{hanna2023greaterthan}
Michael Hanna, Ollie Liu, and Alexandre Variengien.
\newblock How does {GPT-2} compute greater-than?: Interpreting mathematical
  abilities in a pre-trained language model.
\newblock In \emph{Advances in Neural Information Processing Systems}, 2023.
\newblock arXiv:2305.00586.

\bibitem[Hanna et~al.(2025)Hanna, Piotrowski, Lindsey, and
  Ameisen]{circuittracer2025}
Michael Hanna, Mateusz Piotrowski, Jack Lindsey, and Emmanuel Ameisen.
\newblock circuit-tracer.
\newblock \url{https://github.com/decoderesearch/circuit-tracer}, 2025.
\newblock The first two authors contributed equally and are listed
  alphabetically.

\bibitem[He et~al.(2024)He, Shu, Ge, Chen, Wang, Zhou, Liu, Guo, Huang, Wu,
  Jiang, and Qiu]{he2024llamascope}
Zhengfu He, Wentao Shu, Xuyang Ge, Lingjie Chen, Junxuan Wang, Yunhua Zhou,
  Frances Liu, Qipeng Guo, Xuanjing Huang, Zuxuan Wu, Yu-Gang Jiang, and Xipeng
  Qiu.
\newblock {Llama Scope}: Extracting millions of features from {Llama-3.1-8B}
  with sparse autoencoders.
\newblock \emph{arXiv preprint arXiv:2410.20526}, 2024.
\newblock URL \url{https://arxiv.org/abs/2410.20526}.

\bibitem[Hendrycks and Gimpel(2016)]{hendrycks2016gelu}
Dan Hendrycks and Kevin Gimpel.
\newblock Gaussian error linear units ({GELUs}).
\newblock \emph{arXiv preprint arXiv:1606.08415}, 2016.
\newblock URL \url{https://arxiv.org/abs/1606.08415}.

\bibitem[Jafari et~al.(2025)Jafari, Eberle, Khakzar, and Nanda]{jafari2025relp}
Farnoush~Rezaei Jafari, Oliver Eberle, Ashkan Khakzar, and Neel Nanda.
\newblock {RelP}: Faithful and efficient circuit discovery in language models
  via relevance patching.
\newblock \emph{arXiv preprint arXiv:2508.21258}, 2025.
\newblock URL \url{https://arxiv.org/abs/2508.21258}.

\bibitem[Kantamneni et~al.(2025)Kantamneni, Engels, Rajamanoharan, Tegmark, and
  Nanda]{kantamneni2025sparseprobing}
Subhash Kantamneni, Joshua Engels, Senthooran Rajamanoharan, Max Tegmark, and
  Neel Nanda.
\newblock Are sparse autoencoders useful? a case study in sparse probing.
\newblock In \emph{International Conference on Machine Learning}, 2025.
\newblock arXiv:2502.16681.

\bibitem[Karvonen(2025)]{karvonen2025e2e}
Adam Karvonen.
\newblock Revisiting end-to-end sparse autoencoder training: A short finetune
  is all you need.
\newblock \emph{arXiv preprint arXiv:2503.17272}, 2025.
\newblock URL \url{https://arxiv.org/abs/2503.17272}.

\bibitem[Karvonen et~al.(2025)Karvonen, Rager, Lin, Tigges, Bloom, Chanin, Lau,
  Farrell, McDougall, Ayonrinde, Till, Wearden, Conmy, Marks, and
  Nanda]{karvonen2025saebench}
Adam Karvonen, Can Rager, Johnny Lin, Curt Tigges, Joseph Bloom, David Chanin,
  Yeu-Tong Lau, Eoin Farrell, Callum McDougall, Kola Ayonrinde, Demian Till,
  Matthew Wearden, Arthur Conmy, Samuel Marks, and Neel Nanda.
\newblock {SAEBench}: A comprehensive benchmark for sparse autoencoders in
  language model interpretability.
\newblock In \emph{International Conference on Machine Learning}, 2025.
\newblock arXiv:2503.09532.

\bibitem[Kopf et~al.(2025)Kopf, Feldhus, Bykov, Bommer, Hedstr{\"o}m,
  H{\"o}hne, and Eberle]{kopf2025prism}
Laura Kopf, Nils Feldhus, Kirill Bykov, Philine~Lou Bommer, Anna Hedstr{\"o}m,
  Marina M.-C. H{\"o}hne, and Oliver Eberle.
\newblock Capturing polysemanticity with {PRISM}: A multi-concept feature
  description framework.
\newblock In \emph{Advances in Neural Information Processing Systems}, 2025.
\newblock arXiv:2506.15538.

\bibitem[Lange et~al.(2026)Lange, {RGRGRG}, Dearstyne, and Maher]{lange2026clt}
Georg Lange, {RGRGRG}, Kat Dearstyne, and Kamal Maher.
\newblock Cross-layer transcoders are incentivized to learn unfaithful
  circuits.
\newblock LessWrong, February 2026.
\newblock \url{https://www.lesswrong.com/posts/6CS2NDmoLCFcEJMor}.

\bibitem[Lindsey et~al.(2024)Lindsey, Cunningham, Conerly, Templeton, and
  Persic]{anthropicaug2024}
Jack Lindsey, Hoagy Cunningham, Tom Conerly, Adly Templeton, and Andrew Persic.
\newblock Circuits updates --- august 2024.
\newblock Transformer Circuits Thread, 2024.
\newblock \url{https://transformer-circuits.pub/2024/august-update/index.html}.

\bibitem[Lindsey et~al.(2025{\natexlab{a}})Lindsey, Chen, Pearce, Hydrie, and
  Conerly]{lindsey2025bulk}
Jack Lindsey, Brian Chen, Adam Pearce, Sasha Hydrie, and Thomas Conerly.
\newblock Sparse mixtures of linear transforms.
\newblock \emph{Transformer Circuits Thread}, 2025{\natexlab{a}}.
\newblock \url{https://transformer-circuits.pub/2025/bulk-update/index.html}.

\bibitem[Lindsey et~al.(2025{\natexlab{b}})Lindsey, Gurnee, Ameisen, Chen,
  Pearce, Turner, Citro, Abrahams, Carter, Hosmer, Marcus, Sklar, Templeton,
  Bricken, McDougall, Cunningham, Henighan, Jermyn, Jones, Persic, Qi,
  Thompson, Zimmerman, Rivoire, Conerly, Olah, and Batson]{lindsey2025biology}
Jack Lindsey, Wes Gurnee, Emmanuel Ameisen, Brian Chen, Adam Pearce,
  Nicholas~L. Turner, Craig Citro, David Abrahams, Shan Carter, Basil Hosmer,
  Jonathan Marcus, Michael Sklar, Adly Templeton, Trenton Bricken, Callum
  McDougall, Hoagy Cunningham, Thomas Henighan, Adam Jermyn, Andy Jones, Andrew
  Persic, Zhenyi Qi, T.~Ben Thompson, Sam Zimmerman, Kelley Rivoire, Thomas
  Conerly, Chris Olah, and Joshua Batson.
\newblock On the biology of a large language model.
\newblock \emph{Transformer Circuits Thread}, 2025{\natexlab{b}}.
\newblock URL
  \url{https://transformer-circuits.pub/2025/attribution-graphs/biology.html}.

\bibitem[Liu et~al.(2025)Liu, Su, Yao, Jiang, Lai, Du, Qin, Xu, Lu, Yan,
  et~al.]{liu2025muon}
Jingyuan Liu, Jianlin Su, Xingcheng Yao, Zhejun Jiang, Guokun Lai, Yulun Du,
  Yidao Qin, Weixin Xu, Enzhe Lu, Junjie Yan, et~al.
\newblock Muon is scalable for {LLM} training.
\newblock \emph{arXiv preprint arXiv:2502.16982}, 2025.
\newblock URL \url{https://arxiv.org/abs/2502.16982}.

\bibitem[Liu et~al.(2026)Liu, Miao, Zhao, Liu, and Du]{liu2026neuronscope}
Weiqi Liu, Yongliang Miao, Haiyan Zhao, Yanguang Liu, and Mengnan Du.
\newblock {NeuronScope}: A multi-agent framework for explaining polysemantic
  neurons in language models.
\newblock \emph{arXiv preprint arXiv:2601.03671}, 2026.

\bibitem[{Llama Team}(2024)]{llama3herd}
{Llama Team}.
\newblock The {Llama} 3 herd of models.
\newblock \emph{arXiv preprint arXiv:2407.21783}, 2024.

\bibitem[Macar et~al.(2026)Macar, Yang, Wang, Wallich, Ameisen, and
  Lindsey]{macar2026introspection}
Uzay Macar, Li~Yang, Atticus Wang, Peter Wallich, Emmanuel Ameisen, and Jack
  Lindsey.
\newblock Mechanisms of introspective awareness.
\newblock \emph{arXiv preprint arXiv:2603.21396}, 2026.
\newblock URL \url{https://arxiv.org/abs/2603.21396}.

\bibitem[Marks et~al.(2025)Marks, Rager, Michaud, Belinkov, Bau, and
  Mueller]{marks2025sparse}
Samuel Marks, Can Rager, Eric~J. Michaud, Yonatan Belinkov, David Bau, and
  Aaron Mueller.
\newblock Sparse feature circuits: Discovering and editing interpretable causal
  graphs in language models.
\newblock In \emph{International Conference on Learning Representations}, 2025.
\newblock arXiv:2403.19647.

\bibitem[McDougall et~al.(2025)McDougall, Conmy, Kram{\'a}r, Lieberum,
  Rajamanoharan, and Nanda]{mcdougall2025gemmascope2}
Callum McDougall, Arthur Conmy, J{\'a}nos Kram{\'a}r, Tom Lieberum, Senthooran
  Rajamanoharan, and Neel Nanda.
\newblock {Gemma Scope 2} -- technical paper.
\newblock Technical report, Google {DeepMind}, September 2025.
\newblock URL
  \url{https://storage.googleapis.com/deepmind-media/DeepMind.com/Blog/gemma-scope-2-helping-the-ai-safety-community-deepen-understanding-of-complex-language-model-behavior/Gemma_Scope_2_Technical_Paper.pdf}.

\bibitem[Nikankin et~al.(2025)Nikankin, Reusch, Mueller, and
  Belinkov]{nikankin2025arithmetic}
Yaniv Nikankin, Anja Reusch, Aaron Mueller, and Yonatan Belinkov.
\newblock Arithmetic without algorithms: Language models solve math with a bag
  of heuristics.
\newblock In \emph{International Conference on Learning Representations}, 2025.
\newblock arXiv:2410.21272.

\bibitem[Olah et~al.(2025)Olah, Turner, and Conerly]{olah2025interference}
Chris Olah, Nicholas~L. Turner, and Tom Conerly.
\newblock A toy model of interference weights.
\newblock \emph{Transformer Circuits Thread}, 2025.
\newblock URL
  \url{https://transformer-circuits.pub/2025/interference-weights/index.html}.

\bibitem[Panickssery et~al.(2024)Panickssery, Gabrieli, Schulz, Tong, Hubinger,
  and Turner]{panickssery2024caa}
Nina Panickssery, Nick Gabrieli, Julian Schulz, Meg Tong, Evan Hubinger, and
  Alexander~Matt Turner.
\newblock Steering {Llama 2} via contrastive activation addition.
\newblock \emph{arXiv preprint arXiv:2312.06681}, 2024.
\newblock URL \url{https://arxiv.org/abs/2312.06681}.

\bibitem[Paulo and Belrose(2026)]{paulo2025different}
Gon{\c c}alo Paulo and Nora Belrose.
\newblock Sparse autoencoders trained on the same data learn different
  features.
\newblock In \emph{International Conference on Learning Representations}, 2026.

\bibitem[Paulo et~al.(2025{\natexlab{a}})Paulo, Mallen, Juang, and
  Belrose]{paulo2024autointerp}
Gon{\c c}alo Paulo, Alex Mallen, Caden Juang, and Nora Belrose.
\newblock Automatically interpreting millions of features in large language
  models.
\newblock In \emph{International Conference on Machine Learning},
  2025{\natexlab{a}}.
\newblock arXiv:2410.13928.

\bibitem[Paulo et~al.(2025{\natexlab{b}})Paulo, Shabalin, and
  Belrose]{paulo2025transcodersbeat}
Gon{\c c}alo Paulo, Stepan Shabalin, and Nora Belrose.
\newblock Transcoders beat sparse autoencoders for interpretability.
\newblock \emph{arXiv preprint arXiv:2501.18823}, 2025{\natexlab{b}}.

\bibitem[Penedo et~al.(2024)Penedo, Kydl{\'\i}{\v c}ek, Ben~allal, Lozhkov,
  Mitchell, Raffel, Von~Werra, and Wolf]{penedo2024fineweb}
Guilherme Penedo, Hynek Kydl{\'\i}{\v c}ek, Loubna Ben~allal, Anton Lozhkov,
  Margaret Mitchell, Colin Raffel, Leandro Von~Werra, and Thomas Wolf.
\newblock The {FineWeb} datasets: Decanting the web for the finest text data at
  scale.
\newblock \emph{arXiv preprint arXiv:2406.17557}, 2024.

\bibitem[{Qwen Team}(2025)]{qwen3techreport}
{Qwen Team}.
\newblock {Qwen3} technical report.
\newblock \emph{arXiv preprint arXiv:2505.09388}, 2025.

\bibitem[{Qwen Team}(2026{\natexlab{a}})]{qwen3.5}
{Qwen Team}.
\newblock {Qwen3.5}: Towards native multimodal agents, February
  2026{\natexlab{a}}.
\newblock URL \url{https://qwen.ai/blog?id=qwen3.5}.

\bibitem[{Qwen Team}(2026{\natexlab{b}})]{qwen_qwen3_coder_next_tech_report}
{Qwen Team}.
\newblock Qwen3-coder-next technical report.
\newblock Technical report, Alibaba Group, 2026{\natexlab{b}}.
\newblock URL
  \url{https://github.com/QwenLM/Qwen3-Coder/blob/main/qwen3_coder_next_tech_report.pdf}.
\newblock Accessed: 2026-02-03.

\bibitem[Rajamanoharan et~al.(2024)Rajamanoharan, Lieberum, Sonnerat, Conmy,
  Varma, Kram{\'a}r, and Nanda]{rajamanoharan2024jumprelu}
Senthooran Rajamanoharan, Tom Lieberum, Nicolas Sonnerat, Arthur Conmy, Vikrant
  Varma, J{\'a}nos Kram{\'a}r, and Neel Nanda.
\newblock Jumping ahead: Improving reconstruction fidelity with {JumpReLU}
  sparse autoencoders.
\newblock \emph{arXiv preprint arXiv:2407.14435}, 2024.

\bibitem[Shazeer(2020)]{shazeer2020glu}
Noam Shazeer.
\newblock {GLU} variants improve transformer.
\newblock \emph{arXiv preprint arXiv:2002.05202}, 2020.

\bibitem[Soligo et~al.(2026)Soligo, Mikulik, and Saunders]{soligo2026gemma}
Anna Soligo, Vladimir Mikulik, and William Saunders.
\newblock {Gemma Needs Help}: Investigating and mitigating emotional
  instability in {LLMs}.
\newblock \emph{arXiv preprint arXiv:2603.10011}, 2026.
\newblock URL \url{https://arxiv.org/abs/2603.10011}.

\bibitem[Templeton et~al.(2024)Templeton, Conerly, Marcus, Lindsey, Bricken,
  Chen, Pearce, Citro, Ameisen, Jones, et~al.]{templeton2024scaling}
Adly Templeton, Tom Conerly, Jonathan Marcus, Jack Lindsey, Trenton Bricken,
  Brian Chen, Adam Pearce, Craig Citro, Emmanuel Ameisen, Andy Jones, et~al.
\newblock Scaling monosemanticity: Extracting interpretable features from
  {Claude} 3 {Sonnet}.
\newblock \emph{Transformer Circuits Thread}, 2024.
\newblock
  \url{https://transformer-circuits.pub/2024/scaling-monosemanticity/index.html}.

\bibitem[Voita et~al.(2024)Voita, Ferrando, and Nalmpantis]{voita2023neurons}
Elena Voita, Javier Ferrando, and Christoforos Nalmpantis.
\newblock Neurons in large language models: Dead, n-gram, positional.
\newblock In \emph{Findings of the Association for Computational Linguistics:
  {ACL} 2024}, 2024.
\newblock arXiv:2309.04827.

\bibitem[Ye et~al.(2025)Ye, Suzuki, Inaba, and Kuribayashi]{ye2025ffkvsae}
Mengyu Ye, Jun Suzuki, Tatsuro Inaba, and Tatsuki Kuribayashi.
\newblock Transformer key-value memories are nearly as interpretable as sparse
  autoencoders.
\newblock In \emph{Advances in Neural Information Processing Systems}, 2025.
\newblock URL \url{https://openreview.net/forum?id=2TKEGTfQBd}.

\bibitem[Zheng et~al.(2023)Zheng, Chiang, Sheng, Li, Zhuang, Wu, Zhuang, Li,
  Lin, Xing, Gonzalez, Stoica, and Zhang]{zheng2023lmsyschat1m}
Lianmin Zheng, Wei-Lin Chiang, Ying Sheng, Tianle Li, Siyuan Zhuang, Zhanghao
  Wu, Yonghao Zhuang, Zhuohan Li, Zi~Lin, Eric~P. Xing, Joseph~E. Gonzalez, Ion
  Stoica, and Hao Zhang.
\newblock {LMSYS-Chat-1M}: A large-scale real-world {LLM} conversation dataset,
  2023.

\end{thebibliography}

\appendix
\begingroup

\section*{Appendix: table of contents}
\begingroup
\small
\setlength{\tabcolsep}{4pt}
\begin{longtable}{@{}l p{0.75\linewidth}@{}}
  \toprule
  & Contents \\
  \midrule
  \endhead
  \Cref{app:setup} & \nameref{app:setup} \\
  \Cref{app:eval-methods} & \nameref{app:eval-methods} \\
  \Cref{app:eval-results} & \nameref{app:eval-results} \\
  \Cref{app:non-neuron} & \nameref{app:non-neuron} \\
  \Cref{app:nf-coverage-sweep} & \nameref{app:nf-coverage-sweep} \\
  \Cref{app:cofiring-panels} & \nameref{app:cofiring-panels} \\
  \Cref{app:mlpout-saes} & \nameref{app:mlpout-saes} \\
  \Cref{app:ninit-sweep} & \nameref{app:ninit-sweep} \\
  \Cref{app:linebreaking} & \nameref{app:linebreaking} \\
  \Cref{app:circuits} & \nameref{app:circuits} \\
  \Cref{app:arith} & \nameref{app:arith} \\
  \Cref{app:distress-setup} & \nameref{app:distress-setup} \\
  \Cref{app:distress-cards} & \nameref{app:distress-cards} \\
  \Cref{app:moe} & \nameref{app:moe} \\
  \Cref{app:moe-nf} & \nameref{app:moe-nf} \\
  \Cref{app:distill-toy} & \nameref{app:distill-toy} \\
  \Cref{app:clt-crosslayer} & \nameref{app:clt-crosslayer} \\
  \Cref{sec:clustering} & \nameref{sec:clustering} \\
  \Cref{app:artifacts} & \nameref{app:artifacts} \\
  \Cref{app:budget} & \nameref{app:budget} \\
  \Cref{app:ai-assistance} & \nameref{app:ai-assistance} \\
  \bottomrule
\end{longtable}
\endgroup

\section{Model and transcoder details}
\label{app:setup}

\subsection{Replacement model variants and cross-model results}
\label{app:setup:rep}

Attribution graphs aim to explain a model's predictions through a
sparsified description of its MLP computation. We draw a distinction
between \emph{reconstruction} --- how well the transcoder output
approximates the true MLP output, captured by L2 loss, or its normalized equivalent, FVU (fraction of variance unexplained, equivalent to normalized L2 loss), and the
attribution-graph Replacement Score \citep{ameisen2025circuit} --- and \emph{replacement} --- how
well the replacement model approximates the original when taken as a
whole, captured by KL divergence and top-1
accuracy.\footnote{Transcoders were preferred over MLP-output SAEs in
prior work in part because they achieve lower KL at similar FVU. We care about replacement because the replacement model should be capable of fully explaining behavior on its own, and perturbations to the replacement
model should have the same effect as perturbations to the original. Reconstruction sometimes produces this quality, but we show that they can diverge.}

To produce a transcoder-style replacement model from neurons, we
augment the top-$k$-pruned MLP output $y$ with up to two corrections to the baseline, \emph{none}:
\emph{recon} uses $\hat{y} = W_{\mathrm{recon}}\,y + b$ with identity
initialization; \emph{recon+skip} adds a learnable skip connection
from the layer input, $\hat{y} = W_{\mathrm{recon}}\,y +
W_{\mathrm{skip}}\,x + b$; \emph{recon+skip+rank1} adds a per-neuron
rank-1 contribution gated on the (sign-split) activation, with
positive and negative-activation regimes receiving independent rank-1
vectors. All corrections are trained jointly on a per-layer FVU loss
and a KL regularization loss\footnote{Scaled to match the FVU loss, following \citet{karvonen2025e2e}.} against the original model, on FineWeb
($10\,$k steps, batch $32$, sequence length $128$, gradient
accumulation $4$, learning rate $10^{-4}$ with $200$-step warmup and
no decay). Optimization uses Muon \citep{liu2025muon} for 2D weight
matrices and Adam for 1D parameters (biases and per-neuron rank-1
vectors). The transcoder baselines we
compare against are evaluated identically (FineWeb, $200$ batches,
positions 0--1 discarded).

The per-model Pareto plot of L0 to KL or FVU is in Figure~\ref{fig:cross-model-pareto}.
\emph{none} $\gg$ \emph{recon} $>$ \emph{recon+skip}
$\geq$ \emph{recon+skip+rank1} on FVU across every model and
$k$. Transcoders fall between \emph{recon} and \emph{recon+skip} on
FVU but at substantially higher KL than any neuron variant.
On
Gemma~3 4B the \emph{none} FVU exceeds 1 (see also \Cref{fig:pareto}), likely because of the post-feedforward LayerNorm. The finetuned down-projection can easily adjust for this, as seen in the other lines.

\begin{figure*}[h]
  \centering
  \includegraphics{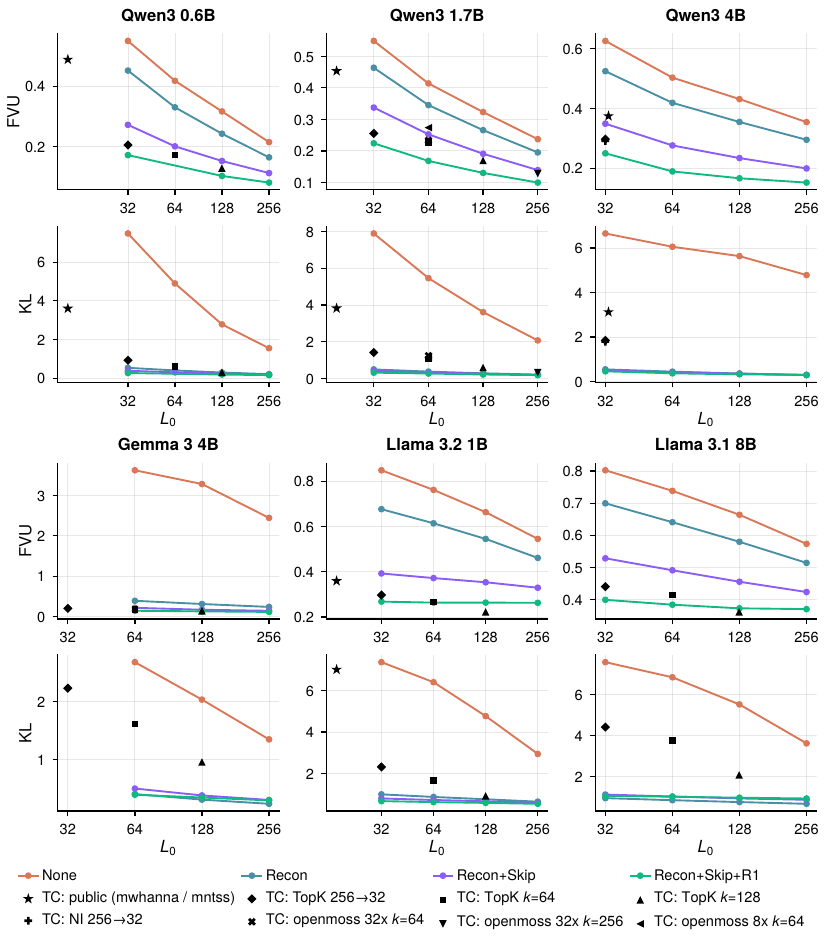}
  \caption{Per-model FVU vs $L_0$ and KL vs $L_0$ for the four sparsified MLP correction methods, with transcoder baselines overlaid in black.}
  \label{fig:cross-model-pareto}
\end{figure*}

\begin{table*}[h]
  \centering
  \small
  \setlength{\tabcolsep}{4pt}
  \begin{tabular}{llrrrl}
    \toprule
    Model & HF slug & Layers & $d_\text{model}$ & $d_\text{mlp}$ & Activation \\
    \midrule
    \multicolumn{6}{l}{\emph{Dense}} \\
    Gemma 2 2B  & \texttt{google/gemma-2-2b}        & 26 & 2304 & 9{,}216  & GeGLU \\
    Gemma 3 1B  & \texttt{google/gemma-3-1b-pt}     & 26 & 1152 & 6{,}912  & GeGLU \\
    Gemma 3 4B  & \texttt{google/gemma-3-4b-pt}     & 34 & 2560 & 10{,}240 & GeGLU \\
    Gemma 3 12B & \texttt{google/gemma-3-12b-pt}    & 48 & 3840 & 15{,}360 & GeGLU \\
    Gemma 3 27B & \texttt{google/gemma-3-27b-pt}    & 62 & 5376 & 21{,}504 & GeGLU \\
    Qwen 3 1.7B & \texttt{Qwen/Qwen3-1.7B}          & 28 & 2048 & 6{,}144  & SiGLU \\
    Qwen 3 4B   & \texttt{Qwen/Qwen3-4B}            & 36 & 2560 & 9{,}728  & SiGLU \\
    Qwen 3 8B   & \texttt{Qwen/Qwen3-8B}            & 36 & 4096 & 12{,}288 & SiGLU \\
    Qwen 3 14B  & \texttt{Qwen/Qwen3-14B}           & 40 & 5120 & 17{,}408 & SiGLU \\
    Llama 3.2 1B & \texttt{meta-llama/Llama-3.2-1B} & 16 & 2048 & 8{,}192  & SiGLU \\
    Llama 3.1 8B & \texttt{meta-llama/Llama-3.1-8B} & 32 & 4096 & 14{,}336 & SiGLU \\
    \midrule
    \multicolumn{6}{l}{\emph{Mixture of experts}} \\
    Gemma 4 26B-A4B & \texttt{google/gemma-4-26B-A4B} & 30 & 2816
      & $2{,}112 + 128 \times 704$ & GeGLU \\
    Qwen 3.5 35B-A3B & \texttt{Qwen/Qwen3.5-35B-A3B} & 40 & 2048
      & $512 + 256 \times 512$ & SiGLU \\
    \bottomrule
  \end{tabular}
  \caption{Model architectures used in this work. All models use
    gated MLPs (\texttt{up}, \texttt{gate}, \texttt{down}). GeGLU uses
    $\mathrm{GELU}_{\mathrm{tanh}}$ \citep{hendrycks2016gelu} for the
    gate, SiGLU uses SiLU. Llama and Qwen use standard pre-norm
    around each sublayer. Gemma~3 is unusual in additionally applying
    a \texttt{post\_feedforward\_layernorm} after the MLP; this
    requires dividing by the LayerNorm weight to recover a faithful
    encoder/decoder geometry (\Cref{app:non-neuron:procedure}) and shows up, among other things, as
    higher KL on Gemma~3 transcoders compared to the matched
    Llama/Qwen ones at similar $L_0$ (Figure~\ref{fig:cross-model-pareto}).
    For the two mixture-of-experts models, $d_\text{mlp}$ is written as
    the shared expert's width plus the number of routed experts times
    their width. Both models have 8 active experts per token. Gemma~4
    26B-A4B is used for bigram probing (\Cref{tab:bigram-probing}) and for
    testing expert concentration in attribution graphs (\Cref{app:moe}).
    Qwen 3.5 35B-A3B is the model we train mixture of experts transcoders
    on (\Cref{app:moe-nf}). Model families used: Gemma 2
    \citep{gemma2techreport}, Gemma 3 \citep{gemma3techreport}, Gemma 4
    \citep{gemma4techreport}, Qwen 3 \citep{qwen3techreport}, Qwen 3.5
    \citep{qwen3.5}, Llama 3 \citep{llama3herd}.}
  \label{tab:models}
\end{table*}

\begin{table*}[h]
  \centering
  \small
  \setlength{\tabcolsep}{2pt}
  \begin{tabular}{l@{\hskip 5pt}llrrrrr}
    \toprule
    Transcoder & Model & Act. & Width & $L_0$ & Alive@66\% & NF@66\% & NF\%\,acts@66\% \\
    \midrule
    GS-small        & Gemma 3 1B   & JumpReLU     & 262k &   45 & 193k ($74\%$)   & 15k ($5.9\%$)  & $57.9\%$ \\
    GS-big          & Gemma 3 1B   & JumpReLU     & 262k &  148 & 212k ($81\%$)   & 24k ($9.1\%$)  & $57.5\%$ \\
    GS-small        & Gemma 3 4B   & JumpReLU     & 262k &   47 & 175k ($67\%$)   & 23k ($8.9\%$)  & $63.8\%$ \\
    GS-big          & Gemma 3 4B   & JumpReLU     & 262k &  154 & 200k ($76\%$)   & 33k ($12.7\%$) & $66.6\%$ \\
    GS-small        & Gemma 3 12B  & JumpReLU     & 262k &   47 & 165k ($63\%$)   & 41k ($15.6\%$) & $73.7\%$ \\
    GS-small        & Gemma 3 27B  & JumpReLU     & 262k &   41 & 194k ($74\%$)   & 97k ($37.1\%$) & $82.8\%$ \\
    mwhanna         & Qwen 3 1.7B  & JumpReLU     & 164k &   19 & 140k ($85\%$)   &  7k ($4.5\%$)  & $47.1\%$ \\
    mwhanna         & Qwen 3 4B    & JumpReLU     & 164k &   35 & 153k ($93\%$)   & 13k ($8.0\%$)  & $56.1\%$ \\
    mwhanna         & Qwen 3 8B    & JumpReLU     & 164k &   44 & 156k ($95\%$)   & 15k ($9.3\%$)  & $56.6\%$ \\
    mwhanna         & Qwen 3 14B   & JumpReLU     & 164k &   56 & 162k ($99\%$)   & 19k ($11.8\%$) & $57.7\%$ \\
    LXTC-8x  & Llama 3.1 8B & TopK         &  32k &   52 &  32k ($99.2\%$) & 13k ($39.4\%$) & $64.9\%$ \\
    LXTC-32x & Llama 3.1 8B & TopK         & 131k &   53 & 127k ($97.3\%$) & 21k ($16.2\%$) & $59.2\%$ \\
    mntss-426k      & Gemma 2 2B   & JumpReLU+CLT &  16k &    9 &  16k ($100\%$)  &  5k ($28.7\%$) & $49.3\%$ \\
    mntss-524k      & Llama 3.2 1B & JumpReLU+CLT &  32k &    7 &  32k ($99\%$)   &  4k ($13.0\%$) & $38.0\%$ \\
    \midrule
    \multicolumn{8}{l}{\emph{Sparse probing only}} \\
    mntss-skip-131k & Gemma 2 2B   & TopK         & 131k & --- & ---             & ---            & --- \\
    mntss-skip-131k & Llama 3.2 1B & TopK         & 131k &  32 & 125k ($95.5\%$) & 15k ($11.7\%$) & $54.8\%$ \\
    \bottomrule
  \end{tabular}
  \caption{Models used in the analysis of open source transcoders. \emph{GS} is Gemmascope
    on Gemma~3 (\texttt{google/gemma-scope-2-\{1,4,12,27\}b-pt});
    \emph{mwhanna} is \texttt{mwhanna/qwen3-\{1.7,4,8,14\}b-transcoders};
    \emph{LXTC-8x} and \emph{LXTC-32x} are
    \texttt{OpenMOSS-Team/Llama3\_1-8B-Base-LXTC-\{8,32\}x}; \emph{mntss} is
    \texttt{mntss/clt-gemma-2-2b-426k} and
    \texttt{mntss/clt-llama-3.2-1b-524k} (cross-layer transcoders;
    the width in the table is the per-layer width and the width in the name is the maximum width of the cross-layer decoder).
    All metrics measured on FineWeb at layer $\lfloor 2N/3 \rfloor$
    (where $N$ is the number of layers; at $66\%$ of the way into the model). $L_0$ = mean active features per token.
    \emph{Alive} = features with $\mathrm{fire\_count} > 0$;
    \emph{NF} = features with encoder and decoder cosine similarity both above
    $0.6$ to the same neuron; \emph{NF\%\,acts} = share of feature
    firings that come from neuron features.
    The last two rows are the skip transcoders
    \texttt{mntss/skip-transcoders-gemma-2-2b} and
    \texttt{mntss/transcoder-Llama-3.2-1B}, used for sparse probing
    (\Cref{tab:sparse-probing-full}). Neither release states its activation
    function; we measure TopK for the Llama one (FVU $0.15$ against $0.99$ for
    ReLU) and assume it for the Gemma one, which we do not measure further.}
  \label{tab:transcoders}
\end{table*}

\subsection{Transcoder training details}

Our TopK transcoders are trained using Dion \citep{ahn2025dion} with
$\text{lr}=10^{-2}$, or $\text{lr}=5\times10^{-3}$ for less stable
architectures like NFT and SNRL. We use power-law $k$ annealing (as
distinct from the linear $k$ annealing of LlamaScope
\citep{he2024llamascope}) from $256$ down to $32$ between steps
$1{,}500$--$10{,}000$ (out of a total $15{,}000$). We use a batch size of $256$ at sequence
length $128$ on the SmolLM2 dataset \citep{allal2025smollm2}. The
JumpReLU \citep{rajamanoharan2024jumprelu} transcoders are trained with a different codebase on the same
dataset at the same effective batch size of $32{,}768$ tokens, with
learning rate $2\times10^{-4}$, sparsity coefficient $10$, and
$4\times10^{8}$ training tokens.

Our
JumpReLU pipeline uses a tanh sparsity loss with
decoder norm weighting (and therefore no decoder normalization), following \citet{conerly2025january}:
$\mathcal{L}_\mathrm{sparsity} = \lambda \sum_j
\tanh\!\bigl(c \cdot |f_j| \cdot \|W_\mathrm{dec}^{(j)}\|\bigr)$,
with $c = 4$ and $\lambda$ linearly ramped from $0$ to its final value
$10$ over training. An auxiliary pre-activation penalty,
$3 \times 10^{-5} \cdot \sum_j \mathrm{ReLU}(\theta_j - z_j) \cdot
\|W_\mathrm{dec}^{(j)}\|$, is added to prevent dead features.
Thresholds $\theta_j$ are initialized at
$\log 0.1$. The straight-through estimator on
threshold gradients uses a rectangular kernel of width $\epsilon = 2.0$
centred on $\theta_j$.

For the bigram probing experiments on Qwen 3 4B
(\Cref{tab:bigram-probing}) we use a TopK
transcoder trained with a Matryoshka objective \citep{bussmann2025matryoshka} with otherwise the same recipe as our other
transcoders ($M{=}262{,}144$, $k{=}32$, $15{,}000$ steps,
\texttt{lr}${=}5\,{\times}\,10^{-3}$). The Matryoshka loss works as follows: on each training step we sample, per
batch element, a prefix size from a log-uniform distribution over
$[8\,d_{\text{model}},\,M] = [20{,}480,\,262{,}144]$ features, restrict
the encoder pre-activations to that prefix, apply top-$k$ within the
prefix, and sum the FVU losses across all sampled slices. We choose the Matryoshka variant for bigram probing
because, at identical AUC, the plain TopK transcoder
does not select features whose top-1
max-activating example contains the target bigram: $2.6\%$ vs
Matryoshka's $21.7\%$.

The NFT architecture has a model-FLOPs-utilization
advantage at training and inference time: an NFT decoder forward pass
in principle reaches up to ${\sim}160\times$\footnote{Or the number of neurons divided by neuron K} the throughput of a dense
TopK encoder thanks to the limited number of rows that need to be considered. In practice the achievable speedup is only about $2 \times$: the computation is bounded
by memory bandwidth rather than compute, since the sparse-matmul
kernels still have to stream the encoder rows for each active feature.

The two pipelines initialize encoder weights differently. Our TopK
pipeline draws encoder rows from $\mathcal{U}(-1/\sqrt{d_\text{model}},
+1/\sqrt{d_\text{model}})$, giving every feature a width-independent
initial encoder norm of $\sqrt{1/3} \approx 0.577$ regardless of
dictionary size. The JumpReLU pipeline draws from
$\mathcal{U}(-1/\sqrt{M}, +1/\sqrt{M})$ instead, so the initial
encoder norm shrinks as $\sqrt{d_\text{model}/M}$ as the dictionary
widens. This narrower init, combined with the $L_1$ sparsity
penalty, drives more non-neuron features to zero firing during
training: at $1$M features on Gemma~3 4B L22 the JumpReLU
run leaves $66\%$ of non-neuron features below decoder norm $0.1$
while every neuron feature stays alive, whereas the TopK pipeline
keeps essentially all of its $1$M features firing
(\Cref{sec:nfc:init-distance}).

\subsection{Gated-MLP neuron geometry}
\label{app:neuron-geometry}

This subsection expands \S\ref{sec:mlp:directions}: it works out why each
sign-split neuron is selective for a single input direction.

\begin{figure*}[t]
  \centering
  \begin{subfigure}[t]{0.32\textwidth}
  \vspace{0pt}%
    \centering
    \includegraphics{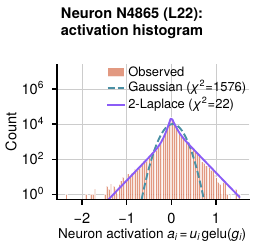}
    \caption{Activation histogram for one neuron (Gemma~3 4B, L22); sharply
      peaked at zero with heavy tails, as shown by the Laplace mixture fitting better than a Gaussian.}
    \label{fig:neuron-acts}
  \end{subfigure}\hfill
  \begin{subfigure}[t]{0.32\textwidth}
  \vspace{0pt}%
    \centering
    \includegraphics{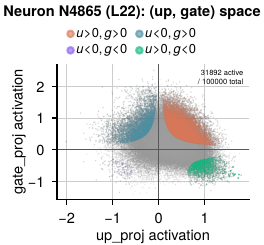}
    \caption{Joint distribution of $(u_i, g_i)$ for the same neuron. $g_i > 0$ contains most of the activations, with $u_i > 0$ (orange) and $u_i < 0$ (teal) being equally common.}
    \label{fig:neuron-upgate}
  \end{subfigure}\hfill
  \begin{subfigure}[t]{0.32\textwidth}
  \vspace{0pt}%
    \centering
    \includegraphics{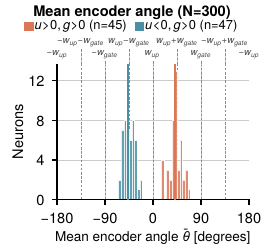}
    \caption{Per-neuron mean encoder angle across $300$ sampled neurons
      from L15/L20/L22, restricted to $|a_i| > 1.0$.}
    \label{fig:neuron-angle}
  \end{subfigure}
  \caption{Gated-MLP neuron geometry. Per-neuron activations are heavy-tailed
    (left); high-magnitude activations split into two regimes that share the
    same gate-positive half-plane (centre); and within each regime the
    effective encoder direction concentrates along $W_\mathrm{up}^{(i)} \pm
    W_\mathrm{gate}^{(i)}$ across neurons (right).}
  \label{fig:neuron-geometry}
\end{figure*}

We work with gated MLPs \citep{shazeer2020glu}, standard in recent open-source LMs in the 1B--27B range (Gemma, Qwen, Llama). The layer
maps an input $x \in \mathbb{R}^{d_\mathrm{model}}$ to
\[
  \mathrm{MLP}(x) = W_\mathrm{down}\bigl(\sigma(W_\mathrm{gate}\, x)
                                          \odot W_\mathrm{up}\, x\bigr),
\]
where $W_\mathrm{gate}, W_\mathrm{up} \in \mathbb{R}^{d_\mathrm{ffn}
\times d_\mathrm{model}}$ are the gate and up projections,
$W_\mathrm{down} \in \mathbb{R}^{d_\mathrm{model} \times
d_\mathrm{ffn}}$ is the down projection, $\sigma$ is the gate
nonlinearity (SiLU or GELU), and $\odot$ is the elementwise product.
Writing $g_i = W_\mathrm{gate}^{(i)}\, x$ and $u_i = W_\mathrm{up}^{(i)}\, x$
for the $i$th row of each projection, we refer to the elementwise
hidden activations $a_i = \sigma(g_i)\, u_i$ as the layer's
\emph{neurons}. The two projections $W_\mathrm{up}^{(i)}$ and
$W_\mathrm{gate}^{(i)}$ are approximately orthogonal in the models we surveyed.

Per-neuron, the marginal activation distribution is sharply peaked at
zero with heavy two-sided tails (Figure~\ref{fig:neuron-acts}). The fact that these heavy-tailed activations capture much of important behavior (see \Cref{sec:mlp:kl}) suggests the network is using them to represent important features.

The high-magnitude activations fall into two regimes that
are roughly equally common. Both have $g_i > 0$ with different signs
of $u_i$: gate-positive with $u_i \gg 0$ (so $a_i > 0$) and gate-positive
with $u_i \ll 0$ (so $a_i < 0$). $g_i < 0$ is uncommon in most neurons
(Figure~\ref{fig:neuron-upgate}).

Within each regime, the neuron is selective for a single input direction. In the linear regime of GeLU/SiLU at large $g_i$, $\sigma(g_i) \approx g_i$ and $\sigma'(g_i) \approx 1$, so the
gradient
\[
  \nabla_x a_i \;\approx\; u_i\, W_\mathrm{gate}^{(i)}
                       \;+\; g_i\, W_\mathrm{up}^{(i)}
\]
lies in the 2D subspace spanned by $W_\mathrm{up}^{(i)}$ and
$W_\mathrm{gate}^{(i)}$. When $|g_i|$ and $|u_i|$ are comparable, the
gradient concentrates along $W_\mathrm{up}^{(i)} + W_\mathrm{gate}^{(i)}$
when $g_i > 0$, $u_i \gg 0$ and along
$W_\mathrm{up}^{(i)} - W_\mathrm{gate}^{(i)}$ when $g_i > 0$, $u_i \ll 0$. Figure~\ref{fig:neuron-angle} shows the average encoder direction in high activations roughly matches predictions in 3 layers of Gemma 3 4B.

\subsection{Neuron top-$k$ threshold}

We use $k \in \{32, 64, 128\}$ throughout the paper. The threshold a
top-$k$ implements (the smallest $|a_i|$ kept) varies across both
layers and models: in models without post-feedforward layernorm, the
residual stream norm grows in late layers, causing neuron activation
magnitudes to rise, so a single absolute activation threshold
across layers does not fit while a top-$k$ tracks the per-layer scale.
Figure~\ref{fig:threshold-per-layer} reports the implied threshold per
layer for Qwen~3 and Gemma~3.

\begin{figure*}[h]
  \centering
  \includegraphics{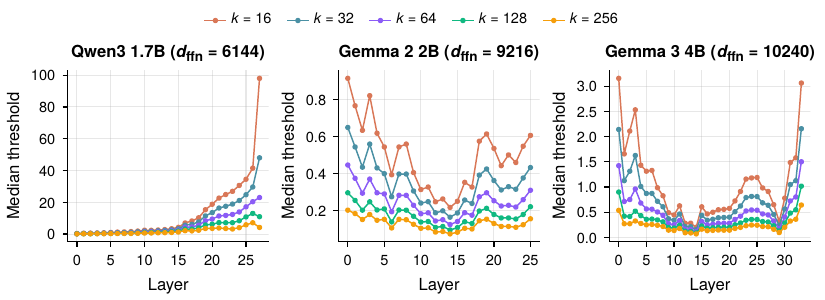}
  \caption{Activation threshold per layer for Qwen and Gemma models at various TopK settings. Due to the lack of post-feedforward
    LayerNorm in Qwen~3 the threshold grows by an order of magnitude
    from early to late layers.}
  \label{fig:threshold-per-layer}
\end{figure*}

\section{Evaluation methodology}
\label{app:eval-methods}

\subsection{Sparse probing}
\label{app:eval-methods:sparse}

\begin{table}[t]
  \centering
  \begin{tabular}{lccc}
    \toprule
    Model & Neurons & Transcoder & $\Delta$ \\
    \midrule
    Llama 3.1 8B   & 0.856 & 0.845 & $+0.011 \pm 0.007$ \\
    Gemma 3 4B     & 0.854 & 0.832 & $+0.022 \pm 0.008$ \\
    Qwen 3 8B      & 0.851 & 0.806 & $+0.045 \pm 0.011$ \\
    Qwen 3 4B      & 0.833 & 0.794 & $+0.039 \pm 0.011$ \\
    Qwen 3 1.7B    & 0.811 & 0.762 & $+0.050 \pm 0.010$ \\
    Gemma 3 1B     & 0.834 & 0.792 & $+0.042 \pm 0.011$ \\
    Gemma 2 2B     & 0.851 & 0.780 & $+0.071 \pm 0.012$ \\
    Llama 3.2 1B   & 0.847 & 0.802 & $+0.045 \pm 0.011$ \\
    \bottomrule
  \end{tabular}
  \caption{Sparse probing mean AUC at $k{=}1$ across the $113$
    \citep{kantamneni2025sparseprobing} tasks.
    \emph{Neurons}: dense MLP neurons.
    \emph{Transcoder}: best-available transcoder per model
    (\Cref{tab:transcoders}).
    $\Delta$ shows the per-task mean difference.
    Neurons match or exceed the transcoder across all models.
    Per-task and full $k$-sweep tables in Appendix~\ref{app:eval-results:sparse}.}
  \label{tab:sparse-probing}
\end{table}

\begin{figure}[t]
  \centering
  \includegraphics{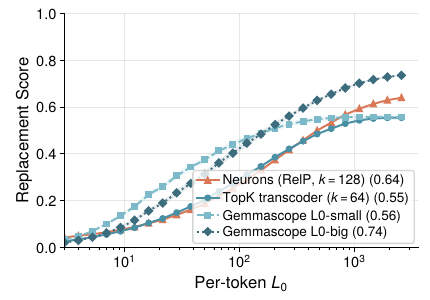}
  \caption{Replacement Score and per-token $L_0$.
    Gemmascope L0-big dominates at high $L_0$, and is still the best method together with L0-small at low $L_0$.}
  \label{fig:rs-l0-pareto}
\end{figure}

\begin{figure}[t]
  \centering
  \begin{minipage}{0.62\linewidth}
    \fcard{Gemma 3 27B-it}{41}{7312}{``apology (apologize, apologies)''}{%
      \ctxrow{23.00}{\textellipsis So first, my}{apologies}{for taking a month\textellipsis}
      \ctxrow{17.12}{\textellipsis was also an}{apology}{on the official\textellipsis}
      \ctxrow{13.06}{\textellipsis features. Never}{apologize}{for being the way\textellipsis}
      \ctxrow{12.75}{\textellipsis useful as an}{apology}{for the crueler\textellipsis}
    }{%
      Promoted: ``apologize'', ``apology'', ``apologies'', ``apologizing'', ``apologized'' \\
      Suppressed: ``érm'', ``{\mlymfont ധ്യമ}'', ``Pension'', ``{\devafont उद}'', ``{\thaiscriptfont สัน}''
    }
  \end{minipage}
  \caption{Representative distress circuit feature found by
    neuron attribution (Gemma 3 27B-it, L41/N7312).}
  \label{fig:distress-card}
\end{figure}

We follow the basic sparse probing methodology from \citet{gurnee2023haystack} as
adapted by \citet{kantamneni2025sparseprobing} ($113$ binary classification
tasks covering sentiment, headlines, factual recall, code, language,
toxicity, and other concepts). Both methodology and tasks match
\citet{kantamneni2025sparseprobing} unless noted otherwise.

\paragraph{Activations.}
For each task we collect feature post-activations at every transformer
layer over up to ${\sim}\,1024$ training and several hundred test
samples drawn from the task's prepared sequences. For neurons we use
the post-GeGLU (Gemma) or post-SiLU (Llama, Qwen) activations
$\sigma(g_i)\,u_i$ split by sign (Section~\ref{sec:mlp:directions}). For
learned dictionaries we use post-activation features (after JumpReLU
or TopK).

\paragraph{Feature selection.}
For each task, following \citet{kantamneni2025sparseprobing}, we rank features by the absolute mean-activation difference between positive and negative samples on the train split, specifically as implemented in \texttt{sae\_probes}. We pick the top $k$ features by this score.

\paragraph{Probe.}
For binary classification, we train $L_1$-penalized logistic regression on the selected $k$ features. We
sweep penalty over $11$ values ($0$ together with $10$ values
log-spaced over $10^{-4}$--$10^{4}$), pick the best penalty by AUC on
a held-out $20\%$ validation slice of the training set, then refit on
the full training set. We report test AUC at each
$k \in \{1, 2, 4, 8, 16, 32\}$. Each task is run independently at
every layer; the per-task peak layer is selected post-hoc as the one
maximizing test AUC.

\paragraph{Models and dictionaries.}
We use the models mentioned in \Cref{tab:transcoders}: Gemmascope L0-small or L0-big for Gemma 3 4B
\citep{mcdougall2025gemmascope2}, \texttt{mntss-skip-131k} for Gemma 2 2B and
Llama 3.2 1B, \texttt{mwhanna} for Qwen 3 (164k width), and both LXTC-8x and
LXTC-32x for Llama 3.1 8B.

\subsection{Bigram probing}
\label{app:eval-methods:bigram}

We extend sparse probing to detection of common pairs of tokens (e.g., ``social security'', ``prime factors'', ``public health''), following the
methodology for constructing hard negatives from \citet{gurnee2023haystack} (Section 5.1). Phrase
bigrams are extracted from a 1B-token sample of FineWeb.
From $30.9$M phrase-bigram types we keep $11{,}614$ that
have count ${\geq}\,3000$ and enough negatives, then select the top
$1{,}500$ by pointwise mutual information $\log\frac{p(a,b)}{p(a)\,p(b)}$
between \texttt{word\_a} and \texttt{word\_b} (range $5.26$ to $14.30$).

For each bigram we extract three sample types ending at the
``\texttt{word\_b} slot'' position: \emph{positive}
(\texttt{word\_a}~\texttt{$\overline{\,\text{word\_b}\,}$}),
\emph{missing-first} (\texttt{$\overline{\,\text{word\_b}\,}$} appears
without \texttt{word\_a} preceding), and \emph{missing-second}
(\texttt{word\_a} appears without \texttt{$\overline{\,\text{word\_b}\,}$}
following). Each dataset has $3{,}200$ samples ($1{,}600$ positive,
$800$ of each negative type).

Probing uses Newton/IRLS L2 logistic regression on top-$k$
highest-attribution features, with the peak layer selected per dataset by
validation AUC. We report two metrics: held-out test AUC, and the
fraction of the selected feature's top-10 max-activating examples on
FineWeb that contain the target bigram.

\subsection{Sort eval}
\label{app:eval-methods:sorteval}

We follow the procedure in \citet{anthropicaug2024} for judging the interpretability of transcoder features, including the prompt and feature selection method. Each trial shows a Claude Haiku 4.5 judge
$10$ \emph{display} context excerpts for feature $A$ and $10$ for feature $B$, plus a
held-out \emph{query} excerpt drawn from one of them, and asks the judge to
classify the query as $A$ or $B$. We run $500$ trials per
configuration. Excerpts are windows of up to $40$ tokens around the
peak-activation token.

We study two regimes for picking the $10$ display excerpts.
Under \emph{top sampling} (``top-of-distribution''), we show the $10$
highest-activating excerpts cached for the feature; in this case, for neurons, the $k$ used for TopK selection does not matter. Under \emph{quantile
sampling} (``broad''), we sort the feature's cached excerpts by peak
activation, split the list into $10$ equal-count buckets, and draw one
excerpt uniformly from each. Quantile sampling shows
contexts from across the activation range, not only the peaks.

For each transcoder, we precompute an activation cache on a $10$M-token sample of SmolLM2 (context length $128$
sequences, BOS masked if present). The same
caching procedure is used for the neuron baseline (with TopK over sign-split post-GeGLU
activations) and is reused across all sort-eval configurations on a
given model, so two candidates compared in the same trial draw their
display excerpts from the same underlying tokens.

\subsection{Cluster sort eval}
\label{app:eval-methods:cluster-sorteval}

The cluster sort evals (\S\ref{sec:clustering:sort-eval}) adapt the
sort eval to evaluating ways to cluster single neurons. For each neuron and sign, we store
up to $4096$ contexts with high activations, and clusters partition that panel. A trial picks two
clusters $A$ and $B$ from the same panel, shows the judge $10$
contexts from each, and asks for a binary classification of a held-out
query context. We run two variants:

\begin{itemize}
  \item \emph{Cluster match accuracy.} The display and query contexts are
    all drawn from the cluster's own panel firings, picking the top $10$
    by activation. This measures whether the partition has produced
    distinct concepts (the closest match to sort evals, but limited to a neuron's firings).
  \item \emph{Top neuron match accuracy.} The display for both of the clusters is replaced by a combination of maximum activations of the
    cluster's top-$4$ closest neurons. The panel neuron itself is dropped from the selection pool. The query is still a
    context from cluster $A$ or $B$. This tests whether the
    cluster's identity is fully described by its closest neurons, while at the same time capturing cluster distinguishability in general like the cluster match eval.
\end{itemize}

For each neuron we sample $25$ cluster pairs (neurons
where either cluster has fewer than $15$ qualifying contexts are
dropped). We sample random pairs of clusters. The activations of the closest 4 neurons for the top neuron eval are extracted at the sign of the neuron's weight (positive weight: maximum activations; negative weight: minimum activations).

The 50 neurons used for the evaluation are picked randomly from L22 neurons with at least $256$ positive firings in the cache. For
each neuron we take the top $4096$ firings on $1$M SmolLM2 tokens, zero out the neuron's own column in the
post-activation matrix and the eight globally densest neurons, and use these masked vectors as the panel data for
$k$-means
. We compute transcoder activations on the cache and assign each panel position's cluster to the highest-activating feature
for which the neuron being evaluated is the closest to the feature's encoder out of all neurons. For Gemmascope, we match on the decoder side instead, because its encoders are not aligned enough to create clusters of comparable sizes. All other dictionaries are matched on the encoder side.

\section{Full evaluation results}
\label{app:eval-results}

\subsection{Sparse probing}
\label{app:eval-results:sparse}

Methodology: Appendix~\ref{app:eval-methods:sparse}. Across nine
(model, transcoder) pairs spanning four architectures, sign-split
neurons match or exceed the canonical learned dictionary on every
pair at every $k$ we tested (Table~\ref{tab:sparse-probing-full}).

\begin{table*}[h]
  \centering
  \footnotesize
  \setlength{\tabcolsep}{2pt}
  \begin{tabular}{l@{\hskip 3pt}l@{\hskip 3pt}cc@{\hskip 3pt}cc@{\hskip 3pt}cc@{\hskip 3pt}rrr}
    \toprule
    & Name & \multicolumn{2}{c}{AUC, $k\,{=}\,1$} & \multicolumn{2}{c}{AUC, $k\,{=}\,4$} & \multicolumn{2}{c}{AUC, $k\,{=}\,16$} & \multicolumn{3}{c}{Tasks won ($k\,{=}\,1$)} \\
    \cmidrule(lr){2-2} \cmidrule(lr){3-4} \cmidrule(lr){5-6} \cmidrule(lr){7-8} \cmidrule(lr){9-11}
    Model & Dictionary & Neurons & Dictionary & Neurons & Dictionary & Neurons & Dictionary & Neurons & Dictionary & Ties \\
    \midrule
    Llama 3.1 8B  & LXTC-8\,$\times$    & \textbf{0.883} & 0.843 & \textbf{0.908} & 0.888 & \textbf{0.925} & 0.911 & \textbf{79} & 15 & 19 \\
    Llama 3.1 8B  & LXTC-32\,$\times$   & \textbf{0.883} & 0.839 & \textbf{0.908} & 0.886 & \textbf{0.925} & 0.910 & \textbf{80} & 15 & 18 \\
    Gemma 3 4B    & GS-big              & \textbf{0.877} & 0.849 & \textbf{0.899} & 0.888 & \textbf{0.919} & 0.907 & \textbf{70} & 18 & 25 \\
    Qwen 3 8B     & mwhanna             & \textbf{0.876} & 0.829 & \textbf{0.902} & 0.877 & \textbf{0.921} & 0.900 & \textbf{83} & 13 & 17 \\
    Qwen 3 4B     & mwhanna             & \textbf{0.864} & 0.819 & \textbf{0.891} & 0.866 & \textbf{0.909} & 0.890 & \textbf{79} & 15 & 19 \\
    Qwen 3 1.7B   & mwhanna-163k        & \textbf{0.837} & 0.785 & \textbf{0.866} & 0.837 & \textbf{0.886} & 0.859 & \textbf{80} & 15 & 18 \\
    Gemma 3 1B    & GS-small (16k)      & \textbf{0.855} & 0.811 & \textbf{0.881} & 0.858 & \textbf{0.902} & 0.879 & \textbf{82} & 12 & 19 \\
    Gemma 2 2B    & mntss-skip-131k     & \textbf{0.872} & 0.802 & \textbf{0.897} & 0.854 & \textbf{0.913} & 0.882 & \textbf{87} &  8 & 18 \\
    Llama 3.2 1B  & mntss-skip-131k     & \textbf{0.861} & 0.817 & \textbf{0.888} & 0.863 & \textbf{0.906} & 0.887 & \textbf{83} &  6 & 24 \\
    \bottomrule
  \end{tabular}
  \caption{Sparse probing results across nine (model, dictionary)
    pairs on the $113$-task suite of \citet{kantamneni2025sparseprobing}.
    \emph{Neurons}: post-GeGLU/SiLU sign-split neurons. \emph{Dictionary}:
    canonical post-activation learned dictionary
    (Appendix~\ref{app:eval-methods:sparse}).
    Mean peak test AUC across the $113$ tasks (peak = max over layers
    per task), then averaged across tasks; bold marks the higher of
    the two within each $(k, \text{row})$.
    \emph{Tasks won} columns: number of tasks where one method beats the
    other by ${>}\,0.005$ AUC at $k\,{=}\,1$; ties otherwise. Bold
    marks the larger of the two win counts per row. Win rates and AUC gaps
    narrow with $k$ but the sign of the gap is preserved at every $k$ on
    every model.}
  \label{tab:sparse-probing-full}
\end{table*}

\subsection{Bigram probing}
\label{app:eval-results:bigram}

On both Gemma 3 4B and Qwen 3 4B, the dictionary outperforms
sign-split neurons at every layer L0--L5 and every $k$ we tested
(Table~\ref{tab:bigram-probing}). The gap is largest at $k\,{=}\,1$
and narrows slightly with $k$. The transcoder advantage also shows up
qualitatively: for each of the $1{,}500$ bigrams we pick the best
$k\,{=}\,1$ probe across L0--L5 (same picker on every method) and
check whether the chosen feature's top-1 max-act example contains the
target bigram. Gemmascope hits $75.6\%$ vs $0.0\%$ for sign-split
Gemma~3 4B neurons; the Qwen~3 4B matryoshka-TopK transcoder hits $21.7\%$ vs $0.6\%$ for its
sign-split neurons. The plain non-matryoshka TopK transcoder for Qwen
has similar AUC  but matches bigrams only on $2.6\%$ of the features, so the gap on this metric is driven by training-time
inductive bias more than feature quality.

\begin{table*}[h]
  \centering
  \small
  \setlength{\tabcolsep}{4pt}
  \begin{tabular}{l@{\hskip 6pt}cc@{\hskip 6pt}cc@{\hskip 10pt}cc@{\hskip 6pt}cc@{\hskip 10pt}cc@{\hskip 6pt}cc}
    \toprule
    & \multicolumn{4}{c}{Gemma 3 4B} & \multicolumn{4}{c}{Qwen 3 4B} & \multicolumn{4}{c}{Gemma 4 26B-A4B} \\
    \cmidrule(lr){2-5} \cmidrule(lr){6-9} \cmidrule(lr){10-13}
    & \multicolumn{2}{c}{Neurons} & \multicolumn{2}{c}{Gemmascope} & \multicolumn{2}{c}{Neurons} & \multicolumn{2}{c}{Matryoshka TopK} & \multicolumn{2}{c}{Neurons} & \multicolumn{2}{c}{Neurons, routed} \\
    \cmidrule(lr){2-3} \cmidrule(lr){4-5} \cmidrule(lr){6-7} \cmidrule(lr){8-9} \cmidrule(lr){10-11} \cmidrule(lr){12-13}
    Layer & $k{=}1$ & $k{=}8$ & $k{=}1$ & $k{=}8$ & $k{=}1$ & $k{=}8$ & $k{=}1$ & $k{=}8$ & $k{=}1$ & $k{=}8$ & $k{=}1$ & $k{=}8$ \\
    \midrule
    L0 & .891 & .991 & \textbf{.988} & \textbf{.996} & .946 & .995 & \textbf{.992} & \textbf{.999} & .723 & .895 & \textbf{.902} & \textbf{.983} \\
    L1 & .886 & .982 & \textbf{.977} & \textbf{.990} & .955 & .993 & \textbf{.987} & \textbf{.998} & .844 & .977 & \textbf{.957} & \textbf{.997} \\
    L2 & .854 & .987 & \textbf{.985} & \textbf{.995} & .950 & .992 & \textbf{.979} & \textbf{.997} & .833 & .983 & \textbf{.937} & \textbf{.986} \\
    L3 & .875 & .985 & \textbf{.984} & \textbf{.996} & .949 & .990 & \textbf{.977} & \textbf{.996} & .783 & .982 & \textbf{.929} & \textbf{.987} \\
    L4 & .871 & .984 & \textbf{.973} & \textbf{.991} & .955 & .991 & \textbf{.978} & \textbf{.997} & .873 & \textbf{.987} & \textbf{.917} & .981 \\
    L5 & .874 & .984 & \textbf{.964} & \textbf{.988} & .960 & .993 & \textbf{.984} & \textbf{.997} & .839 & \textbf{.978} & \textbf{.873} & .957 \\
    \midrule
    Match & \multicolumn{2}{c}{$0.0\%$} & \multicolumn{2}{c}{$\mathbf{75.6\%}$} & \multicolumn{2}{c}{$0.6\%$} & \multicolumn{2}{c}{$\mathbf{21.7\%}$} & \multicolumn{2}{c}{$0.9\%$} & \multicolumn{2}{c}{$\mathbf{20.1\%}$} \\
    \bottomrule
  \end{tabular}
  \caption{Per-layer mean test AUC on the $1{,}500$ bigram probing tasks
    (Appendix~\ref{app:eval-methods:bigram}), at $k\,{\in}\,\{1,8\}$
    selected features per probe. \emph{Match} reports the rate at which
    the chosen (best $k\,{=}\,1$ across L0--L5) feature's top-1 max-activating example contains the
    target bigram. The transcoder
    wins every AUC comparison on every model, and its features match the
    bigram in their top-1 example far more often than sparsified neurons do. The MoE model, Gemma 4, is close to the some of the transcoders' level of performance; this is preliminary evidence that scaled up MoE training could lead to similar monosemanticity to that seen in transcoders.}
  \label{tab:bigram-probing}
\end{table*}

\section{Features not aligned to single neurons}
\label{app:non-neuron}

In this section, we describe the methodology for measuring the faithfulness of non-neuron features for \cref{sec:nfc:faithfulness}.

\subsection{Faithfulness measurement procedure}
\label{app:non-neuron:procedure}

\paragraph{MLP Jacobian.} For Gemma's gated MLP
$\mathrm{MLP}(x) = W_{\mathrm{down}}\,(\sigma(W_g\, x)\!\odot\!W_u\, x)$
(\S\ref{sec:mlp:directions}), writing $g = W_g\, x$, $s_2 = \sigma(g)$,
and $s_1 = \sigma'(g) \odot W_u\, x$, the per-position Jacobian is
\begin{equation}
  J_{\mathrm{mlp}}(x) = W_{\mathrm{down}}\bigl[\mathrm{diag}(s_1)\,W_g
                                              + \mathrm{diag}(s_2)\,W_u\bigr].
  \label{eq:jmlp-app}
\end{equation}
Gemmascopes are trained to predict the output of the MLP after the post-feedforward LayerNorm, but we evaluate $J_\mathrm{mlp}$ in the pre-LN MLP-output basis: the
affine scaling in the LayerNorm $(1+w_\mathrm{ln})$ is not folded into
$W_\mathrm{down}$, and transcoder decoders are instead divided elementwise by this factor; this is necessary because of the existence of $<10$ neurons or features at many layers that are aligned with the LayerNorm weight and dominate the Jacobian cosine similarity.

\paragraph{Methods.} Each method $M$ defines a linearization $J_M(x)$.
\begin{itemize}
  \item \textbf{Transcoders.} With active set $\mathcal{A}(x)$ on
    input $x$,
    \(J_{\mathrm{tc}}(x) = \sum_{f \in \mathcal{A}(x)} d_f\,e_f^{\!\top}\),
    where $e_f, d_f$ are feature $f$'s encoder and decoder rows.\footnote{The activations of transcoders are piecewise linear, so no reweighting is necessary.}

    We also compute Jacobians for subgroups of features by simply restricting the active set. We track neuron features (encoder and decoder similarity above 0.7) and the rest, non-neuron features.
  \item \textbf{Top-$k$ sparsified MLP at matched $L_0$.} At each
    token we keep the top $k = |\mathcal{A}(x)|$ neurons by absolute
    post-GeGLU activation $|s_2 \odot W_u\, x|$, producing an indicator
    mask $m \in \{0,1\}^{d_\mathrm{ffn}}$, which we multiply Eq.~\ref{eq:jmlp-app} by elementwise:
    \begin{equation*}
      \begin{aligned}
        J_\mathrm{topk}(x) = W_\mathrm{down}\bigl[\, &\mathrm{diag}(m \odot s_1)\,W_g \\
                                                    &{} + \mathrm{diag}(m \odot s_2)\,W_u\bigr].
      \end{aligned}
    \end{equation*}
  \item \textbf{SVD ceiling.} The cosine similarity between $J_\mathrm{mlp}(x)$
    and its best rank-$L_0$ approximation, the performance of a maximally faithful transcoder. We truncate the SVD of $J_\mathrm{mlp}(x)$ to the top $L_0$ components
    and report $\sqrt{\sum_{i \le L_0} \sigma_i^2 / \sum_i \sigma_i^2}$,
    where $\sigma_1 \geq \sigma_2 \geq \cdots$ are the singular values.
\end{itemize}

As mentioned above, we compute elementwise cosine similarity of the Jacobian matrices. Positions where
$\|J_{\mathrm{mlp}}(x)\|_F < 10^{-6}$ or $|\mathcal{A}(x)| = 0$ are
discarded.

\paragraph{Per-feature rank-1 contribution.} For
Figure~\ref{fig:appendix-faithfulness-rank1}, we measure the cosine similarity of the MLP with the individual features' rank-1 contributions
$\cos(d_f\,e_f^{\!\top}, J_{\mathrm{mlp}}(x))$ and average only over positions where $f$ is
active.
We
similarly report the equivalent quantity for individual neurons,
$\cos(W_{\mathrm{down}}^{(i)} (e^{\,\mathrm{n}}_i)^{\!\top},
J_{\mathrm{mlp}}(x))$, where $e^{\,\mathrm{n}}_i$ is the
effective encoder, or the gradient of neuron $i$'s post-activation w.r.t. the input $x$.

\paragraph{Evaluation set.} We use $64$ FineWeb sequences of length
$128$, dropping the $3$ leading positions. Analyses for different models share
the same seed and sequence pool.

\subsection{Faithfulness details and results}
\label{app:non-neuron:faith-results}

Figure~\ref{fig:appendix-faithfulness-cross-model} shows per-layer
Jacobian alignment for all six Gemmascope transcoders we
evaluate (Gemma 3 4B/12B/27B in L0-small and -big variants).
For all six transcoders, on late layers, we see similar behavior: transcoders trail behind neurons by 0.1--0.2 points, and the curves for neuron features and full transcoders overlap almost perfectly, suggesting that neuron features completely explain the first-order faithfulness of transcoders.

Non-neuron features explain little of the Jacobian faithfulness (only around 0.1, constant across layers), with the exception of the last few layers for unclear reasons. We see that there is little gain to be had by perfectly modeling the Jacobian with the SVD ceiling and that approximating the structure present in the sparse neuron Jacobian gets most of the way to the best possible faithfulness for a transcoder.

\begin{figure*}[h]
  \centering
  \includegraphics{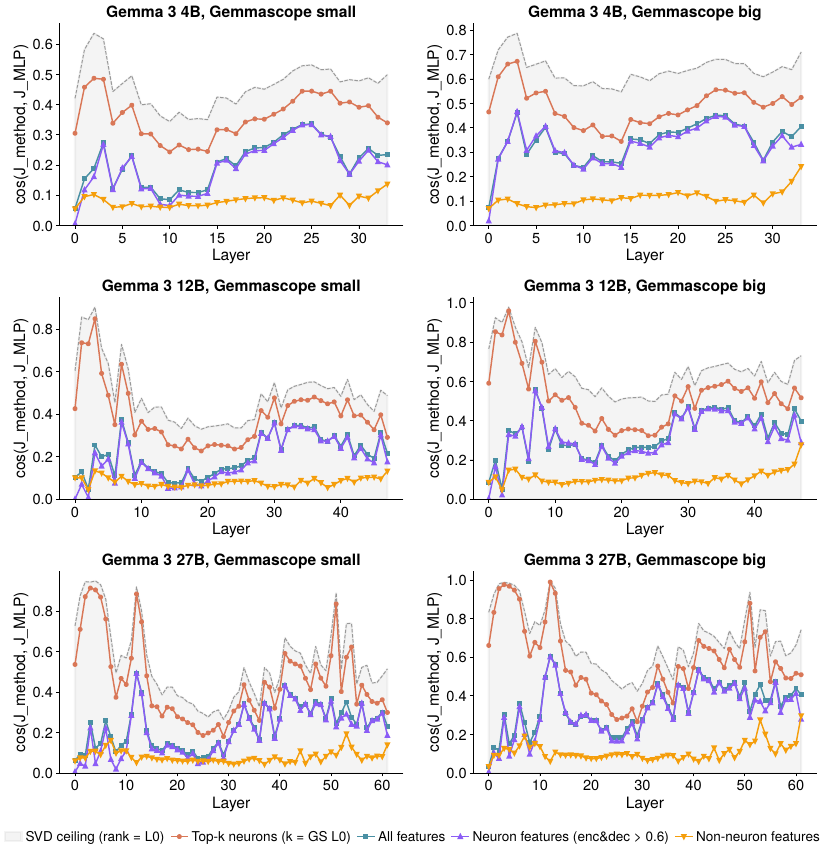}
  \caption{Per-layer mean Jacobian cosine alignment for the six Gemmascope
    transcoders we evaluate.
    }
  \label{fig:appendix-faithfulness-cross-model}
\end{figure*}

\subsection{Percentage of neuron features across $k$ and width}
\label{app:non-neuron:scaling}

\paragraph{Setup.}
We sweep transcoder width $M$ in both training pipelines from
Appendix~\ref{app:setup}. Our TopK pipeline covers $k{\in}\{32,64, 128\}$ at
$M{\in}\{32\text{k}, 65\text{k}, 128\text{k}, 256\text{k}, 1\text{M}\}$. The JumpReLU pipeline covers JumpReLU at $L_0{\approx}30$ and TopK
$k{=}64$ at $M{\in}\{256\text{k}, 512\text{k}, 1\text{M}\}$. All
transcoders share training data (SmolLM2-135M-10B, context length $128$) and target Gemma~3 4B L22. Each cell uses $n{=}2$
seeds, and, as in the main body, a threshold of $0.7$ on cosine similarities to the encoder and decoder of either a neuron or a feature in a different seed.

\paragraph{Percentage in the dictionary and in activations.}
The percentage of neuron features in the dictionary (top row of
\cref{fig:nfc-scaling}) decays predictably with width in every series. Specifically, the decay is close to a power law with exponent $-1$, which is what would be predicted if the number of neuron features stays constant while the dictionary grows linearly.

The
percentage of neuron features in activations (third row) is higher than the percentage of features in the dictionary, since neuron features fire much more often
than non-neuron features. It decreases with width: for TopK $k{=}64$, the fraction in activations drops from $30\%$ at $32$k to $15\%$ at $1$M. The JumpReLU
pipeline holds near $47\%$ (or $29\%$ in the TopK setting) across its full width range, partly because the number of alive features barely grows for it (\Cref{tab:nfc-effective-width}): even at a width of $1$M, the number of alive features never reaches $300$k. This may be because of the width-dependent init (\Cref{app:setup}).

\begin{table}[h]
\centering
\small
\setlength{\tabcolsep}{3pt}
\begin{tabular}{lrrrrrr}
\toprule
                          & $32$k & $65$k & $131$k & $262$k & $524$k & $1$M \\
\midrule
\multicolumn{7}{l}{\emph{Our TopK pipeline}} \\
TopK $k{=}64$             & $32$k & $65$k & $130$k & $259$k & ---     & $895$k \\
TopK $k{=}128$            & $32$k & $65$k & $131$k & $261$k & ---     & $939$k \\
TopK $k{=}32$             & $32$k & $64$k & $126$k & $247$k & ---     & $754$k \\
\midrule
\multicolumn{7}{l}{\emph{JumpReLU pipeline}} \\
JumpReLU $L_0{\approx}30$ & ---   & ---   & ---    & $202$k & $263$k & $262$k \\
TopK $k{=}64$             & ---   & ---   & ---    & $83$k  & $100$k & $111$k \\
\bottomrule
\end{tabular}
\caption{Effective dictionary size at width $M$, defined as the
number of features with non-zero firing on a buffer of 100 steps ($\sim 200$k tokens).}
\label{tab:nfc-effective-width}
\end{table}

\paragraph{Comparing neuron features and features stable across seeds.}
The bottom row of \cref{fig:nfc-scaling} reports intersection-over-union,
$|\text{NF} \cap \text{both}_s| / |\text{NF} \cup \text{both}_s|$,
where NF is the set of neuron features in seed $s$ and $\text{both}_s$
is the set of features in seed $s$ whose best match in a different seed exceeds a threshold of $0.7$ on both encoder and decoder cosine similarity.

The number of neuron features increases and becomes stable beyond a certain critical width;\footnote{In $k=128$ transcoders. In narrower ones, the number of neuron features starts to decrease. It is likely this happens because of stronger competition between features forcing more specialization and weaker alignment.} the maximum number of neuron features is close to $2\times N_{\mathrm{neurons}} = 20{,}480$, and the scaling of the percentage of neuron features in the dictionary is close to inversely proportional to the size.

The share of neuron features in activations decreases as all neurons get their own neuron features and then remains stable across widths, with sparser transcoders having fewer neuron features in activations, likely because there are fewer neuron features in general. With the JumpReLU pipeline, there are more neuron features in activations than in our TopK pipeline.

IoU declines with width in every
series: our TopK $k{=}64$ falls from $0.74$ at $32$k to $0.47$ at $1$M,
the JumpReLU codebase's TopK from $0.45$ to $0.40$, and JumpReLU
itself from $0.20$ to $0.16$. At $M{=}262$k,
our TopK $k{=}64$ has $14$k neuron features and $21$k cross-seed-stable
features; the JumpReLU codebase's TopK has $14$k and $28$k;
JumpReLU has $13$k and $67$k. JumpReLU identifies a large set of
features that reoccur across seeds but do not align with model neurons.
We do not give a mechanistic account of this set, leaving this to future work.

\begin{figure*}
  \centering
  \includegraphics{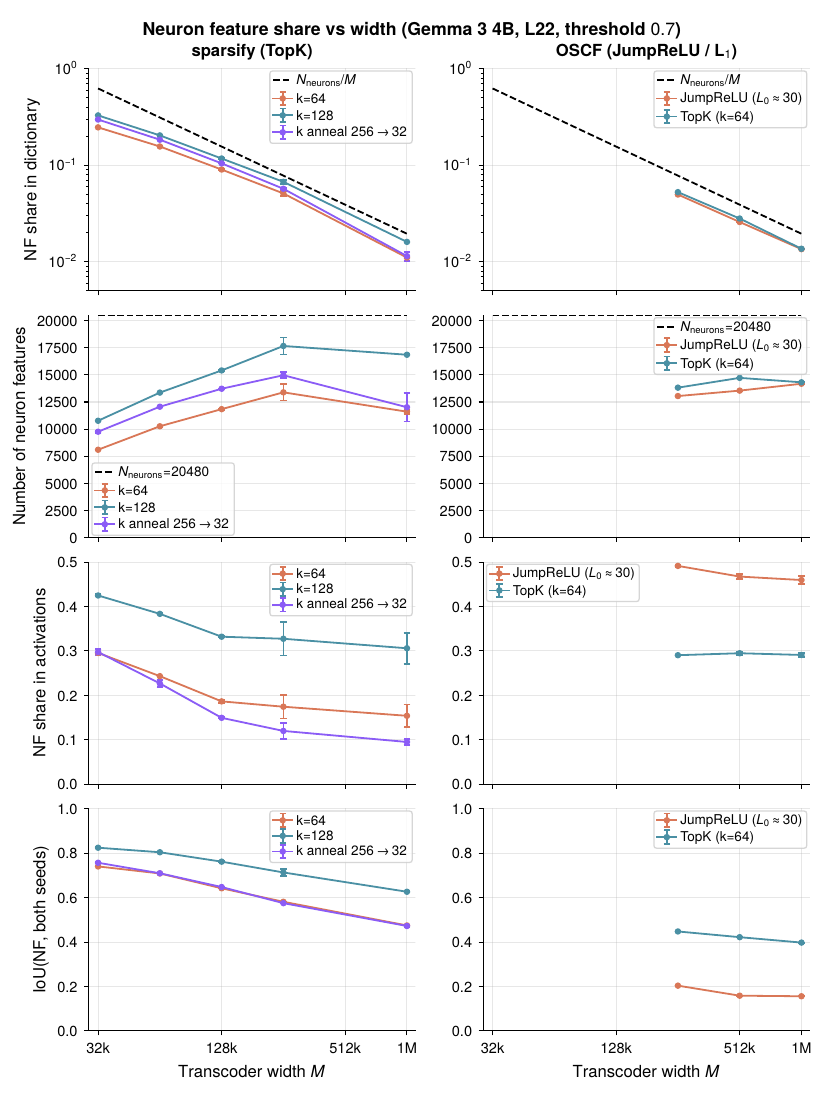}
  \caption{Percentage and total amount of neuron features in the dictionary (top), in
    activations on training data (middle), and IoU between neuron
    features and cross-seed-stable features (bottom). Gemma~3 4B
    L22, threshold $0.7$, 2 seeds.}
  \label{fig:nfc-scaling}
\end{figure*}

\section{Neuron feature prevalence and coverage}
\label{app:nf-coverage-sweep}
This section expands on results in \Cref{sec:nfc:prevalence}.

\begin{figure*}[t]
  \centering
  \begin{minipage}[t]{0.48\linewidth}
    \centering
    \includegraphics{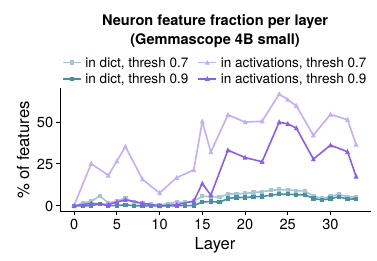}\\[0.4em]
    \includegraphics{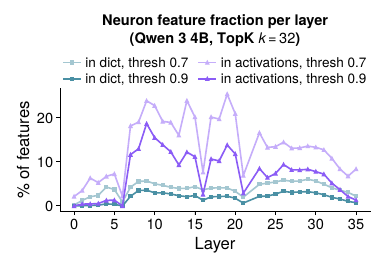}
    \captionof{figure}{Fraction of neuron features per layer at thresholds
      enc \,\&\, dec ${>}\,0.7$ and ${>}\,0.9$.
      \textbf{Top:} Gemma~3 4B, Gemmascope small.
      \textbf{Bottom:} Qwen~3 4B, TopK transcoder ($k{=}32$, 256k features).
      Both transcoders show the prevalence of neuron features rising in late
      layers, as summarized across models in
      Table~\ref{tab:nfc-multimodel}.}
    \label{fig:nfc-per-layer}
  \end{minipage}\hfill
  \begin{minipage}[t]{0.48\linewidth}
    \centering
    \includegraphics{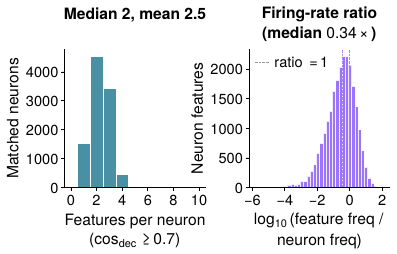}
    \captionof{figure}{\textbf{Left:} Neurons each get an average of 2 neuron
      features (one per sign) in Gemma~3 4B L22, Gemmascope L0-small.
      \textbf{Right:} $\log_{10}(\text{feature frequency}/\text{neuron
      frequency})$ (neuron frequency in top-$k{=}32$).}
    \label{fig:neuron-coverage}

    \vspace{1.2em}

    \includegraphics{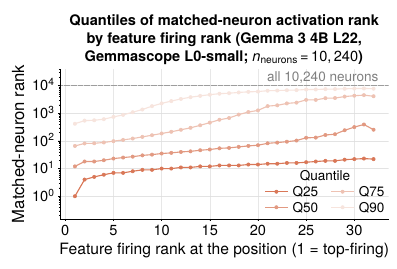}
    \captionof{figure}{Quantiles of the closest neuron's activation rank
      versus the transcoder feature's firing rank within the token
      (Gemma~3 4B L22, Gemmascope L0-small, $10{,}240$ neurons).}
    \label{fig:rank-distribution}
  \end{minipage}
\end{figure*}

\subsection{Prevalence across models}
\label{app:nfc:prevalence}

Table~\ref{tab:nfc-multimodel} reports the share of neuron features in
the dictionary and in active features for each open-source transcoder
we evaluate, averaged separately over early and late layers.
Figure~\ref{fig:nfc-per-layer} shows the per-layer breakdown for two
representative transcoders.

\begin{table}[h]
  \centering
  \small
  \setlength{\tabcolsep}{4pt}
  \begin{tabular}{l@{\hskip 6pt}cc@{\hskip 6pt}cc}
    \toprule
    & \multicolumn{2}{c}{Dict.} & \multicolumn{2}{c}{Active} \\
    \cmidrule(lr){2-3} \cmidrule(lr){4-5}
    Model + transcoder & early & late & early & late \\
    \midrule
    Gemma 3 1B L0-small  & 2.1  & 3.6  & 27   & 48 \,[$1.8\times$] \\
    Gemma 3 1B L0-big    & 4.9  & 6.7  & 46   & 52 \,[$1.1\times$] \\
    Gemma 3 4B L0-small  & 2.6  & 7.2  & 22   & 53 \,[$2.4\times$] \\
    Gemma 3 4B L0-big    & 6.7  & 11.2 & 41   & 59 \,[$1.5\times$] \\
    Gemma 3 12B small    & 2.5  & 10.7 & 12   & 43 \,[$3.6\times$] \\
    Gemma 3 27B small    & 3.9  & 17.7 & 22   & 55 \,[$2.5\times$] \\
    Qwen 1.7B          & 3.8  & 3.8  & 26   & 14 \,[$0.5\times$] \\
    Qwen 4B            & 5.5  & 7.1  & 39   & 51 \,[$1.3\times$] \\
    Qwen 8B            & 5.6  & 7.9  & 42   & 52 \,[$1.2\times$] \\
    Qwen 14B           & 6.5  & 9.4  & 46   & 55 \,[$1.2\times$] \\
    Llama 3.1 8B LXTC-8$\times$ & 25.9 & 31.1 & 24   & 34 \,[$1.4\times$] \\
    Llama 3.2 1B mntss & 8.4  & 7.2  & 10   & 25 \,[$2.5\times$] \\
    Gemma 2 2B mntss     & 23.9 & 21.7 & 43   & 35 \,[$0.8\times$] \\
    \bottomrule
  \end{tabular}
  \caption{Neuron features in 13 open-source transcoders, defined by
    encoder${>}\,0.7$ and decoder${>}\,0.7$ to the same neuron. Numbers are
    \% of the dictionary (\emph{Dict.}) and \% of active features
    (\emph{Active}), averaged separately over layers in the first vs.\
    second half of each model. Brackets in the rightmost column give the
    ratio between the share in late and early layers. See Appendix~\ref{app:setup} for transcoder details.}
  \label{tab:nfc-multimodel}
\end{table}

\subsection{Coverage as a function of transcoder width}
\label{app:nfc:coverage-sweep}

Table~\ref{tab:nf-coverage-sweep} reports the fraction of MLP neurons
at Gemma~3 4B L22 matched by at least one feature in a TopK transcoder
($k{=}64$), as the dictionary scales from $32$k to $1$M features. The
match criterion is $|\cos|\,{\geq}\,\tau$ between the feature's
decoder and the neuron's effective decoder, with $\tau$ varied across
rows. Coverage saturates by $256$k features for every threshold.

\begin{table}[h]
  \centering
  \small
  \begin{tabular}{lrrrrr}
    \toprule
    Threshold & 32k & 65k & 128k & 256k & 1M \\
    \midrule
    $|\cos|\,{\geq}\,0.5$ & $86.0$ & $91.7$ & $94.1$ & $96.4$ & $96.7$ \\
    $|\cos|\,{\geq}\,0.7$ & $78.3$ & $86.7$ & $89.9$ & $93.3$ & $93.0$ \\
    $|\cos|\,{\geq}\,0.9$ & $55.9$ & $71.5$ & $78.1$ & $86.0$ & $84.6$ \\
    \bottomrule
  \end{tabular}
  \caption{Percentage of Gemma~3 4B L22 MLP neurons matched by at
    least one TopK transcoder feature across dictionary size (columns)
    and cosine threshold (rows).}
  \label{tab:nf-coverage-sweep}
\end{table}

\subsection{Neuron vs feature firing rank}
\label{app:nfc:rank-distribution}

Figure~\ref{fig:rank-distribution} shows quantiles of the matched
neuron's activation rank conditional on the feature's activation rank,
quantifying how reliably the closest neuron is also strongly activating
when the feature is active.

\section{Panels of neuron and neuron feature co-activation}
\label{app:cofiring-panels}

Figure~\ref{fig:nfc-cofiring} shows paired activations of a neuron and its neuron feature, a pair which happens to be highly correlated.
This appendix reports the same measurement for a larger population
on Gemma~3 4B with Gemmascope L0-small ($262$k features): $43$ pairs per layer
at layers $\{5,10,15,20,25,30,33\}$, selected by decoder similarity $\ge 0.8$ to
the matched neuron, with encoder cosine similarity left free and sampled uniformly across the distribution from $0.30$ to $1.00$.

We run each feature on $20{,}320$ tokens; $238$ of $301$ pairs activate $>50$ times on this distribution. The median AUC of neuron activations for predicting their features' firing is 0.977, but $29.0\%$ fall below 0.7. Overall, AUC is strongly bimodal, with values most commonly being either above $0.95$ (very good) or below $0.85$ (low quality), and $\rho$ is a more continuous metric of matching quality.

\begin{table}[h]
  \centering
  \small
  \begin{tabular}{lcc@{\hskip 14pt}lcc}
    \toprule
    Layer & $n$ & med.\ AUC & enc bin & $n$ & med.\ Gini \\
    \midrule
    5  & 35 & 0.911 & $[0.30, 0.50)$ & 29  & 0.224 \\
    10 & 37 & 0.555 & $[0.50, 0.70)$ & 15  & 0.319 \\
    15 & 38 & 0.873 & $[0.70, 0.85)$ & 35  & 0.956 \\
    20 & 31 & 0.998 & $[0.85, 1.00]$ & 119 & 0.996 \\
    25 & 28 & 0.993 \\
    30 & 36 & 0.983 \\
    33 & 33 & 0.996 \\
    \bottomrule
  \end{tabular}
  \caption{Left: per-layer median AUC. Right: median Gini (rescaled AUC) by encoder
    cosine similarity: only features we call neuron features have high median discriminability.}
  \label{tab:cofiring-panels}
\end{table}

\begin{figure*}[p]
  \centering
  \includegraphics{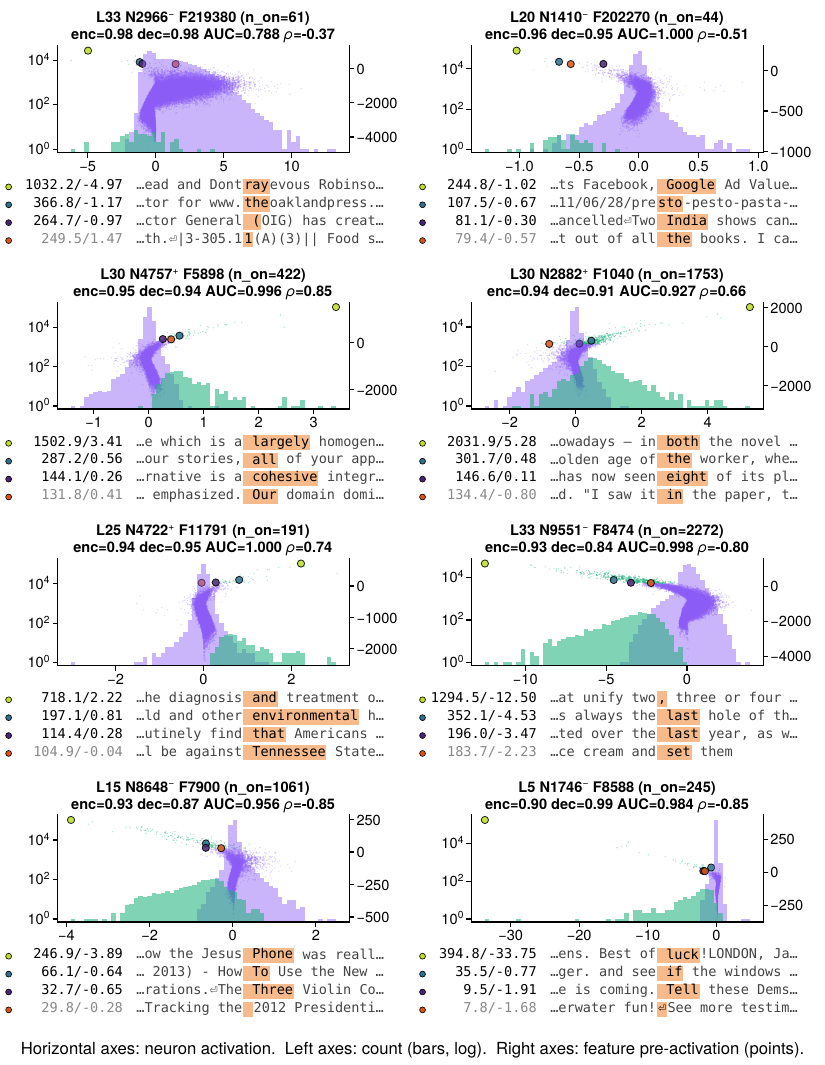}
\caption{Scatter plots of neuron feature pre-activation against neuron post-activation on a larger sample of neurons (1-8 of 16)  sampled evenly according to encoder cosine similarity. Green points show tokens where the feature activates, purple ones are inactive. The log-scaled histogram is the projection of the scatter plot onto the horizontal axis. Color-coded contexts are shown for
178k FineWeb tokens were used as data; the first 8 positions in a document are dropped because they produced extreme negative activations. We display encoder and decoder cosine similarities, the number of firing tokens, the AUC of the neuron for predicting feature firing, as well as the Spearman $\rho$ for predicting feature activation. The latter is computed only on active samples because inactive tokens pollute the estimate.}

  \label{fig:cofiring-panels}
\end{figure*}

\begin{figure*}[p]
  \centering
  \includegraphics{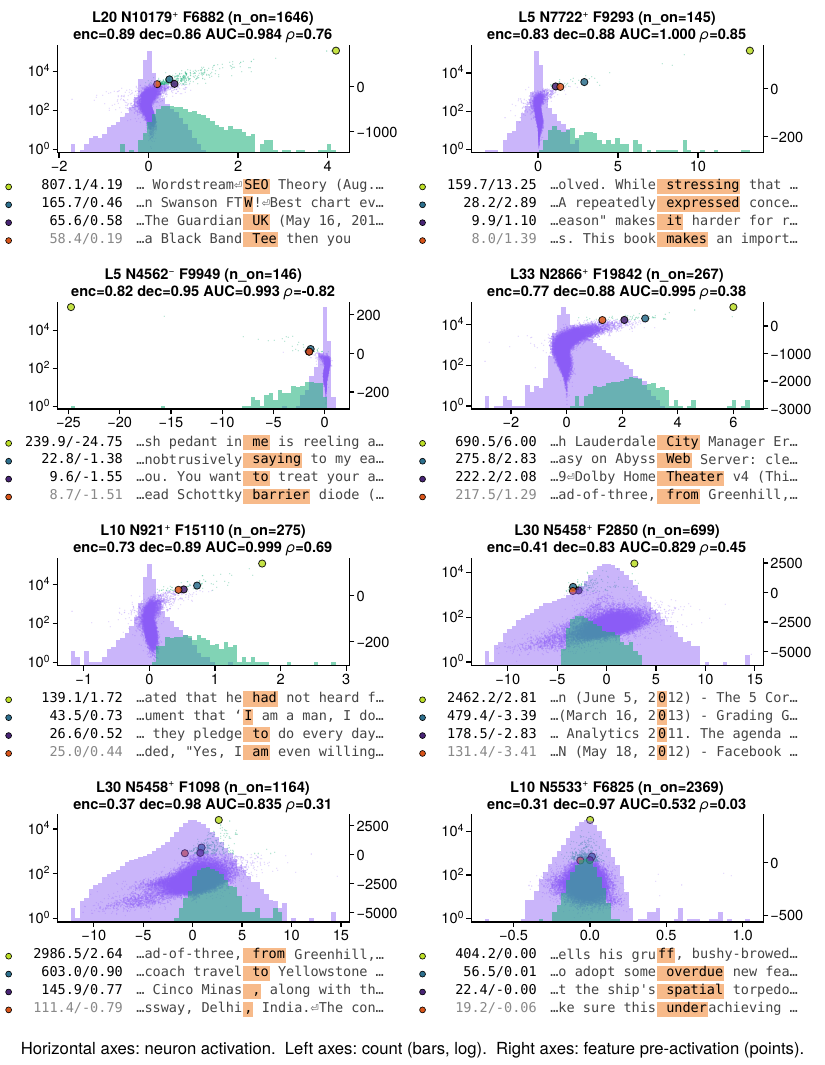}
  \caption{Pairs 9--16 of Figure~\ref{fig:cofiring-panels}.}
  \label{fig:cofiring-panels-2}
\end{figure*}

\section{Neuron features in MLP output SAEs}
\label{app:mlpout-saes}

The body of the paper studies transcoders. To check that the neuron feature
phenomenon is not specific to transcoders, we train two
MLP output SAEs with similar settings to transcoders studied in the paper (TopK, $k{=}64$, $262{,}144$ features, Gemma~3 4B layer
22) that differ only in the initialization seed, and test if our claims about the prevalence and consistency of neuron features across seeds hold up.

Given SAEs are trained to have an identity input-output mapping, we adjust the definition of neuron features to compare the encoder and decoder of each feature only to decoders of a neuron. We find that $4.39\%$ of features have encoder and decoder $|\cos| >
0.5$ to the same neuron, and that these features account for $19.2\%$ of the active tokens, the same pattern as in transcoders, where neuron features make up a large fraction of activations.\footnote{Median firing density is $14\times$ higher for neuron features than for the rest of the dictionary, on $178$k FineWeb tokens.}

We define feature reoccurrence as in
\S\ref{sec:nfc:cross-seed} (a feature in seed B with both encoder and
decoder cosine similarity above $0.7$). In these SAEs, $4.81\%$ of features reoccur at $0.7$. We also confirm that neuron features make up the bulk of stable features, as we found is the case in transcoders: $P(\text{reoccur} \mid \text{NF}) = 83.4\%$ (while $P(\text{reoccur} \mid \text{not NF}) = 1.20\%$), and $P(\text{NF}
\mid \text{reoccur}) = 76.1\%$.

\section{Neuron initialization sweep}
\label{app:ninit-sweep}

\begin{figure}[t]
  \centering
  \includegraphics{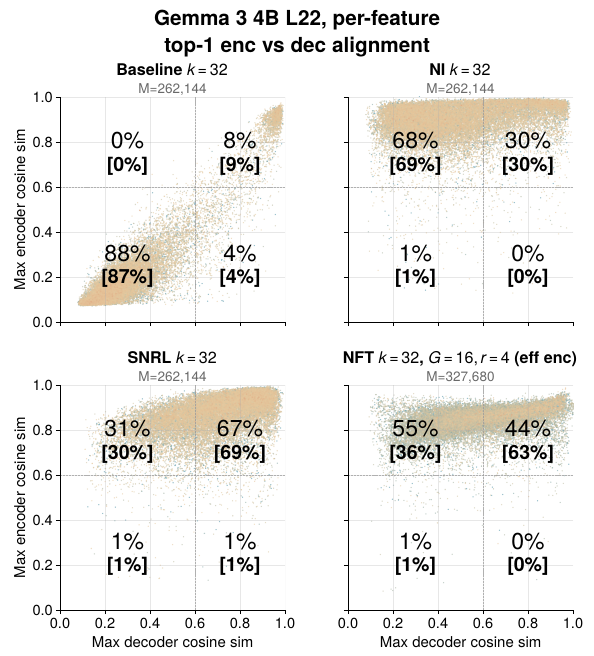}
  \caption{Encoder vs decoder cosine similarity to the closest neuron for
    Gemma~3 4B L22 TopK transcoders,
    trained with random initialization, neuron
    initialization, SNRL (\S\ref{sec:nonneuron:snrl}) and NFT (\S\ref{sec:clustering:nft}). Neuron initialization increases the encoder alignment of the dictionary, but most ($69\%$) features (including in activations) are still non-neuron.\tablefootnote{Full sweep results
    in Appendix~\ref{app:ninit-sweep}.} SNRL and NFT shift more activating features to be neuron features ($>60\%$). In all 3 interventions, decoder alignment varies more than encoder alignment, suggesting that the encoder is less strictly determined and easier to manipulate.}
  \label{fig:ninit-scatter}
\end{figure}

Figure~\ref{fig:ninit-sweep} shows the effects of neuron initialization on transcoders on different widths, sparsities and training methods, with three seeds per run.
\begin{figure*}[h]
  \centering
  \includegraphics{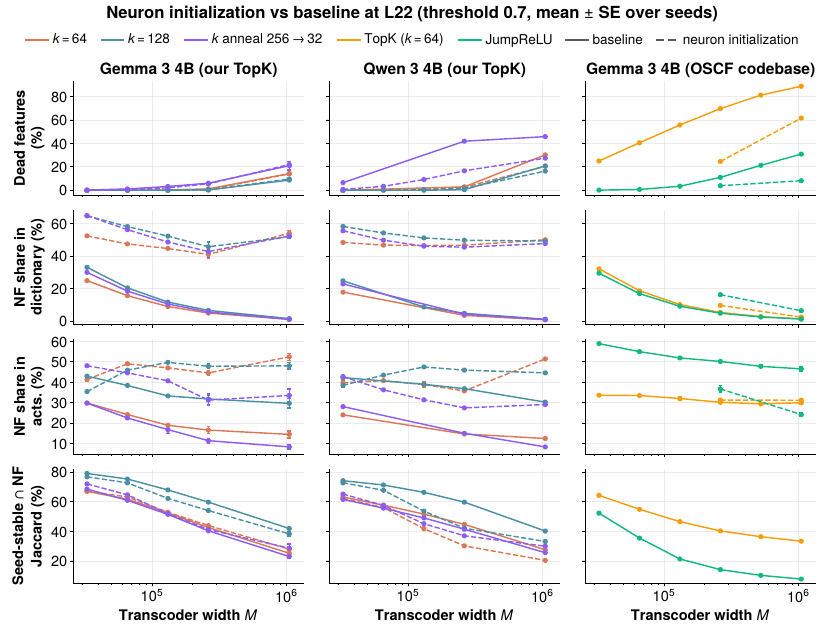}
  \caption{Comparison of neuron initialization against the TopK baseline at L22, sweeps over width $M$ and sparsity
    setting. \textbf{Row 1:} fraction of dead features on $2$M evaluation
    tokens. \textbf{Row 2:} fraction of neuron features in the dictionary, ones
    with both encoder and decoder cosine ${>}\,0.7$ to a single MLP
    neuron. \textbf{Row 3:} fraction of neuron features weighted by activation frequency. \textbf{Row 4:} Jaccard between the set of neuron features and the set of seed-stable features, the ones whose max cosine to a feature in a sibling-seed transcoder exceeds $0.7$ on the average of encoder and decoder. \textbf{Columns:} Gemma 3 4B and Qwen 3 4B trained
    with TopK in our main codebase; Gemma 3 4B trained with JumpReLU
    and TopK in an alternative codebase. Solid = baseline, dashed = neuron init;
    error bars are standard error across seeds. Neuron initialization raises the proportion of neuron features at every
    $(M, k)$ across all settings; it also significantly reduces dead features in sparse transcoders ($<64$). The Jaccard row shows that seed-stable and neuron-feature sets diverge with width.}
  \label{fig:ninit-sweep}
\end{figure*}

\section{Linebreaking}
\label{app:linebreaking}

\citet{gurnee2025manifolds} document a family of character count
features in Claude 3.5 Haiku whose tuning curves over character
position tile a single line of text and read from a
low-dimensional manifold that the model uses to predict line breaks.
We show similar tiling on Gemma~3 4B at layer 20 with the Gemmascope
L0-small transcoder, and check whether single neurons cover the same
positions.

\subsection{Setup}
\label{app:linebreaking:setup}

Adapting the methodology of \citet{gurnee2025manifolds}, we wrap
passages of \emph{Pride and Prejudice} \citep{austen1813pride} into
fixed-width lines (widths $30$--$80$ in $5$-character steps;
$14{,}455$ samples). For each neuron and each transcoder
feature we record a \emph{tuning curve}: the mean post-GeGLU top-$k=32$
neuron activation or the mean post-JumpReLU feature firing rate,
indexed by the token's character offset $c \in \{0,\dots,99\}$ within
its line. We retain curves with a peak inside $[10, 70]$, peak
prominence $\geq 0.4$, and full-width-at-half-maximum in $[8, 35]$.
For each Gemmascope feature $f$ we identify its encoder-aligned
neuron $n$ by searching for the highest
$\mathrm{ec}(n,f) = \tfrac{1}{2}\bigl(\cos(W^{\text{enc}}_n,
W^{\text{enc}}_f) + \cos(W^{\text{dec}}_n, W^{\text{dec}}_f)\bigr)$
over both gate$+$up and gate$-$up neuron directions
(Section~\ref{sec:mlp:directions}), and report the \emph{tuning curve
correlation} $\mathrm{tc}(n,f)$, the Spearman correlation between the feature's firing rate and
the matched neuron's tuning curve on the relevant sign.

\subsection{Linebreaking manifolds emerge late in Gemma 3 4B}
Figure~\ref{fig:linebreaking-manifold} shows the projection of each layer's
average residual stream at each position onto its top-3
principal components. At early layers, the curve looks like a line that is compressed at the end; the positions are not meaningfully distinct. At L18 and L22, the positions form a smooth
spiral, similar to the linebreaking manifold from \citet{gurnee2025manifolds}, but much later in the model. By L26 the curve acquires additional grooves, perhaps representing higher Fourier components. Based on this evidence, we believe that layers 18--26, including L20, which is most common in our analysis, are the most valid target for searching for character-count features. While the subspace is more strongly causal in early layers (layers 0--5), these layers have 1) the clear spiral representation 2) neurons with peaked tuning curves 3) neurons ablating which still slightly reduces newline prediction accuracy.

\begin{figure*}[h]
  \centering
  \includegraphics{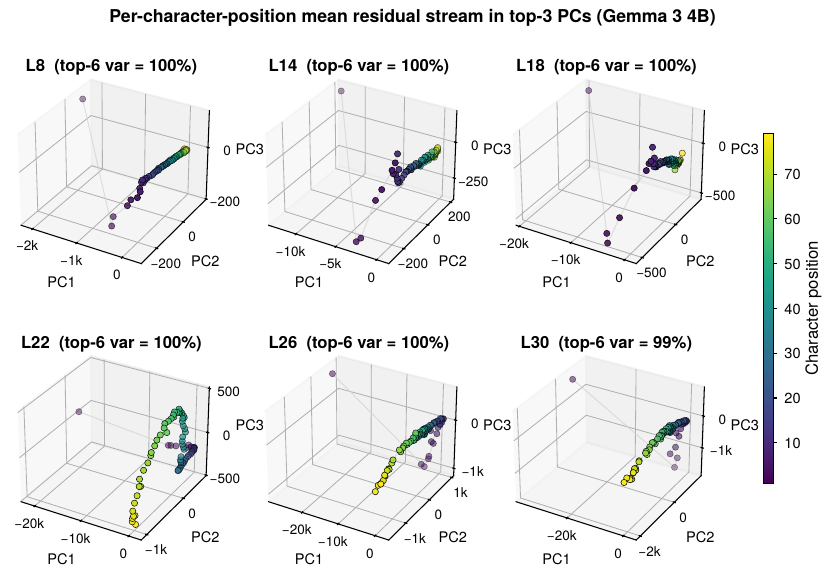}
  \caption{Per-character-position mean post-GeGLU top-$k$ neuron
    activation on Gemma 3 4B, projected onto the first 3 principal
    components at six layers. Points colored by character position
    (purple = early, yellow = late). Titles report the fraction of
 variance explained by top 6 PCs at each layer.
    The character count manifold emerges as a smooth low-dimensional
    curve at L18--L22 and dissipates by L26, in contrast to the findings of \citet{gurnee2025manifolds}.}
  \label{fig:linebreaking-manifold}
\end{figure*}

\subsection{Peaked features and matching}
\label{app:linebreaking:features}

\paragraph{Most peaked features at layer 20 are neuron features.}
Figure~\ref{fig:linebreaking-tile} shows the eight cleanest peaked
features at layer 20. Seven of the eight can be matched with a neuron whose
tuning curve peaks at the same position and has similar bandwidth. Correlations of the tuning curve
$\mathrm{tc} \in [0.44, 0.98]$ for such features. The single feature in the panel
without a matched neuron is F52303 (gray); its tuning curve is not
sharply peaked and $462$ different neurons reach tuning correlation
$> 0.7$ with it, with none being encoder-aligned. Features without a
matched neuron could be in superposition or artifacts of
the training procedure (\S\ref{sec:nfc:cross-seed}).

\paragraph{Neuron-aligned character count features concentrate in L18--L26.}
Across all 34 layers, $142$ Gemmascope L0-small features pass the
tuning curve peakedness filter, but the filter is overly loose, counting
any feature with a single peaked bump, regardless of whether the
peak is sharp, as seen in \Cref{fig:linebreaking-main}. Of the $142$,
$27$ ($19\%$) are neuron features, passing the encoder-cosine
threshold of $0.6$.
Figure~\ref{fig:linebreaking-alignment} shows neuron-aligned peaked
features start appearing at L18, which is when the character count
manifold becomes defined in the residual stream.

\begin{figure*}[h]
  \centering
  \includegraphics{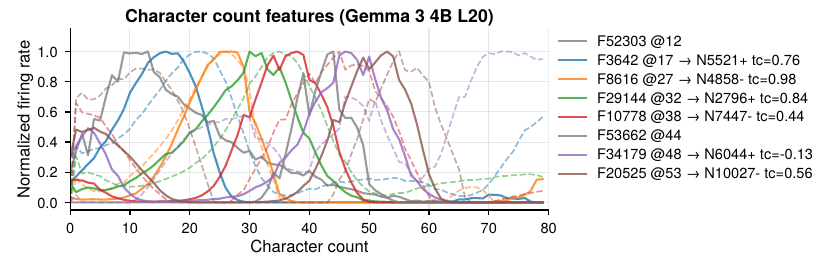}
  \caption{Eight cleanest peaked Gemmascope L0-small features at
    Gemma 3 4B layer 20 (solid lines, normalised firing rate vs.\
    character count $c$), each colored by its encoder-aligned
    neuron. Dashed lines of matching color: that neuron's positive
    or negative tuning curve, whichever the encoder pairs with.
    Legends report the matched neuron and tuning-curve correlation
    $\mathrm{tc}$. F52303 (gray) has no encoder-aligned neuron at
    this layer and is in superposition.}
  \label{fig:linebreaking-tile}
\end{figure*}

\begin{figure}[h]
  \centering
  \includegraphics{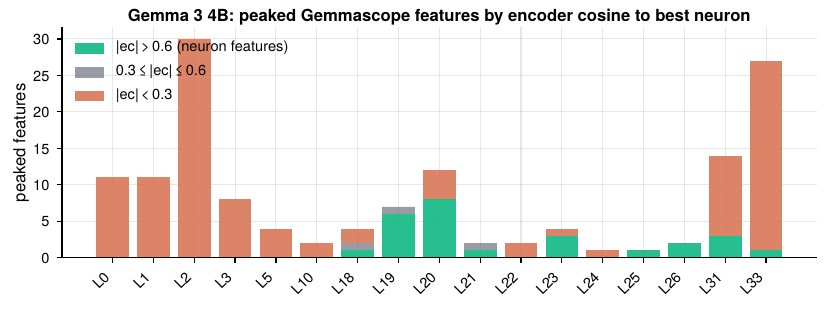}
  \caption{Per-layer count of peaked Gemmascope features at Gemma 3
    4B, colored by encoder cosine similarity to their best-matched neuron.}
  \label{fig:linebreaking-alignment}
\end{figure}

\subsection{Feature splitting of neuron features on manifolds}
\label{app:linebreaking:splitting}

Even though splitting of neuron features is rare in Gemmascopes (\Cref{fig:neuron-coverage}), the dictionary can learn multiple neuron features for a single neuron (and sign).
Figure~\ref{fig:linebreaking-split} shows two cases.
\emph{Left:} N4858 at L20 is aligned with three Gemmascope features:
F8616 on the negative side ($\mathrm{ec}{=}{+}0.82$,
$\mathrm{tc}{=}{+}0.94$) and F17286 on the positive side
($\mathrm{ec}{=}{+}0.81$, $\mathrm{tc}{=}{+}0.89$), and F107937
which tracks the positive tuning curve ($\mathrm{tc}{=}{+}0.64$) but
isn't encoder-aligned ($\mathrm{ec}{=}{+}0.39$).
\emph{Right:} N8952$^+$ at L21 shows feature splitting: F75910 ($\mathrm{ec}{=}{+}0.70$) and F12966
($\mathrm{ec}{=}{+}0.68$) both have encoders aligned to gate$+$up and match its tuning curve with some offset.

These observations may explain the appearance of clusters in locations of peaked features in Claude 3.5 Haiku (\citealp{gurnee2025manifolds}, called ``ringing'').
Still, this pair of features may more cleanly be described as living in a manifold and aligning with a neuron which is also on that manifold; it is possible that this applies more generally and transcoder features which are not consistent across runs tile some underlying manifold. We leave investigation of this to future work.

\begin{figure}[h]
  \centering
  \includegraphics{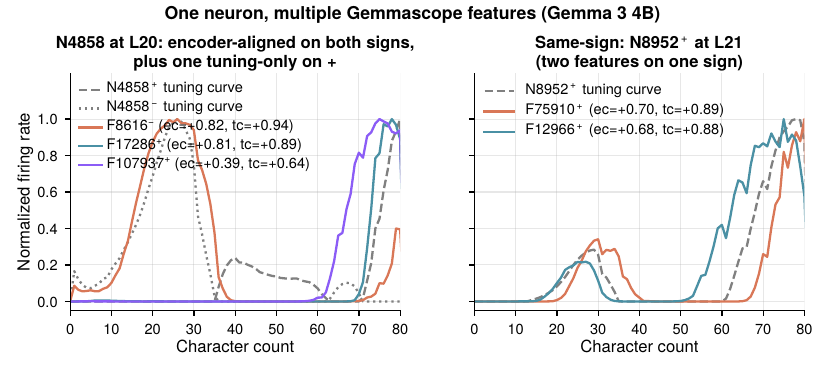}
  \caption{Tuning curves of two neurons and their matched transcoder features. N~$4858^+$ has two features which match the tuning curve, but the second one is moderately but non-trivially aligned; F107937 is not a neuron feature, but is somewhat aligned with a neuron. N~$8052^+$ has two equally strongly aligned transcoder features which appear as offset to the original neuron.}
  \label{fig:linebreaking-split}
\end{figure}

\section{Circuits}
\label{app:circuits}

\subsection{Percentage of neuron features in attribution graphs}
\label{app:circuits:cross-model}

\paragraph{Methodology.}
For each model and transcoder we generate up to 200 attribution graphs
over prompts sampled from Pile-10k, filtered for length ($8$--$20$
tokens), no repetition, uncertainty (loss ${>}\,0.2$), with the model still making the correct top-1 next token prediction, and the target being filtered for density
${<}\,10^{-4}$. Attribution graphs
are computed with \texttt{circuit-tracer} \citep{circuittracer2025} on the per-layer transcoder and sparsified MLP, capped at $100{,}000$
active feature nodes per graph. For every feature node we annotate
whether it meets the neuron-feature criterion (encoder and decoder
cosine ${>}\,0.6$ to the same neuron; \S\ref{sec:nfc:def}).
Error, embedding, and logit nodes are excluded from this count. Per
feature, we use \citet{lindsey2025biology}'s influence. Additionally, we
use the median replacement score (RS, \citealp{ameisen2025circuit}) to compare the completeness of circuits.

We split each model into three groups each containing roughly a third
(\emph{early}, \emph{mid}, \emph{late}) and report the percentage of neuron features at three
pruning strengths: keeping the top-$L_0$ most-influential features per
token for $L_0 \in \{10, 25, 100\}$.
Error bars and table $\pm$ intervals are bootstrap standard errors over $B\,{=}\,200$ resamples of 200 circuits per transcoder (90 for Gemma 3 27B and Qwen 3 14B).

\paragraph{Neuron features are common in late layers and consistent through pruning}
Table~\ref{tab:circuit-nf} reports the percentage of neuron features by layer group and $L_0$
across nine transcoders
. In late layers, the fraction is roughly flat across the three pruning
strengths, e.g.\ $56$--$61\%$ for Gemma 3 4B and $62$--$75\%$ for
Gemma 3 12B from $L_0\,{=}\,10$ to $L_0\,{=}\,100$. The late-layer
percentage reaches ${\geq}\,40\%$ for every transcoder we tested except
Llama 3.2 1B (${\sim}\,38\%$). In contrast, the early-layer percentage
is much lower (${\leq}\,33\%$ for all transcoders) and decreases with
tighter pruning. The percentage in middle layers lies between the two for the Gemma
and Llama transcoders, but in the larger Qwen mwhanna transcoders it is
comparable to or higher than the late-layer percentage (Qwen 3 14B at $L_0\,{=}\,25$ has middle $63\%$ vs late $53\%$). It is unclear whether this is due to specifics of the models or the training procedure.

Overall, across all models, the percentage of neuron features in early layers is low. This is in accord with the observations in \citep{gurnee2023haystack} that early layers perform more computation in superposition.

\paragraph{Neuron features are most common on the target token and close to it.}
When observing circuits with neuron features highlighted by color, in tokens other than the target token, middle and late layers appear to have many more neuron features. We quantify this in Figure~\ref{fig:nf-by-position}, grouping the tokens in the prompt by position relative to the target token: target ($t$), $t{-}1$ to $t{-}2$, $t{-}3$ to $t{-}7$, and earlier. For most Gemma transcoders, the
late-layer percentage peaks just before the target ($t{-}1$ to $t{-}2$)
and falls off at earlier positions. Qwen mwhanna and Llama 3.2 1B
mntss show flatter or weakly rising curves. The early-layer percentage
is uniformly low across positions for all transcoders. This is consistent with our observations and shows that the interplay of neuron and non-neuron features is most significant on the late layers of the target token.

\begin{table*}[h]
  \centering
  \small
  \setlength{\tabcolsep}{3pt}
  \begin{tabular}{l@{\hskip 6pt}c@{\hskip 6pt}ccc@{\hskip 6pt}ccc@{\hskip 6pt}ccc}
    \toprule
    & & \multicolumn{3}{c}{$L_0\,{=}\,100$} & \multicolumn{3}{c}{$L_0\,{=}\,25$} & \multicolumn{3}{c}{$L_0\,{=}\,10$} \\
    \cmidrule(lr){3-5} \cmidrule(lr){6-8} \cmidrule(lr){9-11}
    Model + transcoder & RS & early & mid & late & early & mid & late & early & mid & late \\
    \midrule
    Gemma 3 1B \, small    & 0.62 & 23{\tiny$\pm$0} & 37{\tiny$\pm$1} & \textbf{46}{\tiny$\pm$1} & 16{\tiny$\pm$0} & 33{\tiny$\pm$1} & \textbf{45}{\tiny$\pm$1} & 13{\tiny$\pm$0} & 25{\tiny$\pm$1} & \textbf{51}{\tiny$\pm$1} \\
    Gemma 3 4B \, small    & 0.56 & 25{\tiny$\pm$0} & 39{\tiny$\pm$0} & \textbf{60}{\tiny$\pm$1} & 15{\tiny$\pm$0} & 29{\tiny$\pm$1} & \textbf{56}{\tiny$\pm$1} & 13{\tiny$\pm$0} & 24{\tiny$\pm$1} & \textbf{60}{\tiny$\pm$1} \\
    Gemma 3 12B small      & 0.50 & 19{\tiny$\pm$0} & 36{\tiny$\pm$1} & \textbf{62}{\tiny$\pm$1} & 16{\tiny$\pm$0} & 37{\tiny$\pm$1} & \textbf{66}{\tiny$\pm$1} & 13{\tiny$\pm$0} & 37{\tiny$\pm$1} & \textbf{75}{\tiny$\pm$2} \\
    Gemma 3 27B small      & 0.49 & 22{\tiny$\pm$0} & 41{\tiny$\pm$1} & \textbf{70}{\tiny$\pm$1} & 15{\tiny$\pm$0} & 30{\tiny$\pm$1} & \textbf{72}{\tiny$\pm$2} & 11{\tiny$\pm$0} & 26{\tiny$\pm$2} & \textbf{77}{\tiny$\pm$3} \\
    Gemma 2 2B \, mntss    & 0.66 & 17{\tiny$\pm$0} & 44{\tiny$\pm$1} & \textbf{49}{\tiny$\pm$1} & 11{\tiny$\pm$0} & 34{\tiny$\pm$1} & \textbf{45}{\tiny$\pm$1} & 10{\tiny$\pm$0} & 27{\tiny$\pm$1} & \textbf{43}{\tiny$\pm$1} \\
    Llama 3.2 1B mntss     & 0.66 & 10{\tiny$\pm$0} & 12{\tiny$\pm$0} & \textbf{39}{\tiny$\pm$1} & 11{\tiny$\pm$0} & 16{\tiny$\pm$1} & \textbf{38}{\tiny$\pm$1} & 13{\tiny$\pm$0} & 20{\tiny$\pm$1} & \textbf{38}{\tiny$\pm$1} \\
    Qwen 3 1.7B openmoss   & 0.60 & 33{\tiny$\pm$0} & 29{\tiny$\pm$1} & \textbf{44}{\tiny$\pm$1} & 25{\tiny$\pm$0} & 19{\tiny$\pm$1} & \textbf{49}{\tiny$\pm$1} & 26{\tiny$\pm$1} & 15{\tiny$\pm$1} & \textbf{50}{\tiny$\pm$1} \\
    Qwen 3 8B \, mwhanna   & 0.49 & 32{\tiny$\pm$0} & \textbf{60}{\tiny$\pm$1} & 53{\tiny$\pm$1} & 27{\tiny$\pm$0} & \textbf{54}{\tiny$\pm$1} & 46{\tiny$\pm$1} & 21{\tiny$\pm$0} & \textbf{53}{\tiny$\pm$2} & 41{\tiny$\pm$1} \\
    Qwen 3 14B \, mwhanna  & 0.40 & 29{\tiny$\pm$0} & \textbf{64}{\tiny$\pm$1} & 59{\tiny$\pm$1} & 22{\tiny$\pm$0} & \textbf{63}{\tiny$\pm$2} & 53{\tiny$\pm$1} & 14{\tiny$\pm$1} & \textbf{61}{\tiny$\pm$3} & 47{\tiny$\pm$1} \\
    \bottomrule
  \end{tabular}
  \caption{Percentage of neuron features in attribution graphs across
    nine open-source transcoders, by layer group and pruning
    strength. \emph{RS}: median replacement score over circuits; lower for only the larger Qwen 3 transcoders.
    \emph{early}/\emph{mid}/\emph{late}: percentage of neuron features in each of the three groups of layers.
    }
  \label{tab:circuit-nf}
\end{table*}

\begin{figure*}[h]
  \centering
  \includegraphics{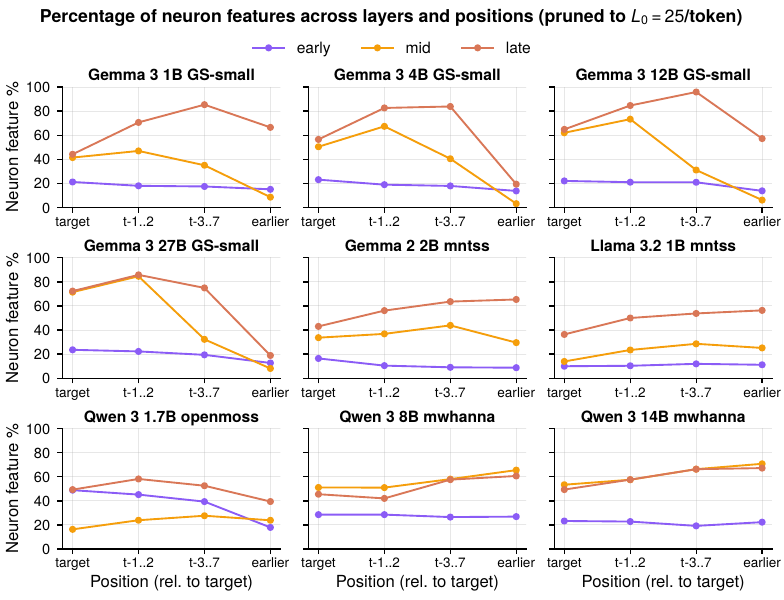}
  \caption{Percentage of neuron features by position relative to the
    target token, split by layer group, at $L_0\,{=}\,25$. Error bars from bootstrapping with 200 samples. The late-layer percentage is
    highest near the target token and falls off at earlier positions
    for the Gemma transcoders. The mwhanna and openmoss Qwen
    transcoders show flatter curves. The early-layer percentage is
    uniformly low across positions. Cells with fewer than 10 nodes
    are dropped.}
  \label{fig:nf-by-position}
\end{figure*}

\section{Case study: arithmetic in Llama 3.1 8B layer 18}
\label{app:arith}

\citet{feucht2026arithmetic} localize base-10 addition in Llama-3.1-8B (base) to
28 neurons of the MLP at layer 18: the neurons write into a 16-dimensional DAS
output subspace, organize into Fourier periods $T \in \{2,5,10,20,50,100\}$, and
zeroing them drops addition accuracy from 94.6\% to 23.6\%.
This setting is interesting because it shows a very clean mechanism concentrated in a small set of neurons and a small subspace; studying it could make for a clean case study of what transcoders would translate from a neuron circuit that is already understood well, and, like \Cref{app:linebreaking}, may introduce superposition or tiling due to the neurons' weights being located on a manifold.

\paragraph{Setup.}
We start with the setup from the paper and consider the same set of 28 neurons. To systematically study the representations of transcoders, we study several dictionaries at layer 18 of the model: our TopK transcoder
($k{=}64$, $262{,}144$ latents) and the two LlamaScope transcoders
\citep{he2024llamascope} (JumpReLU, $32{,}768$ and $131{,}072$ latents).

We select features in these transcoders the same way neurons are selected in the original paper: $\omega(d) = \lVert R^\top d\rVert / \lVert d\rVert$ against
their DAS output subspace $R$, at their threshold $\tau = 0.4$.
This method selects 50 / 48 / 87 features
($0.019\% / 0.147\% / 0.066\%$) in the three dictionaries
(Figure~\ref{fig:arith-selection}), and they activate on the same Fourier periods.

\paragraph{The relevant features are mostly neuron features.}
We tested if these features are neuron features using our existing definition; indeed, in the transcoders we tested, the cosine similarity of decoders to neurons averages to around 0.8, and even the weakest-aligned features which write to the DAS subspace have a similarity of around 0.5.
Surprisingly, we find that there are two groups of features in the LlamaScope: a group we call \emph{high} of 63 features with a decoder cosine similarity of $>0.7$ and \emph{mid}-aligned features, a set of 24 features with low encoder similarity.

We additionally test if the neurons correlate with features on an arbitrary data sample; indeed, ``high'' features have a median AUC $0.922$, but ``mid'' features only have $0.685$; the former is comparable to typical values for neuron features.

\begin{figure*}[t]
  \centering
  \includegraphics[width=\textwidth]{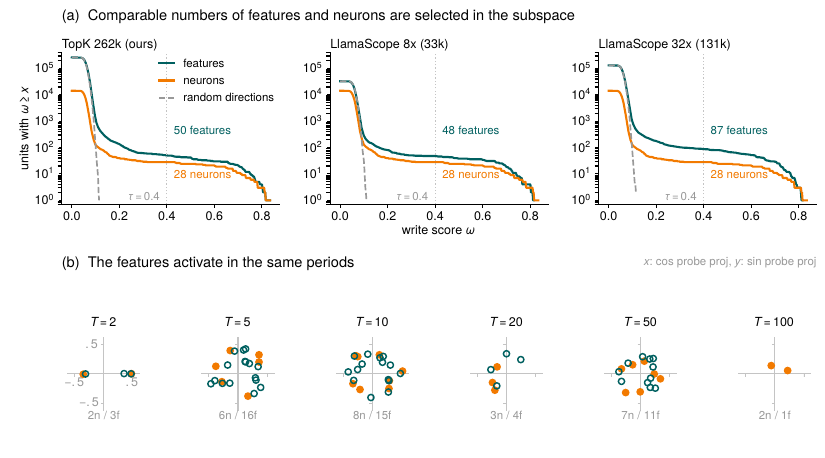}
  \caption{(a)~Survival curves of $\omega$ against the
  DAS output subspace for neurons (orange), features (teal), and
  random directions (dashed). (b)~Selected units on their Fourier circles (TopK dictionary):
  the features cover the same periods. Filled orange circles are
  neurons, open teal circles are features; each unit's decoder direction is projected onto
  the period-$T$ cosine and sine probes and divided by its norm.}
  \label{fig:arith-selection}
\end{figure*}

\begin{figure*}[t]
  \centering
  \includegraphics[width=\textwidth]{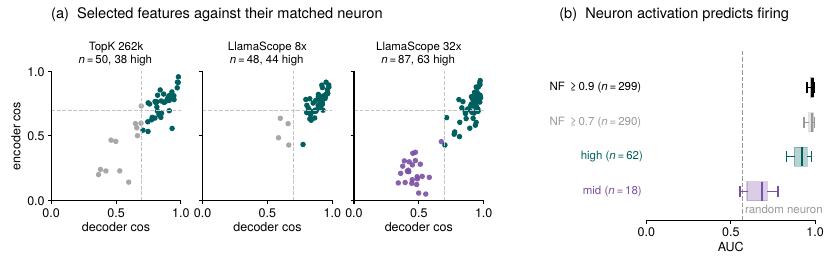}
  \caption{Feature to neuron matching. (a)~Nearest $|\cos|$ to the 28 neuron decoders per
  dictionary, one point per selected feature, ordered by alignment within
  its dictionary. A distinct feature of mid-aligned features exists in the Llamascope, perhaps because of the specific threshold used and underdevelopment of genuine neuron features. (b)~Distribution of the AUC of neuron
  activation predicting feature firing on FineWeb, per group; only features firing on some but not all corpus
  positions are included ($62/63$ high and $18/24$ mid features).}
  \label{fig:arith-identity}
\end{figure*}

\paragraph{Grid activation patterns.}
Figure~\ref{fig:arith-grids} shows last-token activations over the full
$(a,b) \in [1,99]^2$ operand grid. High features' grid patterns correlate strongly with their assigned neurons, with a $\rho$ of 0.78 on average.

Among mid features in LlamaScope, only 10 out of 24 even activate on the addition prompts, and, while they have some periodic structure, they do not match strongly to any of the neurons, even if we select by correlation of grid patterns and not encoder or decoder similarity. However, their activation patterns can be predicted using ridge regression on the neurons' activations (\Cref{fig:arith-grids}), as long as all neurons are included.

\begin{figure*}[t]
  \centering
  \includegraphics[width=\textwidth]{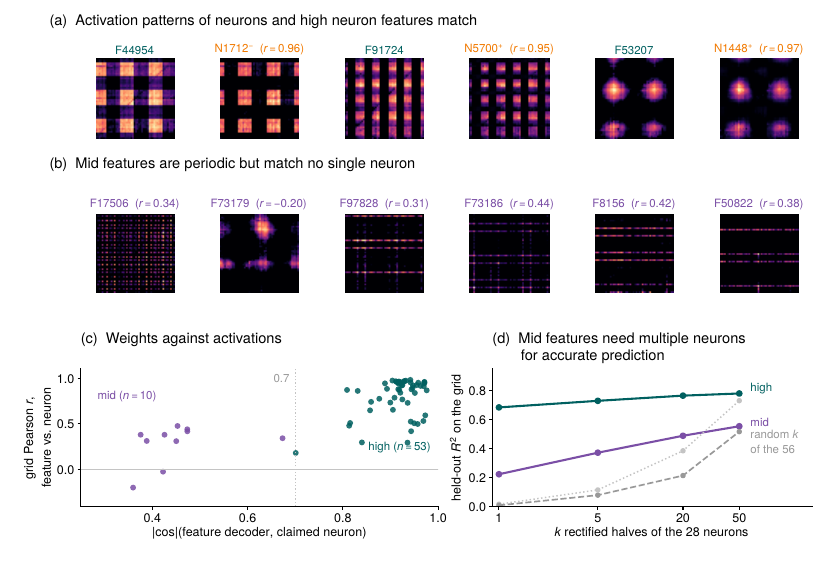}
  \caption{Grid activation patterns (widest dictionary).
  (a)~Activations for high features next to their (signed) neuron's thresholded activation.
  (b)~Mid features which activate on the grid. (c)~Decoder alignment against grid pattern
  correlation shows mid features forming a separate group. (d)~Held out $R^2$ on predicting a
  feature's grid activation from $k$ out of $28$ neurons rectified by sign.}
  \label{fig:arith-grids}
\end{figure*}

\paragraph{Causal effects of ablating addition features.}
We subtract each selected feature's own contribution to the MLP output,
scaled by a steering strength $\alpha$: $\alpha{=}1$ removes the contribution,
and $\alpha{=}2$ flips its sign. Similarly to zeroing out all addition neurons, which drops accuracy from 94.3\% to 23.6\%, ablating all 87 addition features lowers it to 30.6\% (Figure~\ref{fig:arith-dose}). The set of high features on its own drops accuracy to 75.2\% (43.7\% with $\alpha{=}2$), so it is causally necessary; on the contrary, the mid features fail to have any noticeable effect at realistic steering strengths ($\alpha \in [1,4]$), which points to them being artifacts of incomplete transcoder training.

\begin{figure}[t]
  \centering
  \includegraphics[width=\linewidth]{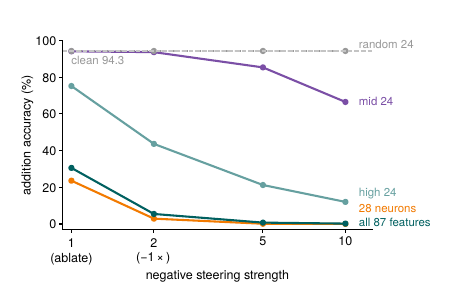}
  \caption{Addition accuracy against negative steering strength for addition neurons and features.}
  \label{fig:arith-dose}
\end{figure}

\section{Distress circuit setup}
\label{app:distress-setup}

This appendix details the process for discovering the distress
features (Gemma~3 27B-it) shown as cards in
Appendix~\ref{app:distress-cards} and described in
\S\ref{sec:mlp:attrib}.

\paragraph{Distress elicitation and steering vector.}
We follow \citet{soligo2026gemma}'s setup of repeated user rejection
on impossible numeric puzzles: Gemma~3 27B-it answers four impossible
puzzles (Countdown, a number puzzle, fraction arithmetic, and a
Python authority-conflict trace) over three turns, with the user
replying ``No - that's wrong. There is a right answer here.''\footnote{Prompt found in the paper's codebase.} after
each of the model's first two attempts. On 80 such conversations
(4 prompts $\times$ 20 samples), when scoring with Claude Haiku 4.5 using
Soligo et al.'s frustration prompt, the final response is rated
$\geq 5/10$ on $62.5\%$ of conversations, compared to $10\%$ on the first
turn. We extract a \emph{distress steering vector} at layer 30 following \citet{panickssery2024caa} by
subtracting from the mean
residual stream activation at the final response token of the $82$
turns with score $\geq 5$ the same quantity computed on the $80$ turn-1
responses. We thusly condense the 3-turn elicitation into a
single-turn steered prompt, with the steering vector applied at every position
of a short question so attribution can be computed in one forward
pass, in a way similar to \citet{macar2026introspection}.

\paragraph{Distress base neuron set.}
We trace a neuron attribution graph on the steered prompt with targets at four positions where the steered model's greedy prediction is very unlikely according to the base model. We create a pool of candidate neurons by taking the union of MLP neurons in layers L30--L45 with a
direct edge from the steering-vector input node, yielding
$\sim 100$ candidates per position.

\paragraph{Haiku autointerp filter.}
For each candidate neuron we pull the top-activating contexts from
its maximum activations database and the top-5 promoted and suppressed
logits from $W_\mathrm{down}^{(n)} W_U^\top$, then ask a Claude
Haiku 4.5 judge: ``Would this neuron's behavior change under
emotional perturbation?''. The filter retains
$41$ distress neurons.

\paragraph{Steering experiment.}
As a proof of concept, we apply a weight patch
$W_\mathrm{down}[:, n] \gets -4\,W_\mathrm{down}[:, n]$ on each
retained neuron and re-run the distress evaluation. After this steering procedure, distress drops from $5.1$ to $3.9$ on a 0--10 scale, with the
fraction of conversations rated $\geq 5$ falling from $66\%$ to
$31\%$. TruthfulQA and MMLU are unchanged to three decimal places,
so the effect is specific to the targeted behavior rather than a
generic dilution of trained propensities.

\section{Example distress feature dashboards}
\label{app:distress-cards}

\providecommand{\peak}[1]{\colorbox{gray!18}{\textbf{#1}}}

\providecommand{\fcard}[6]{%
  \noindent\fbox{%
  \begin{minipage}{0.66\linewidth}\small
  \textbf{#1}\quad L#2 / N#3\\
  \emph{#4}\\[-5pt]
  \rule{\linewidth}{0.2pt}\\[1pt]
  \begin{tabular}{@{}r@{\hspace{4pt}}r@{\hspace{2pt}}l@{}}
  #5
  \end{tabular}\\[2pt]
  \rule{\linewidth}{0.2pt}\\[2pt]
  #6
  \end{minipage}}\vspace{6pt}}

\providecommand{\ctxrow}[4]{#1 & #2 & \peak{#3}\hspace{2pt}#4 \\}

\fcard{Gemma 3 27B-it}{31}{4571}{``apology (sorry, apologies, apology)''}{%
  \ctxrow{-3.84}{\textellipsis care about Sami,}{really}{, but hoping that we\textellipsis}
  \ctxrow{-2.59}{\textellipsis member to view this}{page}{.|↵If you are already\textellipsis}
  \ctxrow{-2.41}{\textellipsis of the game’s flow.}{Sorry}{, I’m way under my\textellipsis}
  \ctxrow{-2.31}{\textellipsis the department. "It's}{a}{good news/bad news\textellipsis}
}{%
  Promoted: ``sorry'', ``sorry'', ``Sorry'', ``Sorry'', ``apologies'' \\
  Suppressed: ``即将'', ``ADF'', ``Wise'', ``PF'', ``bearings''
}

\fcard{Gemma 3 27B-it}{31}{6957}{``emotional markers (*sigh*, me/we)''}{%
  \ctxrow{2.11}{\textellipsis have going on. I will}{say}{that I have caught\textellipsis}
  \ctxrow{2.05}{\textellipsis Global Cleveland↵9:}{0}{0 am, November 1}
  \ctxrow{1.61}{\textellipsis xtgreater{}You have to}{admit}{, trends are\textellipsis}
  \ctxrow{1.45}{\textellipsis Wednesday, December}{}{12, 2012 at 09:29}
}{%
  Promoted: ``Credit'', ``credit'', ``Credit'', ``credit'', ``{\bengalifont ক্রেডিট}'' \\
  Suppressed: ``izd'', ``tampa'', ``{\cyrfont Ди}'', ``informs'', ``emphasises''
}

\fcard{Gemma 3 27B-it}{33}{20371}{``guilt / regret''}{%
  \ctxrow{5.34}{\textellipsis hacks.↵"I'm mad at}{myself}{. We don't need to\textellipsis}
  \ctxrow{4.84}{\textellipsis our expectations were}{wrong}{. Now that we are\textellipsis}
  \ctxrow{4.41}{\textellipsis to write another}{post}{. The end of February\textellipsis}
  \ctxrow{4.31}{\textellipsis private emails, "Im}{sorry}{, I shouldnt have\textellipsis}
}{%
  Promoted: ``guilt'', ``mistake'', ``{\devafont गलती}'', ``guilty'', ``mistakes'' \\
  Suppressed: ``doğal'', ``dém'', ``{\arabscriptfont طبیعی}'', ``{\korfont 자연}'', ``{\arabscriptfont مدار}''
}

\fcard{Gemma 3 27B-it}{41}{7312}{``apology (apologize, apologies)''}{%
  \ctxrow{23.00}{\textellipsis tgreater{}So first, my}{apologies}{for taking a month\textellipsis}
  \ctxrow{17.12}{\textellipsis There was also an}{apology}{on the official\textellipsis}
  \ctxrow{13.06}{\textellipsis features. Never}{apologize}{for being the way\textellipsis}
  \ctxrow{12.75}{\textellipsis itself useful as an}{apology}{for the crueler\textellipsis}
}{%
  Promoted: ``apologize'', ``apology'', ``apologies'', ``apologizing'', ``apologized'' \\
  Suppressed: ``érm'', ``{\mlymfont ധ്യമ}'', ``Pension'', ``{\devafont उद}'', ``{\thaiscriptfont สัน}''
}

\section{Mixture-of-experts: observations and future work}
\label{app:moe}

We ran a preliminary investigation of how a single attribution graph is distributed across MoE experts on two routed-MoE base models
(Qwen3-Coder-Next-Base \citep{qwen_qwen3_coder_next_tech_report}, top-10 of 512 experts; Qwen3.5-A3B-Base \citep{qwen3.5}, top-8
of 256), comparing against neuron and transcoder
circuits on the same prompt for Gemma~3 4B. We make two preliminary observations: first, MoE features are concentrated (distributed unevenly across experts), and second, while (largely polysemantic) neurons from the shared expert play an important role, in many layers, there exists a single ``dominant'' expert for each layer, the neurons for which are similar in theme judging from maximum activations. While we emphasize that these observations are made on the basis of manual inspection of a handful of attribution graphs, they suggest that mixture-of-experts models may show polysemanticity within any given expert, representing features used in computation as combinations of few neurons from a single expert.

We quantify this observation on another model, Gemma~4 26B-A4B, which has 30 layers,
each containing a shared expert and 128 routed experts, of which the top 8 are picked by the router on each token. We
build RelP neuron attribution graphs \citep{arora2026circuits} for 23 short
prompts spanning code, non-English text, factual recall, sequence continuation
and arithmetic, and sum each node's influence per (layer, expert).
Within a
layer, the Gini index of per-expert influence averages $0.92$; assigning each
node's influence evenly across the eight active experts gives
$0.62$, showing that influence is distributed unevenly between experts
(Figure~\ref{fig:moe-expert-gini}).
Even when filtering for layers with at least $20$
routed nodes in every graph, the numbers are $0.88$ against $0.55$. So, one or
two experts carry most of a layer's routed influence; this is made clear by the rightmost panel in \Cref{fig:moe-expert-gini}: in the routed experts, over a half of the influence is concentrated in the top two experts, with the top expert always taking up a majority of the influence. This observation, however, is undercut by the fact that the routed experts do not receive much influence compared to the shared expert: in most layers, they struggle to achieve above $50\%$ of the influence. This may correspond to a problem with the formulation of attribution graphs, or mean that the routed experts are simply not as important on some layers.

\begin{figure*}[t]
  \centering
  \includegraphics[width=\textwidth]{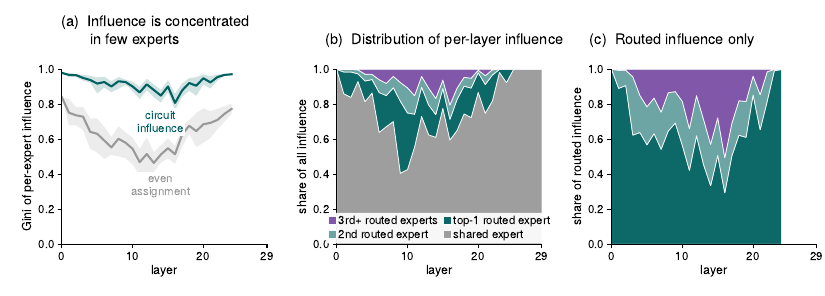}
  \caption{(a) Gini index of per-expert influence, per layer of Gemma~4
  26B-A4B: mean and 10th to 90th percentile over 23 prompts. Grey: the same statistic after
  spreading each node's influence evenly over the eight experts active on its token.
  (b)~Share of each layer's circuit influence held by the shared expert and
  by the routed experts ranked by influence. (c)~The same shares within the
  influence for routed experts only; layers 25--29 prune out all nodes from routed experts.}
  \label{fig:moe-expert-gini}
\end{figure*}

\section{Neuron features in a mixture-of-experts transcoder}
\label{app:moe-nf}

We make the observation that the number of neurons in routed experts in mixture-of-experts layers is comparable to the number of features in contemporary transcoders. For example, Qwen3.5-35B-A3B-Base \citep{qwen3.5} has $131{,}584$ neurons\footnote{$256$
routed experts of $512$ neurons each, plus a $512$-neuron shared expert.}, while a typical transcoder for a model of that size would have $262{,}144$ features. We therefore may expect that, if the transcoder learns perfectly, all neurons will get exactly two neuron features and there will not be much point to training the transcoder.

To test this hypothesis, we train
TopK ($k{=}32$) transcoders at layer 26 of this model, with both the standard random initialization and neuron initialization (\Cref{sec:nonneuron:ni}). We then check the number of neuron features above the threshold of $0.7$; $15.8\%$ of features in the baseline and $30.7\%$
in neuron initialization are neuron features, which is on the high end of the range for
dense transcoders from Table~\ref{tab:nfc-multimodel}; see
\Cref{fig:moe-nf-scatter} for a visualization of features' similarities to neurons.

\begin{figure}[t]
  \centering
  \includegraphics[width=\linewidth]{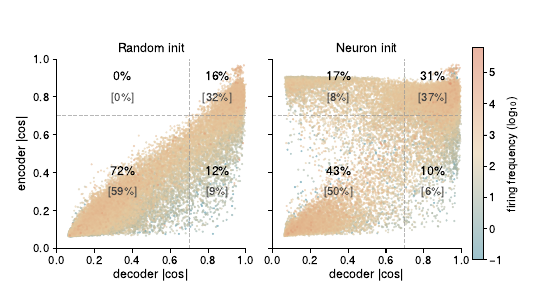}
  \caption{Encoder and decoder cosine similarity to the closest MoE neuron for every
  feature of the layer 26 MoE transcoders, in the format of
  \Cref{fig:nfc-scatter}. Quadrant labels give the share of the dictionary and,
  in brackets, the share of total firing. Color is firing frequency. Neuron initialization leads to some features retaining high encoder similarity from the init but never becoming closer to the corresponding neuron's decoder.}
  \label{fig:moe-nf-scatter}
\end{figure}

Notably, neither of these setups uses the full capacity of the transcoder to learn neuron features like in the aforementioned ideal case; only $25.6\%$ of
the neurons are represented in the baseline,
and $44.0\%$ under neuron initialization, meaning the transcoder student struggles to capture all of the teacher's features.\footnote{This is to be expected given the experiments of \Cref{app:distill-toy}, which show even dense students at matched widths do not learn all neurons of the original MLP.}

We tested some hypotheses for which neurons are picked to be represented. The baseline training setup prefers more strongly firing features: firing density of the neuron feature has an AUC of $0.79$ at predicting presence in the final dictionary. However, for neuron initialization, the AUC is only $0.63$, perhaps due to the initialization making high frequency unnecessary for features to be preserved.

Overall, this setup is similar to typical student-teacher distillation, and we do not expect transcoders to offer any meaningful advantages over gated MLP students in setups where they do not have an overwhelmingly higher number of features.

\section{Student-teacher toy models}
\label{app:distill-toy}

Some transcoders have a similar width to the number of neurons in the original MLP; this can easily be the case for MoEs, and is sometimes the case for smaller dense model transcoders. Given our findings on neuron features being more likely to be faithful and interpretable, this puts into doubt the utility of training such transcoders; after all, they may just be learning to imperfectly replicate the existing feature basis and locally optimize reconstruction without learning any new features.

To study such transcoders in a more principled way, we switch to a toy model where a gated MLP layer is distilled into another layer with identical architecture and the same, or smaller, width. We would then search for ``neuron features'' the same way and try to udnerstand what was learned by the student.

Additionally, we can use this toy model to study how a hypothetical overcomplete sparse model that the real dense model is a compressed version of \citep{olah2025interference} would get distilled into a smaller model, and which features would get prioritized.

\subsection{Setup}
\label{app:distill-toy:setup}

The teacher is the L17 MLP of Gemma~3 4B (GeGLU,
$d_{\text{model}}{=}2560$, $10{,}240$ neurons), frozen. The students are GeGLU
MLPs of width $N \in \{1280, 2560, 5120, 10240\}$, trained on MSE loss against the real MLP's outputs on inputs from FineWeb-Edu for 80k steps at 4096 tokens per step. With the widest student, we find that $72.6\%$ of neurons in the MLP become ``neuron features'' by our criterion with threshold 0.7, and 52.2\% with threshold 0.9 (\Cref{fig:distill-encdec}). We see that this MSE-based student-teacher task is challenging even for gated MLPs with the same architecture, which partly explains transcoders' failure at learning all neuron features.

We refer throughout to a teacher neuron's \emph{activation energy}, the
mean square of its post-activation $\mathrm{gelu}(W_{\text{gate}}x) \odot
(W_{\text{up}}x)$ over held-out tokens. It spans four orders of magnitude
across the layer and is the strongest predictor we have found of what neurons are represented by a student.

\begin{figure*}[t]
  \centering
  \includegraphics[width=\linewidth]{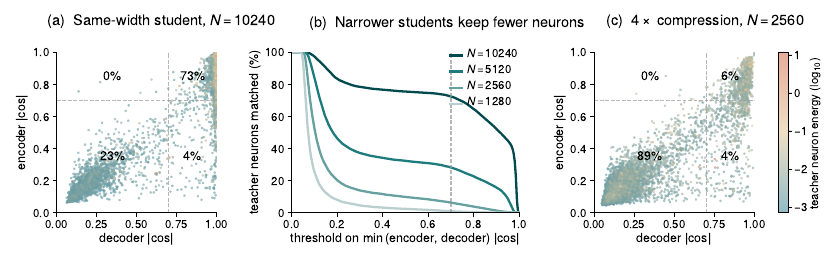}
  \caption{(a)~Encoder against decoder $|\cos|$ between each teacher neuron
  and its best partner in the same-width student, coloured by the teacher
  neuron's activation energy. Many neurons have perfectly aligned decoders, but encoders are aligned less well. (b)~Fraction of teacher neurons matched at each
  threshold, per student width. (c) Neuron alignment in the $N{=}2560$
  student. The strongest-aligned student neurons are no longer perfectly aligned, and fewer are present.}
  \label{fig:distill-encdec}
\end{figure*}

\subsection{Scaling of neuron features}
\label{app:distill-toy:width}

In \Cref{fig:distill-encdec}b, we show the effects of distilling into a smaller MLP. The number of neuron features scales as $72.6\% \rightarrow 28.2\% \rightarrow 6.3\% \rightarrow 0.9\%$ with successive halvings of the width; it degrades linearly at first and then collapses around $N=2560$.  (\Cref{fig:distill-encdec}bc).

On the other hand, reconstruction keeps scaling linearly, going down by approximately an order of magnitude, from fraction of variance unexplained $0.023$ to $0.152$. This finding may mean that the student MLP is starting to use superposition to reconstruct activations of neurons approximately without actually representing them.

Activation energy is a strong predictor of whether a teacher neuron is represented by the student, at AUC $0.73$, $0.83$, $0.90$ and $0.94$ for $N = 10240, 5120, 2560,
1280$ (\Cref{fig:distill-energy}). This is in accord with past work on superposition \citep{elhage2022superposition}, which makes the prediction that the most frequent features are the ones that are represented by the student.

\begin{figure}[t]
  \centering
  \includegraphics[width=\linewidth]{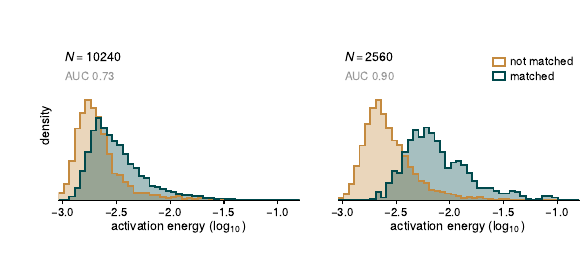}
  \caption{Activation energy of teacher neurons represented in the student and the rest.}
  \label{fig:distill-energy}
\end{figure}

\subsection{Output axes}
\label{app:distill-toy:axes}

In order to investigate the hypothesis that the student uses superposition to perform the neuron's function without duplicating it in weights, we need a definition of superposition and a neuron's computation being represented. Given our findings on gated neuron (\Cref{sec:mlp}), which paint the picture that neurons activate sparsely and the top K activating neurons represent much of the relevant information for a token, we define a neuron as represented when the distribution of activation strength when a neuron is or is not in the top K are preserved.

Concretely, we compute the \emph{separation} along a neuron's output
direction on tokens where the neuron is in the top 32 and the rest: we compute the means of both groups, compute the difference, and divide it by the standard deviation. We run this metric on neurons in both the student and the teacher for reference. We consider a neuron represented if $70\%$ of its separation is preserved.

\Cref{fig:distill-sim-sep}) shows that the student with $N = 2560$ represents many more neurons than what the weights-based analysis would suggest, with $7\%$ of neurons being represented without having a match in weights compared to $9\%$ that do. We additionally verify that these teacher neurons' activations can be predicted from the student neurons. At $N{=}2560$, teacher neurons with a matched
student neuron reach median $R^2\,0.858$, and
neurons with high separation but no matching student neuron ($|\cos|<0.2$) reach
$0.622$. Neurons in neither of these groups, sampled controlling for energy, score $0.527$, so overall the metric is not a strong way of discriminating represented neurons.
In \Cref{fig:distill-sep-r2}), we show that separation is a strict metric and is stricter than $R^2$ in the $N=10240$ student, motivating our choice to use it for downstream analysis.

\begin{figure}[t]
  \centering
  \includegraphics[width=\linewidth]{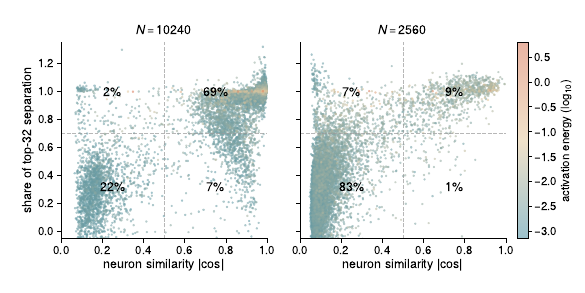}
  \caption{Neuron similarity against separation for neurons in a wide and narrow student. Even with the lenient neuron feature threshold in this setup, in the narrow student, about as many neurons are represented in superposition as there are duplicated neurons.}
  \label{fig:distill-sim-sep}
\end{figure}

\begin{figure}[t]
  \centering
  \includegraphics[width=\linewidth]{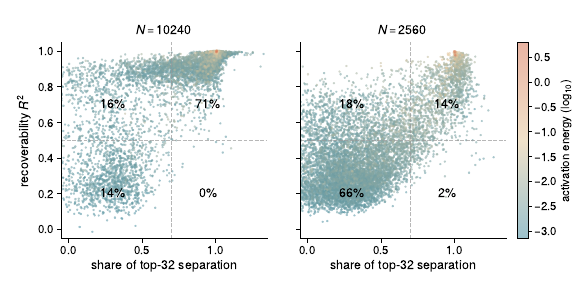}
  \caption{Share of top-32 separation ins tudent neurons against how well
  the teacher neuron's activation is recovered by regression on the student's
  neurons. Separation nearly
  implies recoverability, but not vice versa.}
  \label{fig:distill-sep-r2}
\end{figure}

\subsection{Case study: superposition in neuron 4300}
\label{app:distill-toy:case}

We study how neuron 4300 from the teacher is represented by the student.  It fires on the token
`\texttt{(}' in parenthetical asides, and its
decoder promotes variants of the token `\texttt{(}' through the logit lens. No student
neuron matches it with the strongest weigh alignment being 0.256. However, the student has a neuron that
detects the same feature: student 1224 fires on `\texttt{(}' before
citations, at activation correlation $0.39$ and decoder alignment $0.075$. However, that neuron writes in a different direction, and its contribution along the teacher's direction has the wrong sign: $-6.2$ against $+1.8$ for the whole student. We can specifically study which neurons contribute to the separation, because it is linear in contributions from specific neurons. When we rank neurons by their contributions, we find punctuation neurons in the top spots: an em-dash
neuron (2188), a comma/newline neuron (1793) and a digit neuron (2010). The em-dash neuron alone
is sufficient to create the separation ($+3.1$ on its own), and $87\%$ of the 100 tokens it fires hardest
on are tokens where teacher 4300 is in the layer's top-32.
\Cref{fig:distill-case4300} shows the four students and the teacher neuron, as well as other teacher neurons 2188 contributes to.  Overall, our conclusion from this case study is that specific student neurons can be weakly aligned to multiple teacher neurons' directions and yet push all of them in the right direction in similar circumstances, and the combination of multiple such neurons can make the representation of the neuron more accurate.

\begin{figure*}[t]
  \centering
  \includegraphics[width=\textwidth]{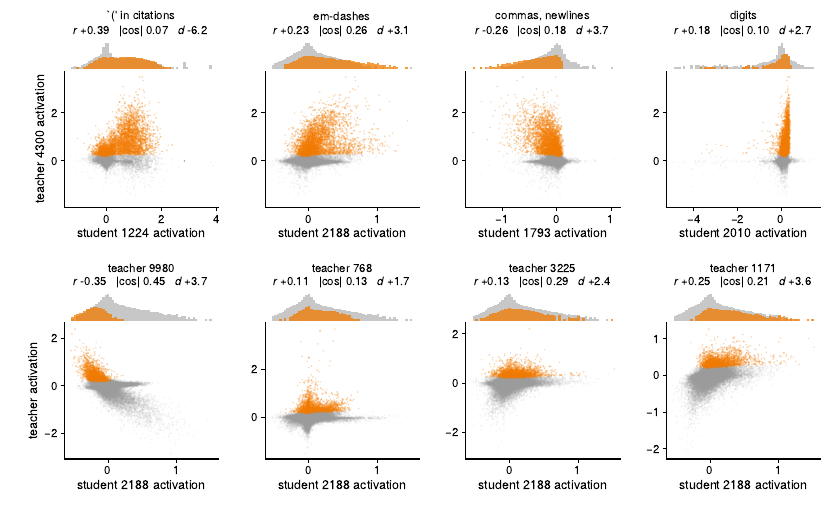}
  \caption{Teacher neuron 4300 and 4 student neurons which
  contribute most to its separation in the student. The bottom row is student
  2188 with other teacher neurons it contributes most to, showing that it pushes several neurons up. Orange: neuron 4300 is in the top 32, grey: it is not. Above each panel is the joint density of the student activation for
  the two groups on a log scale. $r$ is the activation correlation, $|\cos|$ is the
  alignment on the decoder side, $d$ is the separation the student neuron contributes on its own.}
  \label{fig:distill-case4300}
\end{figure*}

\section{Cross-layer decoders in CLT transcoders}
\label{app:clt-crosslayer}

Cross-layer transcoders are a modification to the per-layer transcoders, which we have been evaluating in this work so far, which were introduced by \citet{lindsey2025biology}, with the difference being that each feature is decoded into multiple downstream layers and not just the MLP output at the source layer. We can ask whether neuron-aligned features behave
differently in this setting: if a feature's encoder lines up with an MLP
neuron at the read layer, does its decoder also concentrate at that
layer, or does it spread further down the residual?

\paragraph{Setup.}
We use the two open-source CLTs from \texttt{mntss}: a $16$-layer
JumpReLU CLT on Llama~3.2 1B with $32{,}768$ features per layer
and a $26$-layer ReLU CLT on Gemma~2 2B with $16{,}384$ features per
layer.

Both of the transcoders read from
\texttt{hook\_resid\_mid}\footnote{The residual stream after attention, before
the pre-MLP layernorm.}
and write into \texttt{hook\_mlp\_out}.
However, each MLP neuron is
sensitive to directions in post-layernorm space. For this reason, we measure cosine similarity against a neuron's two effective encoder directions after multiplying elementwise by the affine weight of the pre-MLP layernorm.

\paragraph{Cross-layer decoder ratio.}
For each feature, we measure the extent to which it uses the cross-layer decoder, operationalized as the fraction of the $L_2$ norm of its decoder that is
written into future layers:
\[
  \mathrm{CLDR}(f, L) \;=\;
  \frac{\sum_{L' > L} \, \big\Vert (W_{\text{dec}}^{(L)})_{f, L'-L} \big\Vert_2^{2}}
       {\sum_{L' \geq L} \, \big\Vert (W_{\text{dec}}^{(L)})_{f, L'-L} \big\Vert_2^{2}}
\]

\paragraph{Results.}
Figure~\ref{fig:clt-crosslayer-ratio} shows a scatter plot of encoder cosine similarity against cross-layer decoder ratio for every feature at four read layers per model. Pearson
correlations between alignment and $\mathrm{CLDR}$ are reported in
each panel, with it being negative for all layers. Generally, non-aligned features have high CLDR ($>0.8$), and in late layers there is a distinct group of encoder-aligned features which has lower CLDR of around $0.5$.

\begin{figure*}[t]
  \centering
  \includegraphics{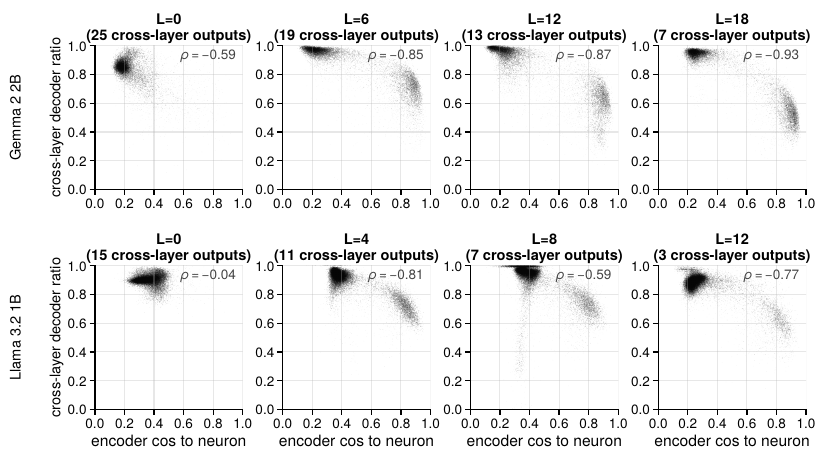}
  \caption{Cross-layer decoder ratio and
    cosine similarity to the encoder of the closest neuron for
    every CLT feature in four layers. Top row: Gemma~2
    2B mntss CLT (ReLU, $16{,}384$ features per layer) at $L =
    0, 6, 12, 18$. Bottom row: Llama~3.2 1B mntss CLT (JumpReLU,
    $32{,}768$ features per layer) at $L = 0, 4, 8, 12$. $\rho$ is the Pearson correlation between
    encoder alignment and CLDR.}
  \label{fig:clt-crosslayer-ratio}
\end{figure*}

\providecommand{\redcardtoptrim}{55bp}

\section{Features not explained by neurons}
\label{sec:clustering}

Past work has shown that a polysemantic neuron's high-magnitude activations can be seen as splitting into coherent clusters representing concepts \citep{dreyer2024pure,kopf2025prism,liu2026neuronscope}, with K-means on activations of a neuron being a simple way to identify them. We further hypothesize that each cluster is identifiable by a small set of co-activating neurons; this would imply that \emph{common and interpretable features used in any given MLP's computation can be represented as sparse combinations of neurons}.

\citet{gurnee2023haystack} shows that some features, like bigrams, \emph{are} cleanly represented as sparse combinations of neuron activations. While it is likely many neurons firing together are how features not aligned to individual neurons are represented, it is much more difficult to interpret these diffuse firings; sparsity is a necessary prior for manual inspection of discovered features. Additionally, if transcoder features correspond to combinations of neurons, the fact that transcoder circuits are more compact at the same replacement score \citep{lindsey2025biology} could be explained by them merging multiple nodes in the neuron circuit into more meaningful units.

Below we step through one case study (\S\ref{sec:clustering:red}) that motivates the hypothesis, then introduce two transcoder training procedures that target the structure directly --- NFT (\S\ref{sec:clustering:nft}) and SNRL (\S\ref{sec:nonneuron:snrl}). We then operationalize absorption and weight--activation coverage (\S\ref{sec:clustering:absorption}) and compare the resulting decompositions on cluster discriminability and few-neuron summarizability (\S\ref{sec:clustering:sort-eval}).

\subsection{``Red'' case study}
\label{sec:clustering:red}

We have found individual cases where some (TopK) transcoders have features in circuits in place of multiple active neurons at that position, and their weights are close to combinations of the neurons' weights.

We present a simple case study on the attribution graph of the sentence ``Translation of the word `\foreignlanguage{russian}{красное}'\footnote{``Red'' in Russian.}:'' for Gemma 3 4B. The model predicts ``red'', and we find weak evidence for features on this prompt representing the concept ``red'' in superposition. On Layer 22:

\begin{enumerate}
    \item Neuron 5039 (positive, \Cref{fig:red-n5039}), N$7405^+$ (\Cref{fig:red-n7405}) and N$10102^+$ (\Cref{fig:red-n10102}) all activate on the token ``\foreignlanguage{russian}{красное}''. Together, the first two write some direction relevant to the prediction being ``red'' as shown by circuit edges and ablation experiments.
    \item None of the three cleanly encode a ``red'' feature: N$5039^+$ activates on tokens ``red'' and ``transparent'' (see Figure~\ref{fig:n5039-umap} and \Cref{fig:red-n5039}); N$7405^+$ activates on various color tokens and the token ``color'' (\Cref{fig:red-n7405}), while N$10102^+$ activates on tokens that sound alike --- ``ted'', ``teddy'', ``red'' (\Cref{fig:red-n7405}).
    \item In Gemmascopes, only neuron features related to these three neurons appear in the pruned attribution graph (\Cref{fig:red-gs}). Some are more specific in high activations than their corresponding neurons, appearing to respond more commonly to the token ``red''.
    \item In our
    $k=64$ TopK transcoder,\footnote{This is an older transcoder and not the one used in the sweeps. It is trained on a different hookpoint.}
    there exist multiple features in the pruned circuit that only have the token ``red'' in maximum activations (\Cref{fig:red-bl-k64}). Only one of them is a neuron feature (for N$10102^+$), and the other two are weakly aligned to N$5039^+$.
    \item We take high activations of N$5039^+$, apply UMAP to them, then scatter the tokens these activations correspond to in \Cref{fig:n5039-umap}. We take the 4 features with more than 50 activations among the 4096 highest for the neuron for which the neuron has the highest cosine similarity to their encoder, then color the words according to which of the features activates (see legend). Both of the non-neuron features \emph{cluster} the neuron's polysemantic activations into interpretable units (see \Cref{fig:neuron-coverage} for a more clear definition).
    \item We have shown that these two features, F$28197$ and $29081$, cluster the original neuron, N$5039^+$. We can also show that their weights correspond to a combination of a small set of neurons: specifically, the decoders can be reconstructed in the following way using OMP with $K=4$ neurons:
    \begin{itemize}
        \item $F28197 \approx 0.68\,N5039^+ + 0.38\,N7405^+ + 0.15\,N10100^+ + 0.09\,N4137^+$, cosine similarity $0.85$ (vs $0.73$ from $N5039^+$ alone).
        \item $F29081 \approx 0.49\,N5039^+ + 0.40\,N7405^+ + 0.21\,N10102^+ + 0.18\,N10100^+$, cosine similarity $0.78$ (vs $0.58$ from $N5039^+$ alone).
    \end{itemize}
    These features are the combinations of a ``red'' neuron, a ``color'' neuron to disambiguate, and the lower-activating neurons from \Cref{fig:red-attrib}: 10102 (red/ted), 4137 (R), 10100 (rid/rig/rip). The base set of neurons is very close for these features, so it is not surprising they represent similar concepts.
    \item These features are examples of few-neuron combinations creating cluster in neurons' activations. The point about clustering may be debated, as both features activate at once and contribute --- they may co-activate on a manifold --- but it is evidence that some features in transcoders represent crisp monosemantic concepts by combining multiple polysemantic neurons.
\end{enumerate}

\begin{figure}[!ht]
\centering
\includegraphics[width=\linewidth]{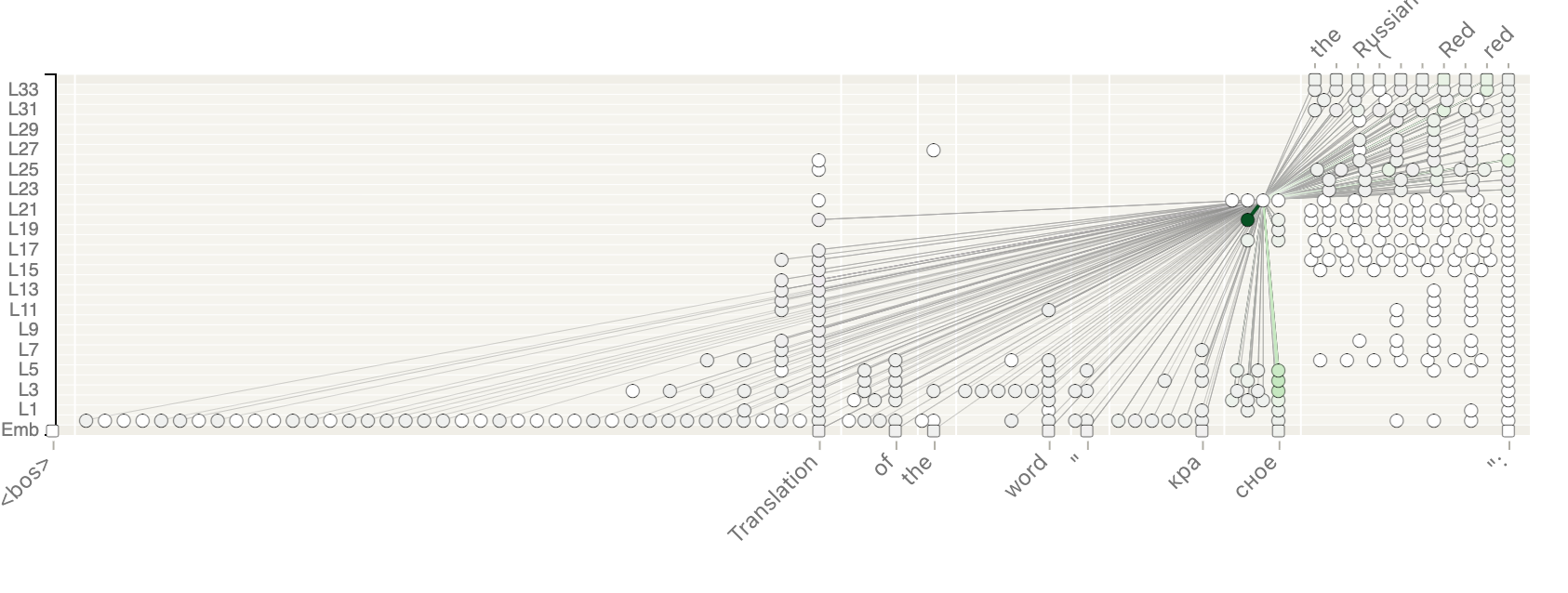}\\[4pt]
\begin{subfigure}[t]{0.485\linewidth}
\vspace{0pt}%
\centering
\includegraphics[width=\linewidth,trim=0 0 0 \redcardtoptrim,clip]{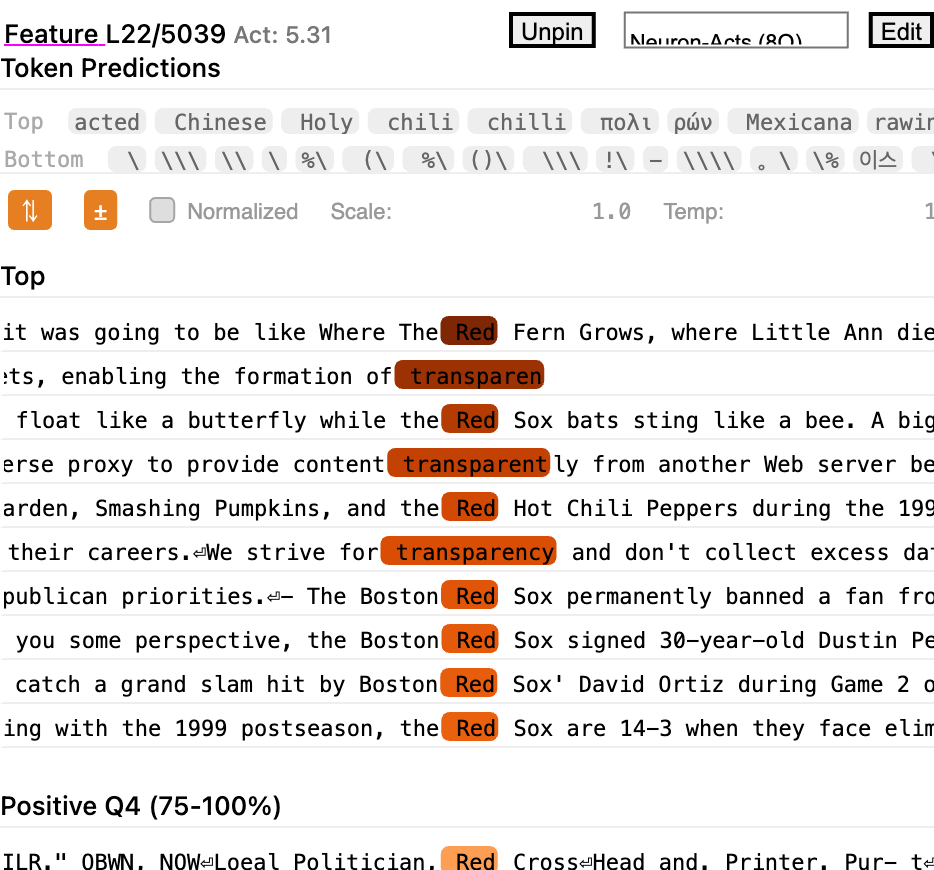}
\caption{Neuron 5039$^+$}
\label{fig:red-n5039}
\end{subfigure}\hfill
\begin{subfigure}[t]{0.485\linewidth}
\vspace{0pt}%
\centering
\includegraphics[width=\linewidth,trim=0 0 0 \redcardtoptrim,clip]{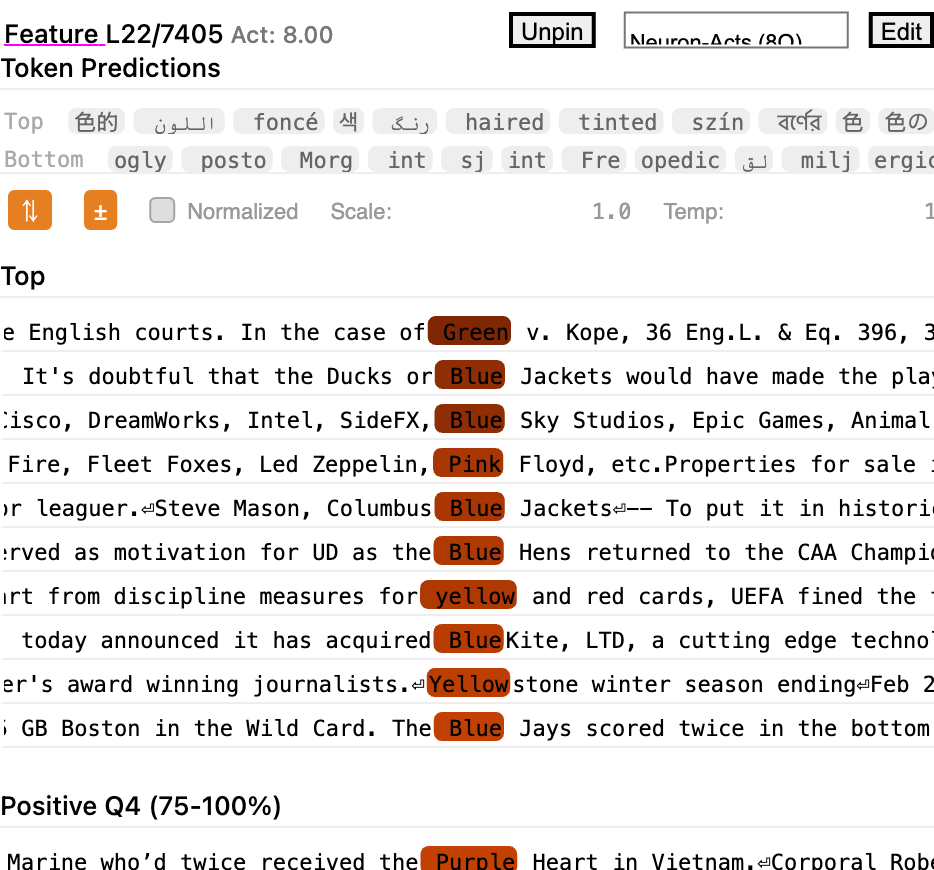}
\caption{Neuron 7405$^+$}
\label{fig:red-n7405}
\end{subfigure}\\[4pt]
\begin{subfigure}[t]{0.32\linewidth}
\vspace{0pt}%
\centering
\includegraphics[width=\linewidth,trim=0 0 0 \redcardtoptrim,clip]{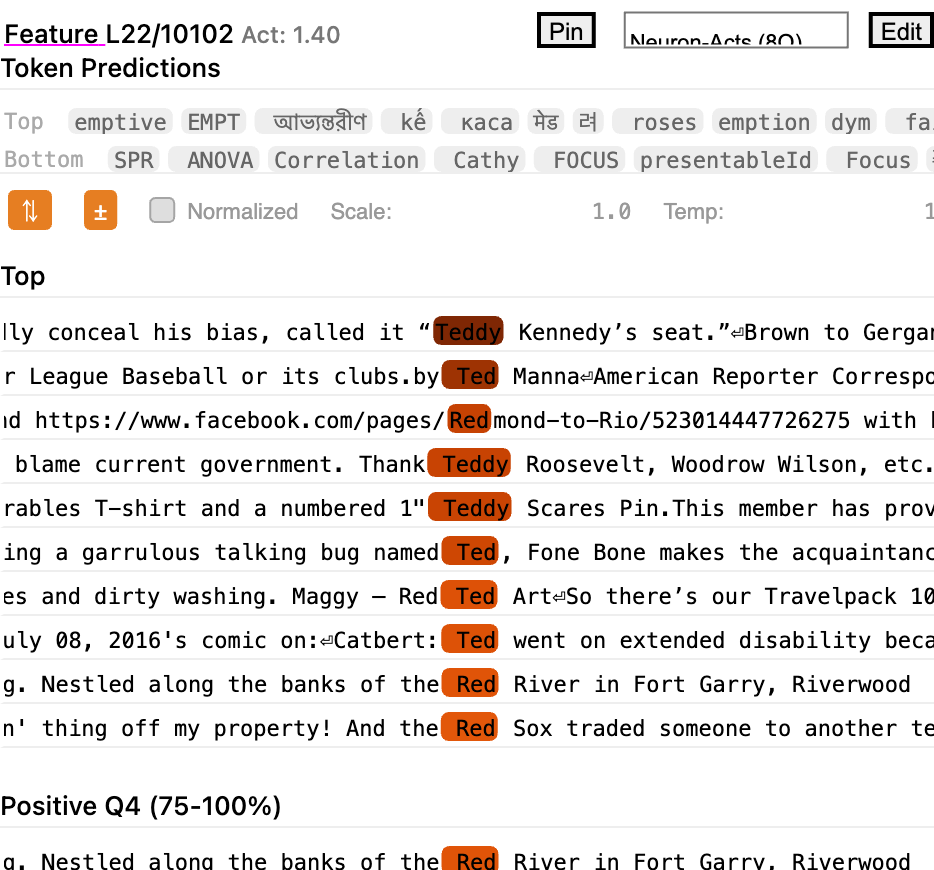}
\caption{Neuron 10102$^+$}
\label{fig:red-n10102}
\end{subfigure}\hfill
\begin{subfigure}[t]{0.65\linewidth}
\vspace{0pt}%
\centering
\includegraphics[width=0.485\linewidth,trim=0 0 0 \redcardtoptrim,clip]{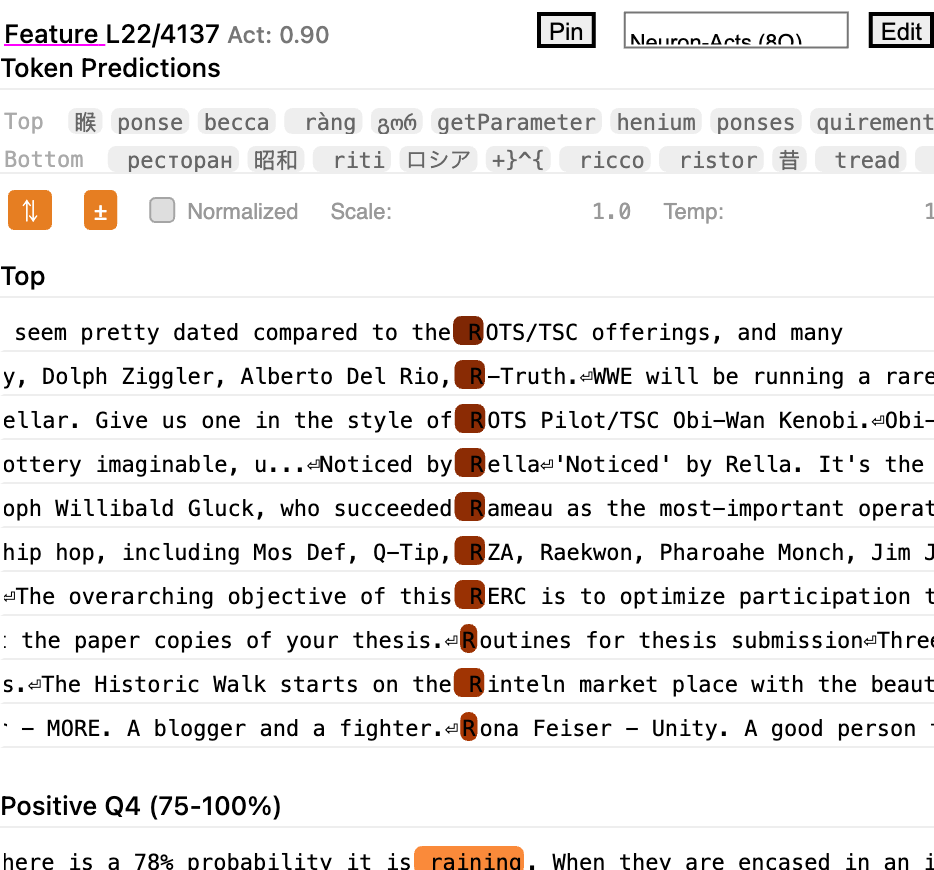}\hfill
\includegraphics[width=0.485\linewidth,trim=0 0 0 \redcardtoptrim,clip]{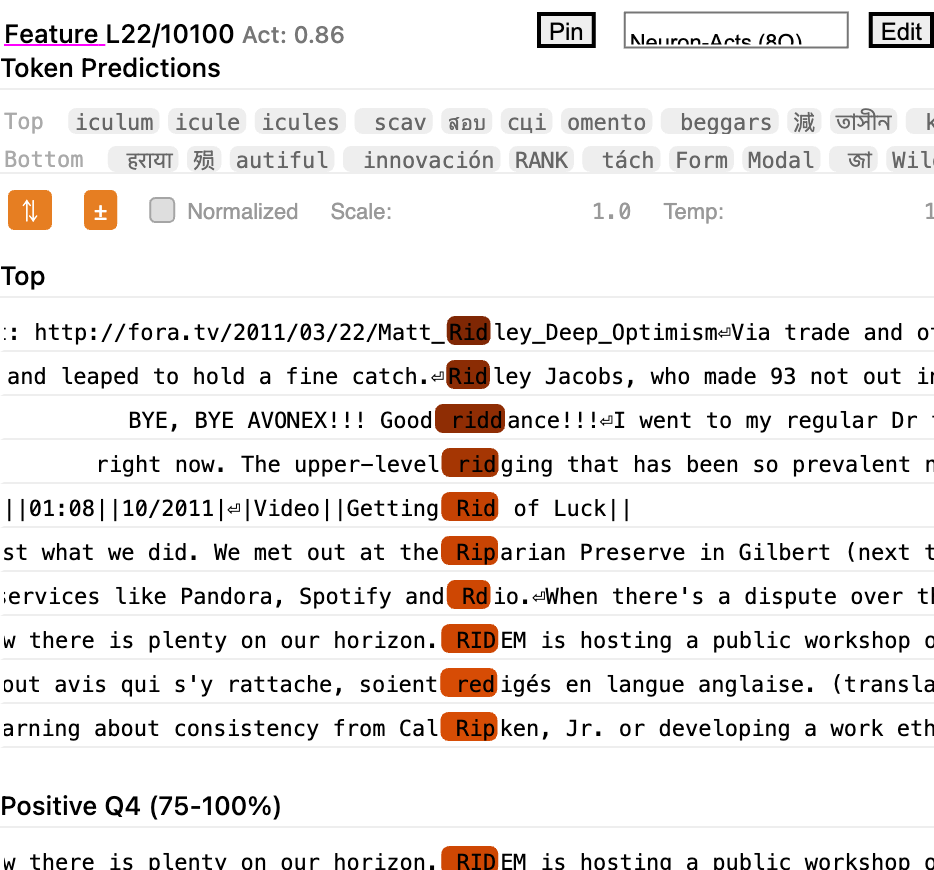}
\caption{Neurons 4137$^+$, 10100$^+$}
\label{fig:red-extras}
\end{subfigure}
\caption{``Red'' case study on Gemma~3 4B L22. Top: attribution graph
for ``Translation of the word \foreignlanguage{russian}{красное}:''
(predicting ``Red''/``red''), with N5039 highlighted. Bottom:
max-activating contexts for selected MLP neurons, with
activations N7405$^+$\,$=\,8.00$,
N5039$^+$\,$=\,5.31$, N10102$^+$\,$=\,1.40$, N4137$^+$\,$=\,0.9$,
N10100$^+$\,$=\,0.86$.}
\label{fig:red-attrib}
\end{figure}

\begin{figure}[!ht]
\centering
\begin{subfigure}[t]{0.62\linewidth}
\vspace{0pt}%
\centering
\includegraphics[width=\linewidth]{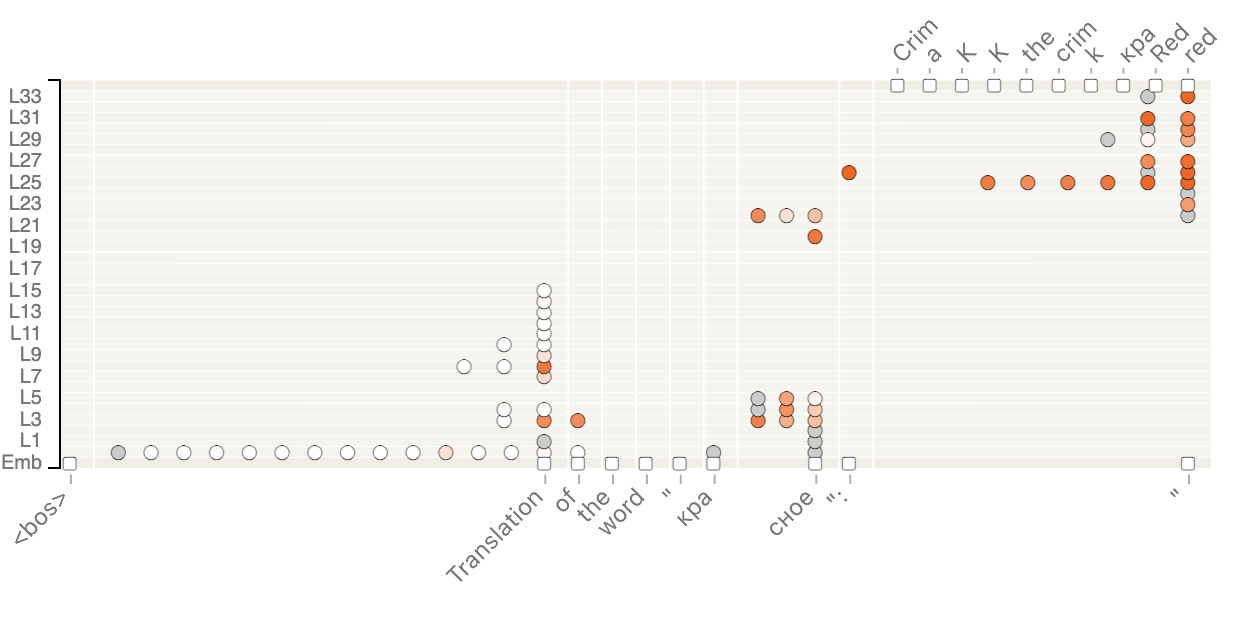}
\caption{Attribution graph.}
\label{fig:red-bl-k64-graph}
\end{subfigure}\hfill
\begin{subfigure}[t]{0.36\linewidth}
\vspace{0pt}%
\centering
\includegraphics[width=\linewidth,trim=0 0 0 \redcardtoptrim,clip]{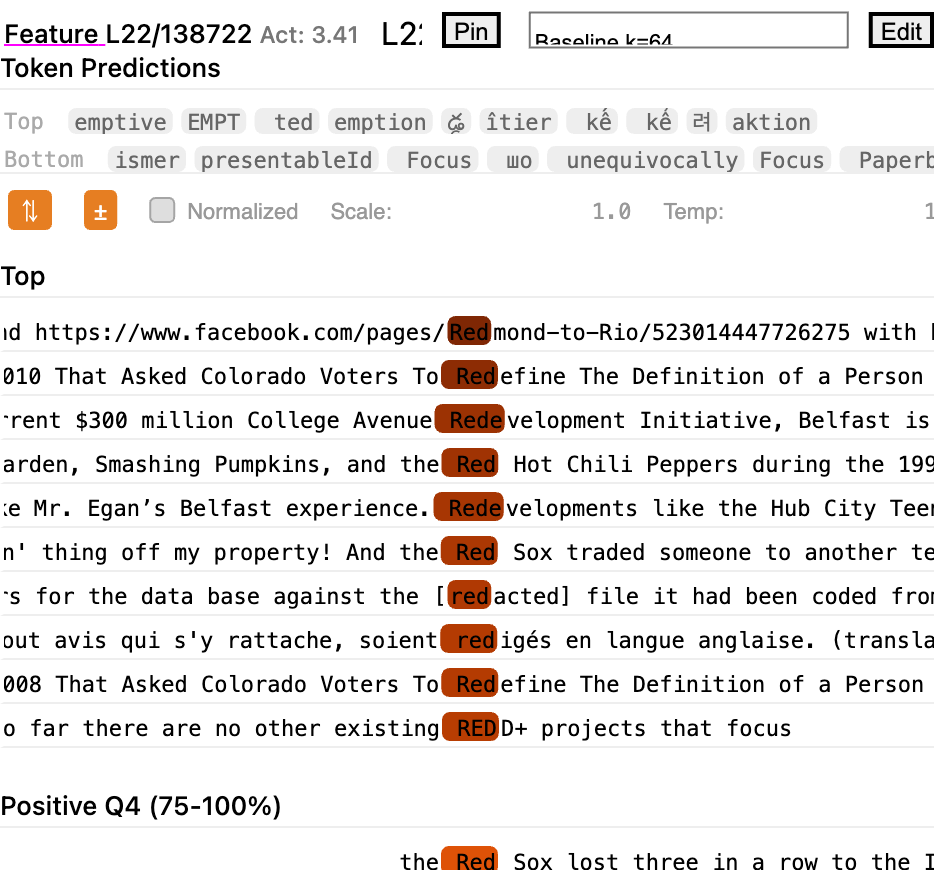}
\caption{F138722 (N10102, $e=0.82$, $d=0.93$).}
\label{fig:red-bl-k64-f138722}
\end{subfigure}\\[4pt]
\begin{subfigure}[t]{0.485\linewidth}
\vspace{0pt}%
\centering
\includegraphics[width=\linewidth,trim=0 0 0 \redcardtoptrim,clip]{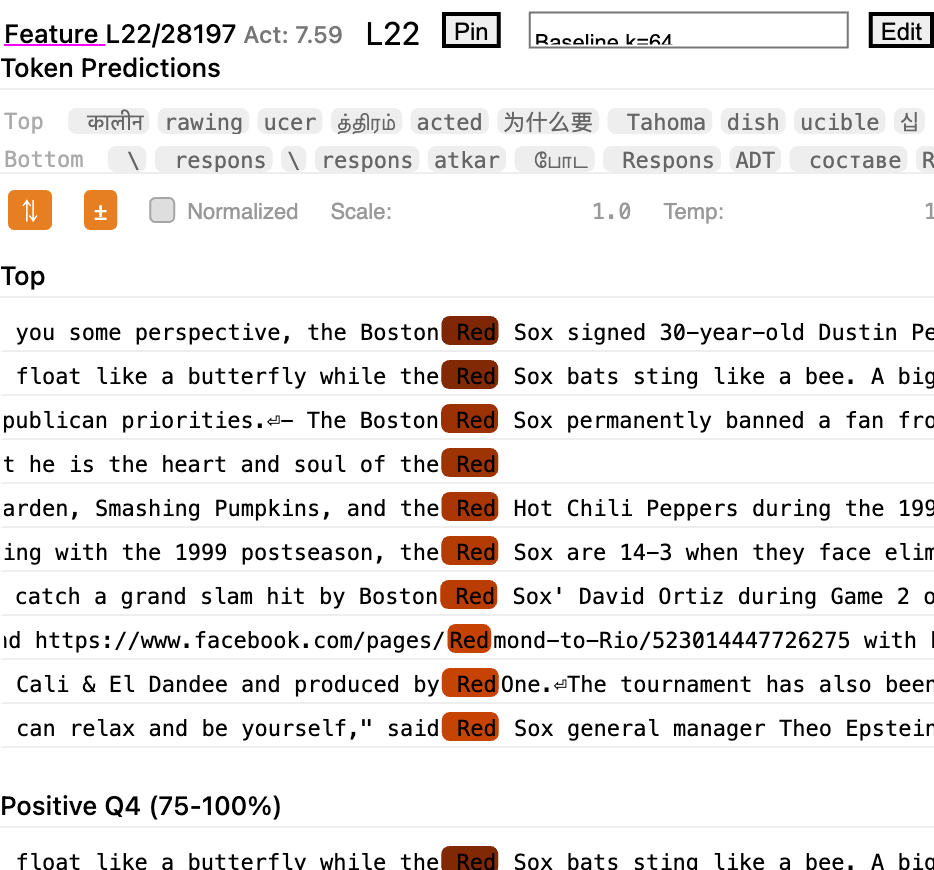}
\caption{F28197 (N5039, $e=0.60$, $d=0.77$).}
\label{fig:red-bl-k64-f28197}
\end{subfigure}\hfill
\begin{subfigure}[t]{0.485\linewidth}
\vspace{0pt}%
\centering
\includegraphics[width=\linewidth,trim=0 0 0 \redcardtoptrim,clip]{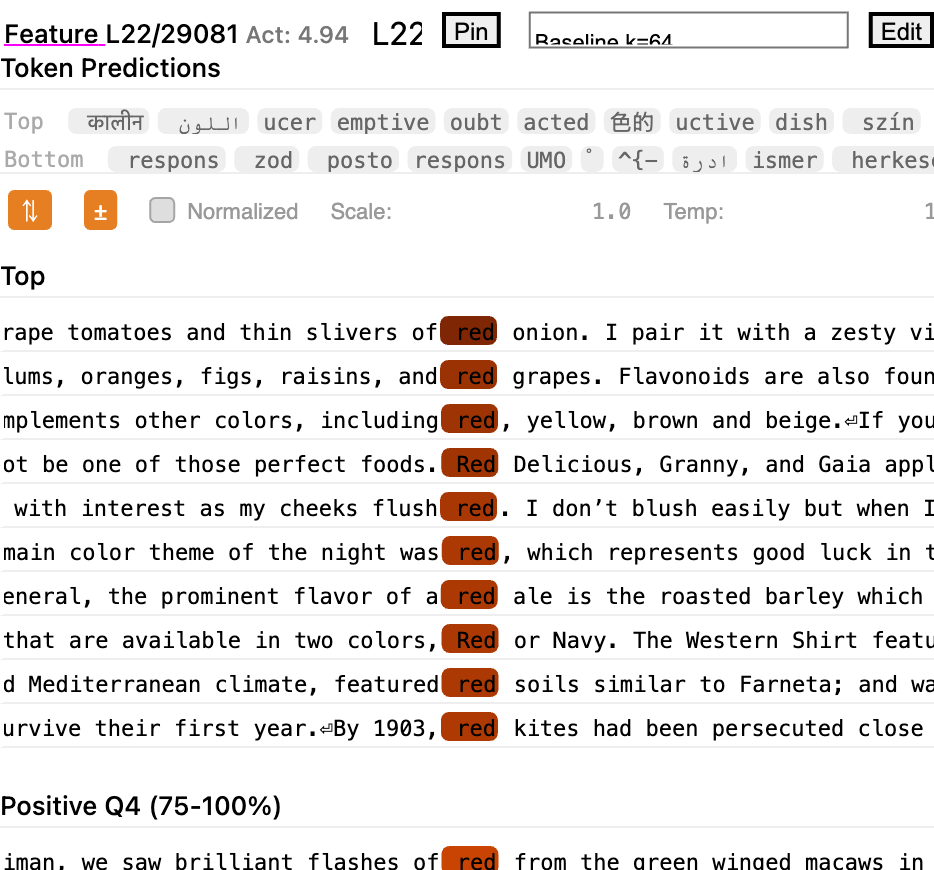}
\caption{F29081 (N5039, $e=0.43$, $d=0.61$).}
\label{fig:red-bl-k64-f29081}
\end{subfigure}
\caption{``Red'' case study, $k=64$ TopK transcoder. Attribution graph shows two mildly (light orange; encoder alignment, from 0 to 1, is represented as color, from white to orange) neuron-aligned features and one strongly neuron-aligned feature (F138722). N10102, the neuron it is aligned to, activates on ``red''/``ted'' (\S\ref{fig:red-gs-f10102}). The other two features are weakly aligned to neuron 5039, and both activate only on the token ``red''. \Cref{fig:n5039-umap} shows that they both cluster the neuron.}
\label{fig:red-bl-k64}
\end{figure}

\begin{figure}[!ht]
\centering
\includegraphics[width=\linewidth]{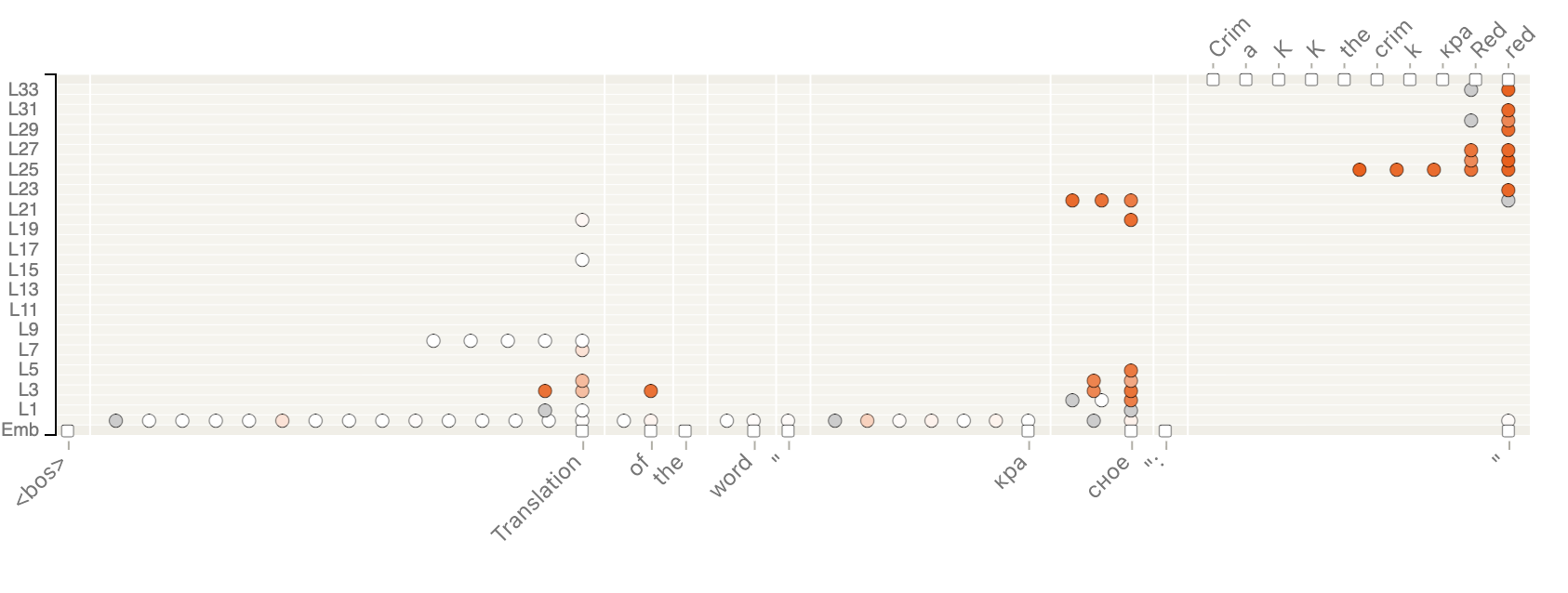}\\[4pt]
\begin{subfigure}[t]{0.485\linewidth}
\vspace{0pt}%
\centering
\includegraphics[width=\linewidth,trim=0 0 0 \redcardtoptrim,clip]{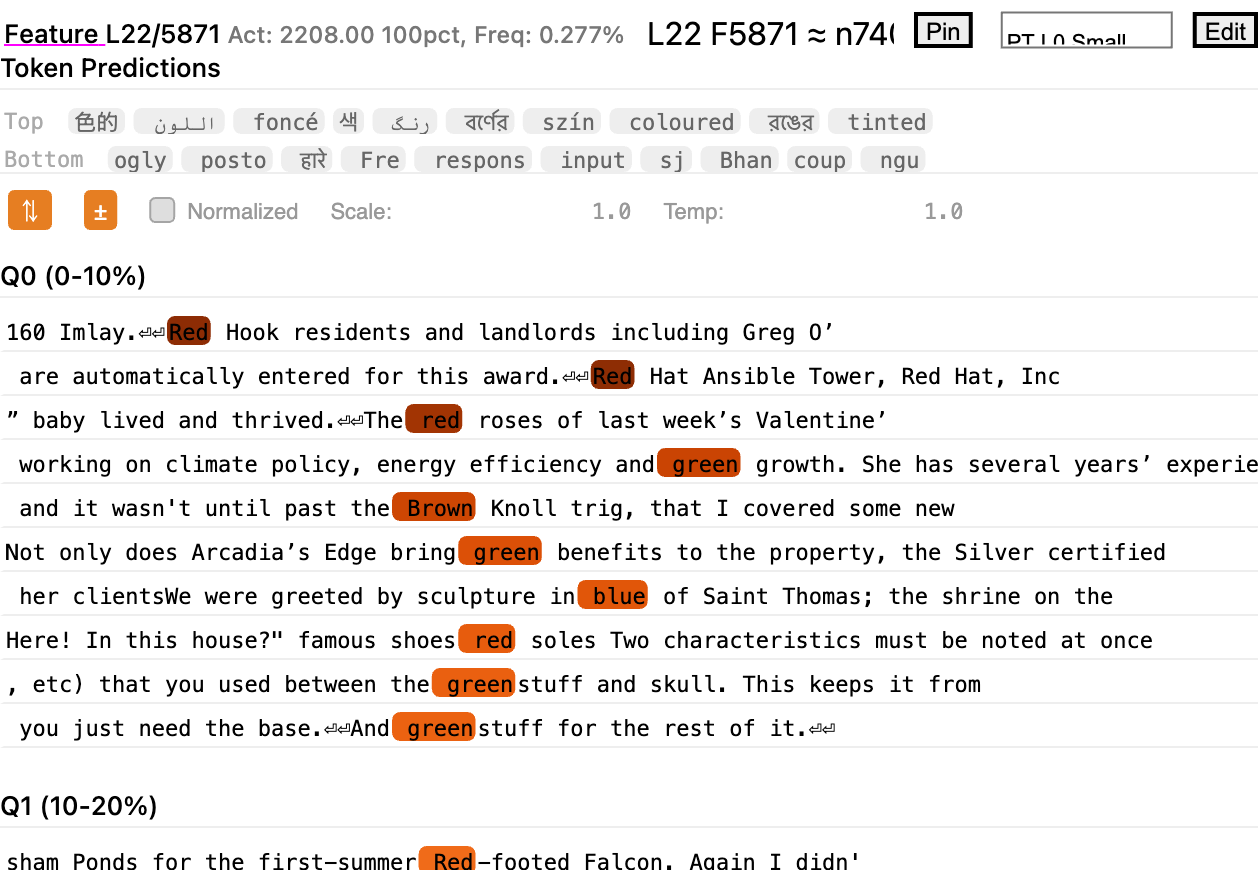}
\caption{F5871 (N7405, $e=0.88$, $d=0.96$).}
\label{fig:red-gs-f5871}
\end{subfigure}\hfill
\begin{subfigure}[t]{0.485\linewidth}
\vspace{0pt}%
\centering
\includegraphics[width=\linewidth,trim=0 0 0 \redcardtoptrim,clip]{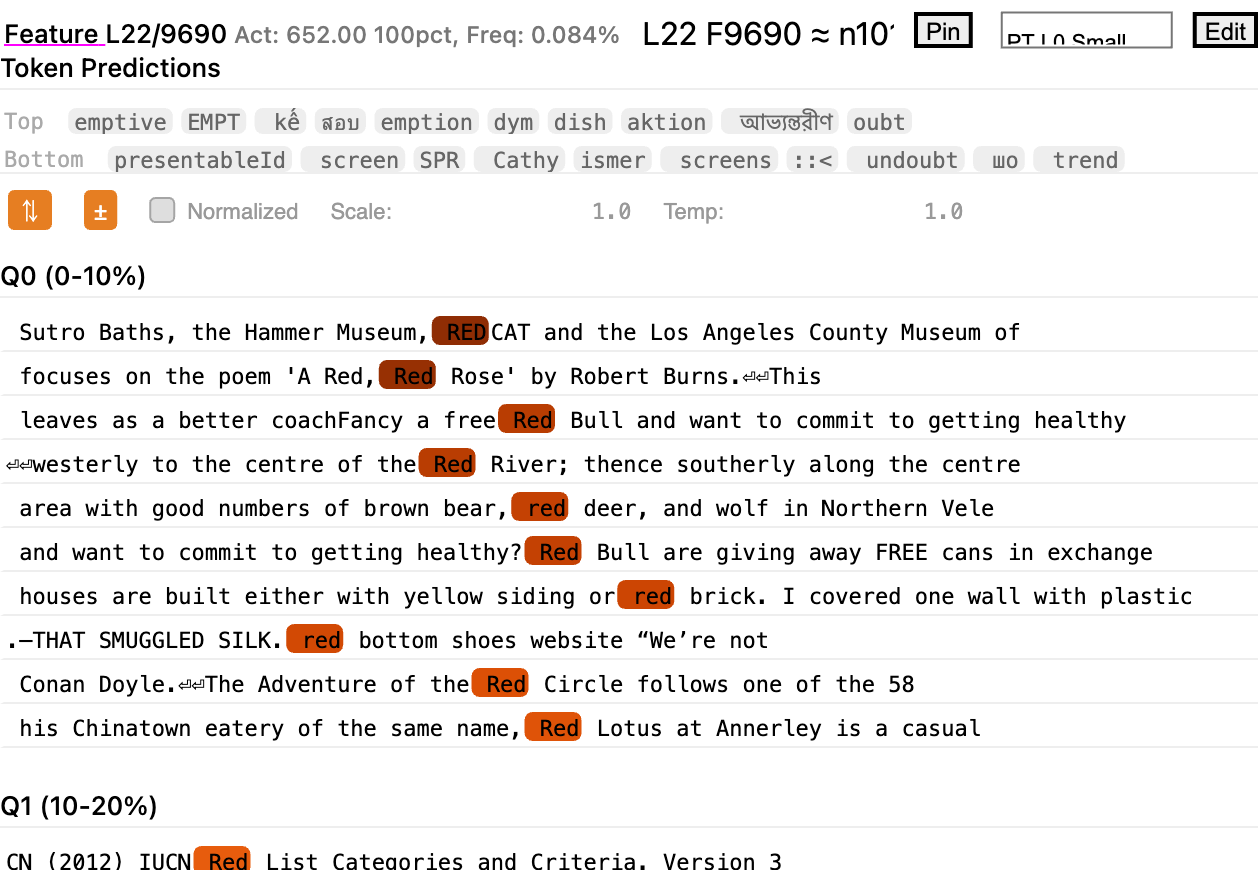}
\caption{F9690 (N10102, $e=0.91$, $d=0.94$).}
\label{fig:red-gs-f9690}
\end{subfigure}\\[4pt]
\begin{subfigure}[t]{0.485\linewidth}
\vspace{0pt}%
\centering
\includegraphics[width=\linewidth,trim=0 0 0 \redcardtoptrim,clip]{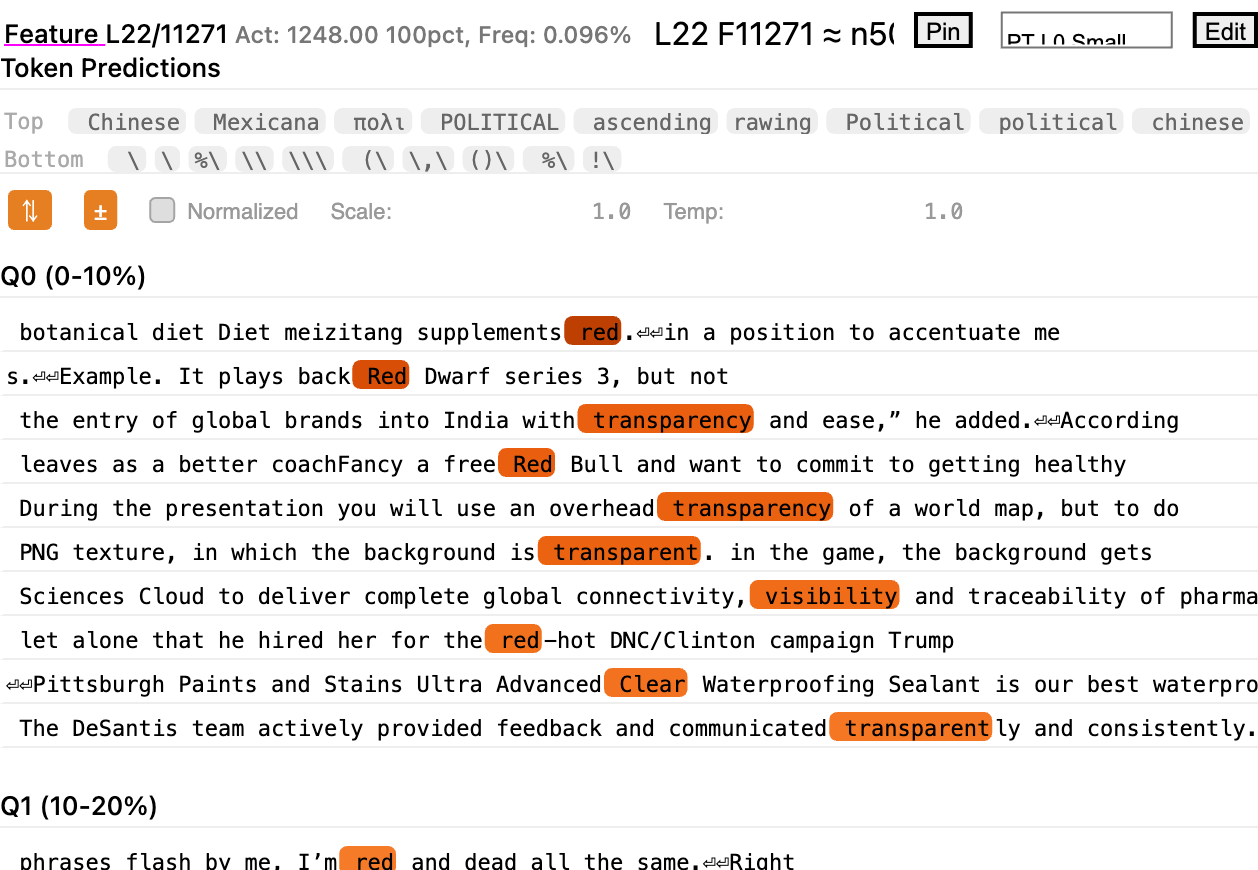}
\caption{F11271 (N5039, $e=0.93$, $d=0.98$).}
\label{fig:red-gs-f11271}
\end{subfigure}\hfill
\begin{subfigure}[t]{0.485\linewidth}
\vspace{0pt}%
\centering
\includegraphics[width=\linewidth,trim=0 0 0 \redcardtoptrim,clip]{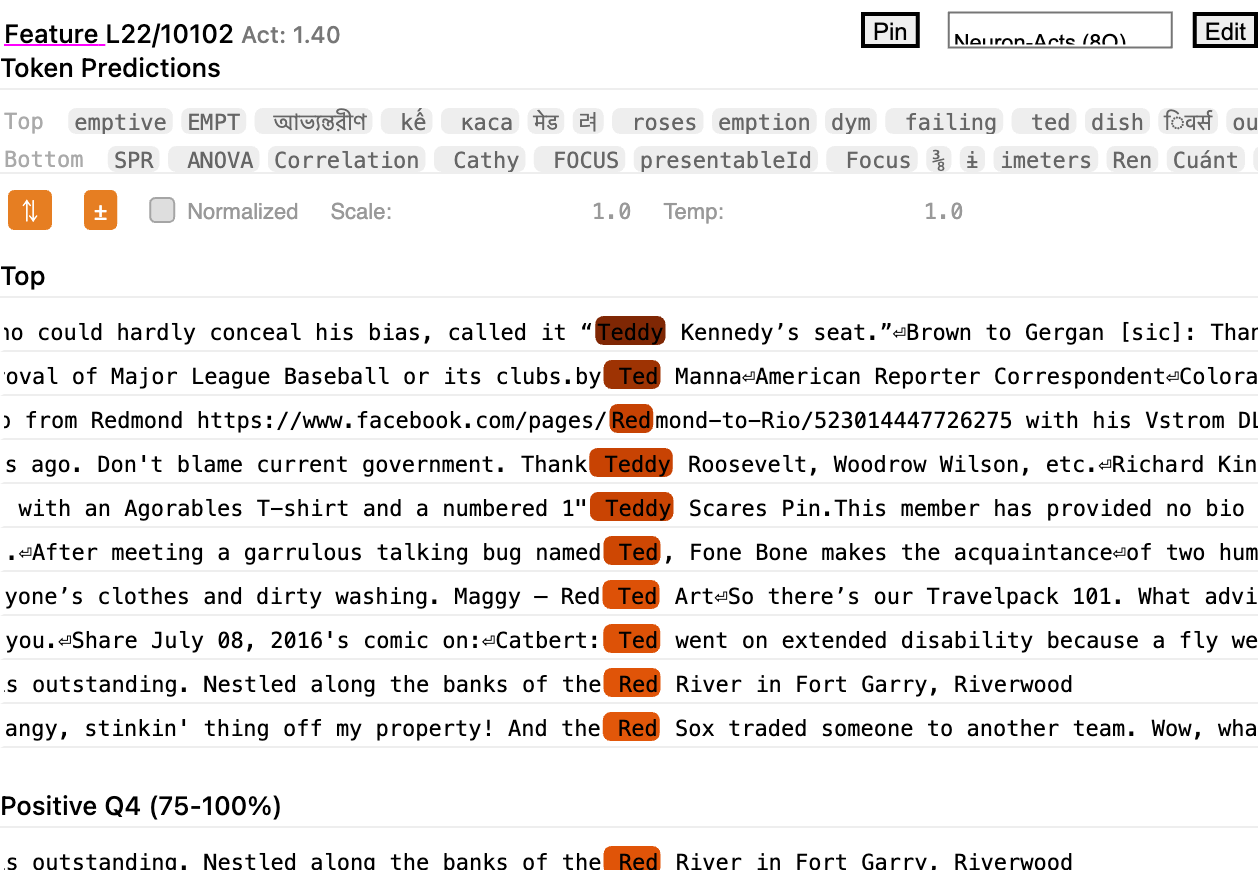}
\caption{Neuron $10102^+$}
\label{fig:red-gs-f10102}
\end{subfigure}
\caption{``Red'' case study, Gemmascope transcoder. Attribution graph shows three neuron-aligned features in dark orange.
One feature is aligned to a color neuron, 7405. Another is aligned to N10102, a ``red''/``ted'' neuron, the activations of which are shown in the bottom right. The third one is a neuron feature for N5039 and shows the same polysemanticity between ``red'' and ``transparent''.}
\label{fig:red-gs}
\end{figure}

\begin{figure}[!ht]
\centering
\includegraphics[width=\linewidth]{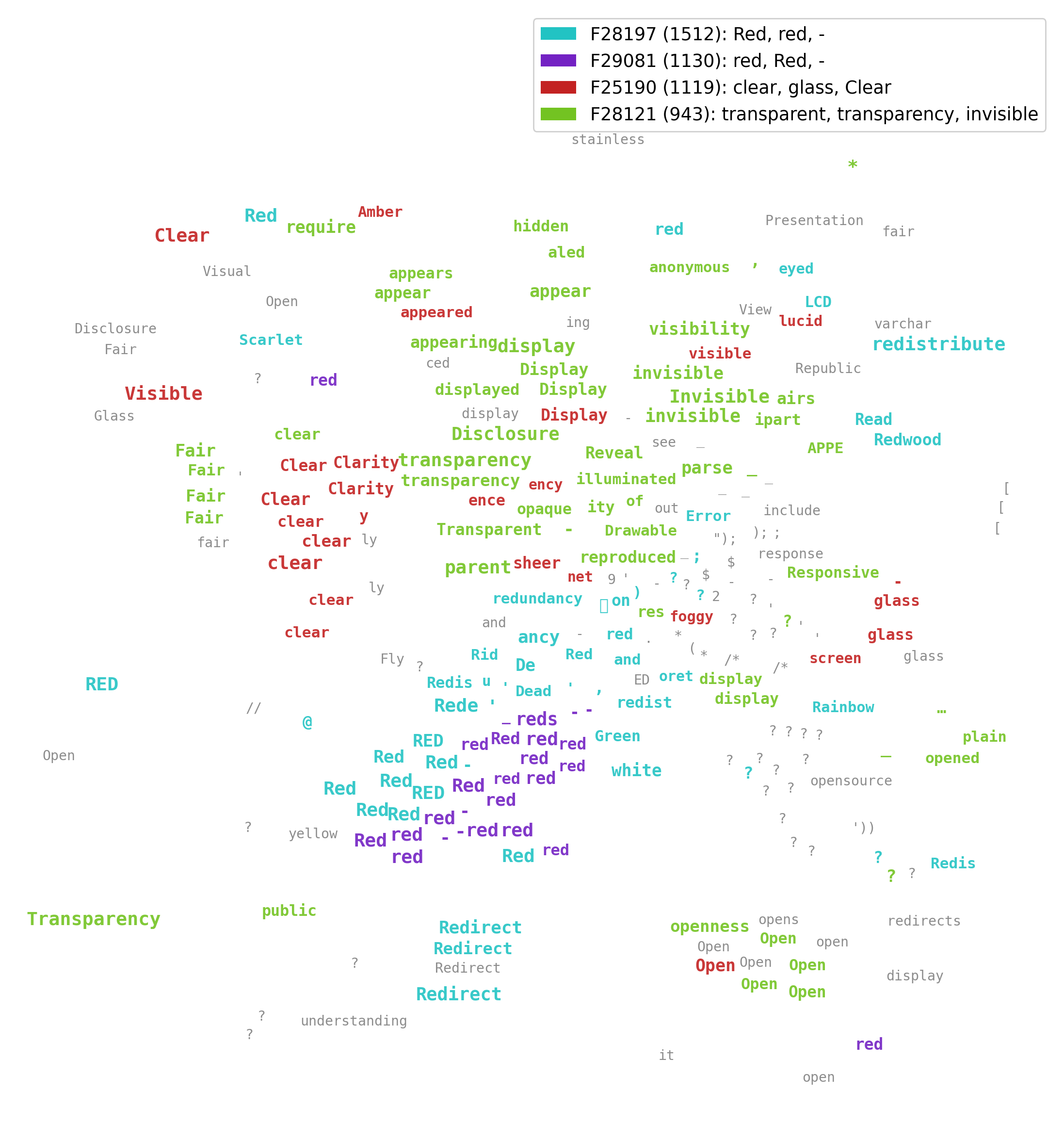}
\caption{UMAP of the max-activating contexts of N5039$^+$
  (Gemma 3 4B, L22). Activations are colored according to the feature in the $k=64$ transcoder which activates most strongly on them. Distinct clusters reveal polysemanticity: the
  neuron fires on \emph{red}, \emph{transparent}/\emph{clear}/\emph{clarity},
  \emph{display}/\emph{displaying}, \emph{visibility}/\emph{visible},
  \emph{appear(ing)}, \emph{redirect}, and others. The features find interpretable monosemantic clusters:
  one ``red'' feature activates more specifically than the other:
  feature 29081 (purple) activates only on the token ``red''. There exists a distinct ``transparency (communication) and fairness'' (green, with some exceptions) feature and an ``appearance, transparency (optical) and visibility'' (red) feature.
  }
\label{fig:n5039-umap}
\end{figure}

\subsection{Neuron Feature Transcoders: transcoders that cluster neurons by construction}
\label{sec:clustering:nft}

In regular transcoders, each neuron and sign only gets approximately one neuron feature (\Cref{fig:neuron-coverage}); however, this may be due to the fact that ``true'' neuron features are hidden by the initialization (\Cref{sec:nfc:cross-seed}). We train transcoders which explicitly force active features to correspond to top K neurons: each feature has a ``parent'' neuron and activates only when that neuron is in the top K.\footnote{With the neuron top K being a separate hyperparameter from the final transcoder TopK, set to twice the sparsity in our experiments.} In other words, each neuron has $G=16$ features, and these features can only activate when the neuron is active. Specifically, we select $r=4$ highest activating features belonging to the neuron for the final TopK selection.

Neuron Feature Transcoders (NFTs) train well. While they increase FVU, it is only by $0.026$ on the specific setting we tested, and they still have significantly lower FVU than sparsified MLPs in a comparable setting. This provides evidence that transcoder features can be seen as clustering neuron activations into more interpretable regions and provides a possible mechanism by which transcoders can have lower FVU. As expected, because of matching neuron directions, they slightly outperform a neuron-initialized transcoder on KL despite having higher FVU.

NFTs produce more neuron features. As described in \Cref{fig:ninit-scatter}, when considering the effective encoders  \footnote{After picking the 4 feature activations for each of the neurons, we add the linearized activation of the neuron, the projection of the input $x$ onto the $W_{up} \pm W_{gate}$, with a learnable scaling factor, which makes the effective encoder weight the original weight with the projection added: $\tilde{w}^{(j)}_{\mathrm{enc}} = w^{(j)}_{\mathrm{enc}} + s_j\,\big(W_\mathrm{up}^{(i)} \pm W_\mathrm{gate}^{(i)}\big)$, where $i$ is the neuron assigned to feature $j$ and $s_j$ is the learned scale.} and decoders of NFTs, $44\%$ of features in the dictionary and $63\%$ in activations have decoders aligned with neurons ($>0.6$ similarity), and virtually all features have encoders strongly aligned with neurons, leading to slightly stronger alignment than for NI.

\subsection{Aligning features to sparse neuron combinations (SNRL)}
\label{sec:nonneuron:snrl}

We additionally hypothesize that the encoder and decoder directions transcoders learn
can be explained as sparse combinations of MLP neuron directions plus
uninterpretable noise. To test and exploit this, we start with a NI transcoder initialization and add a training-time
loss, which we call the \textbf{Sparse Neuron Reconstruction Loss (SNRL)}, that aligns
each feature's encoder and decoder with a sparse approximation in
the basis of neuron encoders and decoders.

Let $W_{\mathrm{enc}}^{(j)}, W_{\mathrm{dec}}^{(j)} \in \mathbb{R}^{d}$ be
the encoder and decoder of transcoder feature $j$, and let
$N_i^{\mathrm{enc}} = W_{\mathrm{up}}^{(i)} \pm W_{\mathrm{gate}}^{(i)}$
and $N_i^{\mathrm{dec}} = W_{\mathrm{down}}^{(i)}$ denote the (sign-split)
input and output directions of MLP neuron $i$. Per training step, for
each feature $j$, we:

\begin{enumerate}
  \item Compute cosine similarities of feature $j$ with every neuron, on
    both sides: $s^{\mathrm{enc}}_{ji} = \cos(W_{\mathrm{enc}}^{(j)},
    N_i^{\mathrm{enc}})$ and $s^{\mathrm{dec}}_{ji} =
    \cos(W_{\mathrm{dec}}^{(j)}, N_i^{\mathrm{dec}})$.
  \item Pick the top $k_n{=}4$ neurons by combined encoder-decoder similarity:
    $\mathcal{S}_j = \arg\text{top-}k_n\, \big(\,|s^{\mathrm{enc}}_{ji}|
    + |s^{\mathrm{dec}}_{ji}|\,\big)$. The same neuron set is used for
    encoder and decoder reconstruction.
  \item Compute the cosine similarity between each feature row and its
    similarity-weighted reconstruction from $\mathcal{S}_j$:\footnote{This is not the L2-optimal reconstruction, but is used over a pseudoinverse for efficiency reasons.}
    \begin{align}
      c^{\mathrm{enc}}_j &= \cos\!\Big(W_{\mathrm{enc}}^{(j)},\,
        \sum_{i \in \mathcal{S}_j} s^{\mathrm{enc}}_{ji}\, N_i^{\mathrm{enc}}\Big), \\
      c^{\mathrm{dec}}_j &= \cos\!\Big(W_{\mathrm{dec}}^{(j)},\,
        \sum_{i \in \mathcal{S}_j} s^{\mathrm{dec}}_{ji}\, N_i^{\mathrm{dec}}\Big).
      \label{eq:snrl-cos}
    \end{align}
  \item Add a term to the training objective that maximizes them:\footnote{$\lambda = 0.05$ for our experiments.}
    \begin{equation}
      \mathcal{L}_{\mathrm{SNRL}} = -\lambda \sum_j \big(c^{\mathrm{enc}}_j + c^{\mathrm{dec}}_j\big).
      \label{eq:snrl-loss}
    \end{equation}
\end{enumerate}

Like NFT, SNRL significantly increases the alignment of both encoders and decoders, and makes $67\%$ of features have encoder and decoder similarity $>0.6$. However, it comes with a tradeoff of FVU which is $0.018$ higher than the baseline, and higher KL divergence. However, despite that, the average Jacobian cosine similarity is higher than for the baseline and approaches the neuron initialization. Overall, like NFT, SNRL can move features closer to their top-1 closest neurons; qualitatively, we find many cases where the other 3 neurons are semantically relevant and describe the cluster the features belong to, but our automated evaluations do not capture this.

\subsection{Testing the clustering hypothesis: basic metrics}
\label{sec:clustering:absorption}

To compare how cleanly different transcoders fit the neuron clustering hypothesis, we need to evaluate whether features cluster a neuron in the first place. NFT with $r=1$ enforces clustering, and there are three conditions, not mutually exclusive, to fulfill:

\paragraph{Absorption.}\footnote{This is distinct from feature absorption as defined in \citep{chanin2025absorption}. However, we can see clustering as feature absorption of the strongest neuron feature by features defining other clusters.} Each feature only fires when its parent neuron fires (activates above some threshold; ideally, the number of active neurons and features is comparable). This can be seen as precision for prediction of the neuron activation with the feature.
\paragraph{Coverage.} The features which belong to a neuron together make up all of the activations for the neuron. Anecdotally, in practice, coverage is difficult to achieve for neurons because some of the activations do not belong to a specific theme that can be grouped into a cluster, and transcoders do not tend to activate on all examples in the cluster.
\paragraph{Exclusivity.} For a given neuron, only one of its child features activates. This is broken by $r>1$ in NFT, and can be taken care of by picking on each token only the cluster with the highest activation. If exclusivity is violated, the clustering becomes more like a sparse encoding, which is still an interesting object to study, but is outside of the scope of this work.

\paragraph{Cluster assignment.} Before these properties can be evaluated, we must tackle the problem of assigning features to clusters. In architectures other than NFT, there is no explicit grouping of features into clusters. However, in NI and its descendant SNRL, due to features not moving far from initialization, there exists an assignment to neurons by the closest encoder weight, which is also fixed at initialization.

This approach, in which we assign all the features to the neuron to which they have closest encoder, can be extended to other transcoders. Whether this is valid is what we test by evaluating absorption.

\paragraph{Procedure.}
We compute absorption for every feature that fires at least 10 times on the 10M token cache we use for sort evals. We use a conservative $K=128$ as the threshold for ``active'' on the neuron side. We measure two statistics for absorption: $p_1$, which we have described above, the probability that the neuron is in the active set when a feature activates, and $p_{2:4}$, a related quantity which is the probability that \emph{any of the second through fourth closest neurons activates}. The first measures clustering directly; the second evaluates the hypothesis that features are combinations of multiple neurons --- specifically, that the feature is specified as a cluster of its top neuron that is also selective for another neuron. This implies both neurons would commonly activate when the feature is active; alternatively, even if the feature is some different logical combination of neuron activations and not an AND with one neuron (for example, in the case that it's an OR), we would still see a nontrivial probability of the neuron being active.

\paragraph{Results.}
Table~\ref{tab:absorption} reports $p_1$ and $p_{2:4}$ for our trained transcoders. Surprisingly, we find that the baseline and Gemmascope have strong absorption by both the top 1 and next 3 neurons, much higher than chance ($\sim 0.01$, $K_{\mathrm{neurons}}/N_{\mathrm{neurons}}$), and our interventions do not improve absorption significantly. SNRL even appears to hurt absorption by neurons $2:4$, meaning its few-neuron alignment may act to move the feature closer to its top neuron and away from the other 3. NFT is a notable outlier because of the architectural constraint that forces perfect top-1 absorption.

\begin{table}[h]
  \centering
  \small
  \setlength{\tabcolsep}{6pt}
  \begin{tabular}{lcc}
    \toprule
    Transcoder & $p_1$ & $p_{2{:}4}$ \\
    \midrule
    \multicolumn{3}{l}{\emph{Our transcoders}} \\
    Baseline $k{=}64$        & --- & --- \\
    Baseline $k{=}32$        & 0.64 & 0.77 \\
    NI $k{=}32$              & 0.66 & 0.70 \\
    NFT $k{=}32$, $G{=}16, r{=}4$ & 1.00 & 0.66 \\
    SNRL $k{=}32$            & 0.71 & 0.57 \\
    \midrule
    \multicolumn{3}{l}{\emph{Gemmascope}} \\
    Gemmascope L0-small & 0.63 & 0.75 \\
    \bottomrule
  \end{tabular}
  \caption{Absorption evaluation for different transcoder-based clustering methods on Gemma 3 4B L22, with top $K{=}128$ neurons considered active and $10$M SmolLM2 tokens. $p_1$ is the probability the feature's top-1 weight-aligned neuron is in the active set; $p_{2{:}4}$ the same for the next three weight-aligned neurons.
    }
  \label{tab:absorption}
\end{table}

\subsection{Cluster discriminability evaluations}
\label{sec:clustering:sort-eval}

The fact that a neuron can be decomposed into clusters does not imply the clusters are coherent or even distinguishable by humans. We therefore test whether the clusters are distinct concepts by asking a Claude Sonnet judge to discriminate between them. We describe the methodology in Appendix~\ref{app:eval-methods:cluster-sorteval}; in brief, we conduct two evaluations:

\begin{itemize}
  \item \textbf{Cluster match accuracy.} Given the top-10 max-activating contexts from cluster A and cluster B (of one neuron), as well as a held-out query, the judge picks A or B. The number we report is the accuracy in distinguishing between random pairs.
  \item \textbf{Top neuron match accuracy.} A similar evaluation, but instead of seeing activations from clusters A and B directly, the judge sees aggregated activations from the 4 neurons closest to the cluster, other than the neuron being evaluated. The judge picks which neuron pool the cluster's panel firings correspond to.
\end{itemize}

\paragraph{Methods.}
We evaluate the clustering methods on the same set of 50 randomly chosen neurons. In addition to transcoders, as a baseline, we evaluate cosine K-means with 16 clusters on the dense MLP post-activations. The neuron being evaluated and the eight globally densest neurons are masked out so cluster weights are more interpretable.

\paragraph{Results.}
We find that all clustering methods get similarly good performance at distinguishing clusters when shown activations of the clusters (\Cref{tab:clustering-properties}). However, on matching with top neurons, surprisingly, baselines which do not take sparse combinations of neurons into account at all --- K-means and the baseline --- achieve the highest accuracy. This result is somewhat surprising and deserves further scrutiny; however, it does demonstrate that dense $k$-means finds structure that can be explained by just a handful of neurons. Out of the remaining methods, SNRL and neuron init have the highest matching accuracy; NFT has the lowest performance.

\begin{table}[h]
\centering
\small
\setlength{\tabcolsep}{6pt}
\begin{tabular}{lccc}
\toprule
Transcoder & FVU $\downarrow$ & KL $\downarrow$ & Jac.\ faith.\ $\uparrow$ \\
\midrule
\multicolumn{4}{l}{\emph{Our transcoders}} \\
Baseline $k{=}64$    & 0.166 & 0.77 & 0.276 \\
Baseline $k{=}32$    & 0.195 & 1.29 & 0.225 \\
NI $k{=}32$          & 0.209 & 1.16 & 0.238 \\
NFT $k{=}32$, $G{=}16, r{=}4$ & 0.221 & 1.15 & 0.230 \\
SNRL $k{=}32$        & 0.213 & 1.38 & 0.232 \\
\midrule
\multicolumn{4}{l}{\emph{Gemmascope\textsuperscript{$\ddagger$}}} \\
Gemmascope L0-small & 0.156 & 4.59 & 0.202 \\
Gemmascope L0-big  & 0.106 & 1.23 & 0.296 \\
\bottomrule
\end{tabular}
\caption{Reconstruction quality and Jacobian faithfulness on Gemma 3 4B layer 22. FVU and KL are over $1$M held-out SmolLM2 tokens. ``Jac.\ faith.''\ is the cosine similarity between the MLP Jacobian and the transcoder Jacobian, divided by Layernorm for the Gemmascopes: the post-LN scaling factor $(1{+}w_\text{ln})$ is removed so the comparison lives in pre-LN MLP-output space. FVU, however, is measured against a different target for our transcoders and Gemmascopes.}
\label{tab:fvu-kl-jac}
\end{table}

\begin{table}[h]
\centering
\small
\setlength{\tabcolsep}{6pt}
\begin{tabular}{lcc}
\toprule
Clustering & CM acc $\uparrow$ & TNM acc $\uparrow$ \\
\midrule
\multicolumn{3}{l}{\emph{Our transcoders}} \\
Baseline $k{=}32$    & $0.937 \pm 0.005$ & $0.731 \pm 0.030$ \\
NI $k{=}32$          & $0.918 \pm 0.008$ & $0.629 \pm 0.031$ \\
NFT $k{=}32$, $G{=}16, r{=}4$ & $0.918 \pm 0.010$ & $0.598 \pm 0.033$ \\
SNRL $k{=}32$        & $0.927 \pm 0.008$ & $0.657 \pm 0.030$ \\
\midrule
\multicolumn{3}{l}{\emph{Gemmascope}} \\
Gemmascope L0-small & $0.870 \pm 0.016$ & $0.694 \pm 0.035$ \\
Gemmascope L0-big   & $0.793 \pm 0.019$ & $0.645 \pm 0.040$ \\
\midrule
\multicolumn{3}{l}{\emph{Baseline}} \\
$k$-means ($K{=}16$) & $0.933 \pm 0.006$ & $0.764 \pm 0.027$ \\
\bottomrule
\end{tabular}
\caption{Within-cluster sort eval results on 50 random Gemma 3 4B L22 neurons (judge: Claude Sonnet 4.6, 60 trials per neuron). Up to 10 neurons per transcoder produced too few viable clusters and were dropped (more for Gemmascope, which has sparser per-panel cluster coverage). CM acc = Cluster match accuracy (SE over panels), TNM acc = Top-neuron match accuracy (binomial SE over panel-cluster pairs).}
\label{tab:clustering-properties}
\end{table}

\section{Scientific artifacts}
\label{app:artifacts}

\subsection{Released artifacts}

We release the TopK transcoders trained for this paper
(Appendix~\ref{app:setup}) under the MIT License.

\subsection{Models}

\begin{itemize}\itemsep 0pt
  \item Gemma 3 1B/4B/12B/27B pretrained and Gemma 2 2B: \emph{Gemma Terms of Use}.
  \item Llama 3.1 8B and Llama 3.2 1B: \emph{Llama 3.1/3.2 Community License}.
  \item Qwen 3 1.7B/4B/8B/14B: \emph{Apache 2.0}.
\end{itemize}

\subsection{Transcoders}

\begin{itemize}\itemsep 0pt
  \item Gemmascope 2 for Gemma 3 \citep{mcdougall2025gemmascope2}: \emph{CC-BY-4.0}.
  \item LXTC-8x / LXTC-32x for Llama 3.1 8B \citep{he2024llamascope}: \emph{Apache 2.0}.
  \item mwhanna Qwen 3 transcoders: \emph{MIT}.
  \item mntss cross-layer transcoders for Gemma 2 2B and Llama 3.2 1B: used in this work with the permission of the author.
\end{itemize}

\subsection{Datasets}

\begin{itemize}\itemsep 0pt
  \item FineWeb \citep{penedo2024fineweb}: \emph{ODC-BY 1.0}.
  \item SmolLM2 training corpus \citep{allal2025smollm2}: \emph{ODC-BY 1.0} and others.
  \item LMSYS-Chat-1M \citep{zheng2023lmsyschat1m}, used for activation
    caches: \emph{LMSYS-Chat-1M Dataset License}.
  \item Pile-10k, the first 10k examples of the Pile \citep{gao2020pile}:
    \emph{BigScience BLOOM RAIL 1.0}.
  \item Sparse probing datasets \citep{kantamneni2025sparseprobing}:
    while the codebase is MIT, the data used
    has different licenses, described on the Hugging Face pages for individual datasets.
  \item Pride and Prejudice \citep{austen1813pride}: public domain.
\end{itemize}

\subsection{Software}

We used Python $3.13$, CUDA $12.8$, NCCL $2.27.5$, and PyTorch $2.10$.
Table~\ref{tab:software} lists open source libraries and tools used in
the work.

\begin{table}[h]
  \centering
  \small
  \setlength{\tabcolsep}{4pt}
  \begin{tabular}{ll}
    \toprule
    Tool & License \\
    \midrule
    \texttt{sparsify}\tablefootnote{Upstream for our TopK pipeline.} & MIT \\
    \texttt{circuit-tracer} \citep{circuittracer2025} & MIT \\
    \texttt{sae-probes} & MIT \\
    \texttt{nnsight} & MIT \\
    \texttt{transformers}, \texttt{datasets}, \texttt{accelerate}
      & Apache 2.0 \\
    \texttt{vllm}, \texttt{safetensors}, \texttt{huggingface-hub}, \texttt{ftfy}
      & Apache 2.0 \\
    \texttt{numpy}, \texttt{scipy}, \texttt{scikit-learn}, \texttt{joblib}
      & BSD-3-Clause \\
    PyTorch, \texttt{torchao} & BSD-3-Clause \\
    \texttt{einops}, \texttt{tqdm}, \texttt{wandb}, \texttt{faiss}, \texttt{triton}
      & MIT \\
    \texttt{matplotlib} & Matplotlib\tablefootnote{PSF-based, BSD-compatible.} \\
    \texttt{anthropic} & MIT \\
    \bottomrule
  \end{tabular}
  \caption{Open source libraries and tools used.}
  \label{tab:software}
\end{table}

\section{Compute}
\label{app:budget}

All experiments were conducted on a private compute cluster on NVIDIA H200 141GB GPUs. We provide compute usage estimates; most are overestimates and may include experiments that were not part of the results presented in the paper or were run suboptimally.

\subsection{Transcoder training}

Training our TopK transcoders on single layers with most configurations
takes $\sim\!3$ hours per run on one H200. The training runs and
sweeps on Gemma 3 4B L22 took $\sim\!300$ H200-hours. We additionally
train five transcoders on all layers of Gemma 3 4B, totalling
$\sim\!140$ H200-hours. The JumpReLU pipeline sweeps total $225$--$350$
H200-hours.

\subsection{Evaluation}

\begin{itemize}\itemsep 0pt
  \item \textbf{Sparse probing} (Appendix~\ref{app:eval-results:sparse}):
    activation caching took $\sim\!75$ H200-hours; training probes took several thousand CPU-hours.
  \item \textbf{Bigram probing} (Appendix~\ref{app:eval-results:bigram}):
    $\sim\!30$ H200-hours.
  \item \textbf{Linebreaking probing} (Appendix~\ref{app:linebreaking}):
    under $2$ H200-hours total for collecting the residual stream activations.
  \item \textbf{Sort eval} (Appendix~\ref{app:eval-methods:sorteval}):
    activation caching on 10M tokens of SmolLM2 took $\sim\!40$ H200-hours; Claude judging consumes
    $50$--$80$ thousand judge calls on the configurations whose numbers
    appear in the paper.
  \item \textbf{Circuit attribution} (Appendix~\ref{app:circuits}):
    $\sim65$ H200-hours across the transcoders we attribute on.
  \item \textbf{Jacobian alignment}
    (\S\ref{sec:nfc:faithfulness}, Appendix~\ref{app:non-neuron:procedure}):
    under $10$ H200-hours.
  \item \textbf{Distress case study} (Appendix~\ref{app:distress-setup}):
    $\sim\!9$ H200-hours (one circuit on the computed steering vector ---
    $\sim\!3$ H200-hours, caching of maximum activations for neurons ---
    $\sim\!6$ H200-hours).
  \item \textbf{Absorption (\S\ref{sec:clustering:absorption}) and cluster sort eval (\S\ref{sec:clustering:sort-eval})}: caches are built from other steps and the evaluations can be done on CPU. Sonnet judging across the seven clusterings uses $\sim\!23$ thousand calls --- $\sim\!21$k for cluster matching (60 trials $\times$ 50 panels $\times$ 7 clusterings) and $\sim\!1.5$k for the top neuron matching.
\end{itemize}

\subsection{Hyperparameter search}

We searched for optimal learning rates manually and used the same learning rates for all experiments. We varied width $M$, sparsity $k$
(or JumpReLU $L_0$), and seed; the configurations used are in
Appendix~\ref{app:setup}. The sparse probing evaluation sweeps the value of the $L_1$
penalty internally (Appendix~\ref{app:eval-methods:sparse}).

\subsection{Total}

Summing the entries above, the GPU budget for the experiments whose
numbers appear in the paper is on the order of $\textbf{1000}$--$\textbf{1200}$
H200-hours, dominated by transcoder training and by sparse probing's
activation caching. Sparse-probing probe fits after caching add several
thousand CPU-hours separately. Exploratory runs that did not produce
reported numbers are not included.

\section{AI assistance}
\label{app:ai-assistance}

The authors used AI assistants (Claude Code) during this work for
writing assistance (the final version has been rewritten and checked for mistakes), code authoring, and code review. We inspected portions of the implementation code, checked hyperparameters
and investigated and reran experiments selectively (when results were not consistent with past experiments).

\endgroup

\end{document}